\documentclass[10pt,twocolumn,letterpaper]{article}

\usepackage{iccv}
\usepackage{times}
\usepackage{epsfig}
\usepackage{graphicx}
\usepackage{amsmath}
\usepackage{amssymb}
\usepackage{microtype}

\usepackage{algorithm}
\usepackage{algorithmic}
\usepackage{booktabs}
\usepackage[table]{xcolor}
\usepackage{boldline}
\usepackage{multirow}
\usepackage{stmaryrd}
\usepackage{makecell}
\usepackage{subcaption}

\renewcommand{\rightarrow}{\shortrightarrow}
\DeclareMathOperator*{\argmax}{arg\,max}
\newcommand{\revised}[1]{\textcolor{black}{#1}}
\newcommand{\notation}[1]{\textcolor{black}{#1}}
\newcommand{\revisedcr}[1]{\textcolor{black}{#1}}

\definecolor{mygreen}{RGB}{44,160,44}
\definecolor{myblue}{RGB}{31,119,180}

\usepackage[breaklinks=true,letterpaper=true,colorlinks,bookmarks=false]{hyperref}

\iccvfinalcopy %

\def\iccvPaperID{1687} %

\ificcvfinal\fi

\begin{document}

\title{RECALL: Replay-based Continual Learning in Semantic Segmentation}

\author{Andrea Maracani\thanks{These authors share the first authorship.}\footnotemark[1], 
Umberto Michieli\footnotemark[1],
Marco Toldo\footnotemark[1]
 \textsuperscript{,}\thanks{Our work was in part supported by the Italian Minister for Education (MIUR) under the ``Departments of Excellence" initiative (Law 232/2016).}\,, 
Pietro Zanuttigh\\
Department of Information Engineering, University of Padova\\
{\tt\small andreamaracani@gmail.com, \{umberto.michieli,toldomarco,zanuttigh\}@dei.unipd.it}
}

\maketitle
\ificcvfinal\thispagestyle{empty}\fi

\begin{abstract}
Deep networks allow to obtain outstanding results in semantic segmentation, however they need to be trained in a single shot with a large amount of data. Continual learning settings where new classes are learned in incremental steps and previous training data is no longer available are challenging due to the catastrophic forgetting phenomenon. Existing approaches typically fail when several incremental steps are performed or in presence of a distribution shift of the background class. We tackle these issues by recreating no longer available data for the old classes and outlining a content inpainting scheme on the background class. We propose two sources for replay data. The first resorts to a generative adversarial network to sample from the class space of past learning steps. The second relies on web-crawled data to retrieve images containing examples of old classes from online databases. In both scenarios no samples of past steps are stored, thus avoiding privacy concerns. Replay data are then blended with new samples during the incremental steps. Our approach, RECALL, outperforms state-of-the-art methods.
\end{abstract}

\section{Introduction}
\label{sec:intro}

A common requirement for many machine learning applications is the ability to learn a sequence of tasks in multiple incremental steps, \eg, progressively introducing novel classes to be recognized, instead of using a single-shot training procedure on a large dataset \cite{parisi2019continual}. This problem has been widely studied in
image classification %
and many methods propose to alleviate the forgetting of previous tasks and intransigence of learning new ones %
\cite{kirkpatrick2017overcoming,rebuffi2017icarl,zenke2017continual}. When the model is exposed to samples of novel classes and  is trained on them without additional provisions, the optimization leads to the so-called \textit{catastrophic forgetting} phenomenon \cite{robins1995catastrophic,goodfellow2013empirical}, \ie, knowledge about previously seen classes tends to be lost. 

Incremental learning on dense tasks (\eg, semantic segmentation), where pixel-wise predictions are performed, has only recently been explored and the first experimental studies show that catastrophic forgetting is even more severe than on the classification task \cite{michieli2019incremental,michieli2020knowledge}. 
Current approaches for class-incremental semantic segmentation re-frame knowledge distillation strategies inspired by previous works on image classification \cite{michieli2019incremental,cermelli2020modeling,klingner2020class,michieli2020knowledge}. Although they partially alleviate forgetting, they often fail when multiple  incremental steps are performed or when background shift \cite{cermelli2020modeling} (\ie, change of statistics of the background across learning steps, as it incorporates old or future classes) occurs.

\begin{figure}[t]
\includegraphics[trim=5.5cm 6cm 6cm 4cm, clip,width=\linewidth]{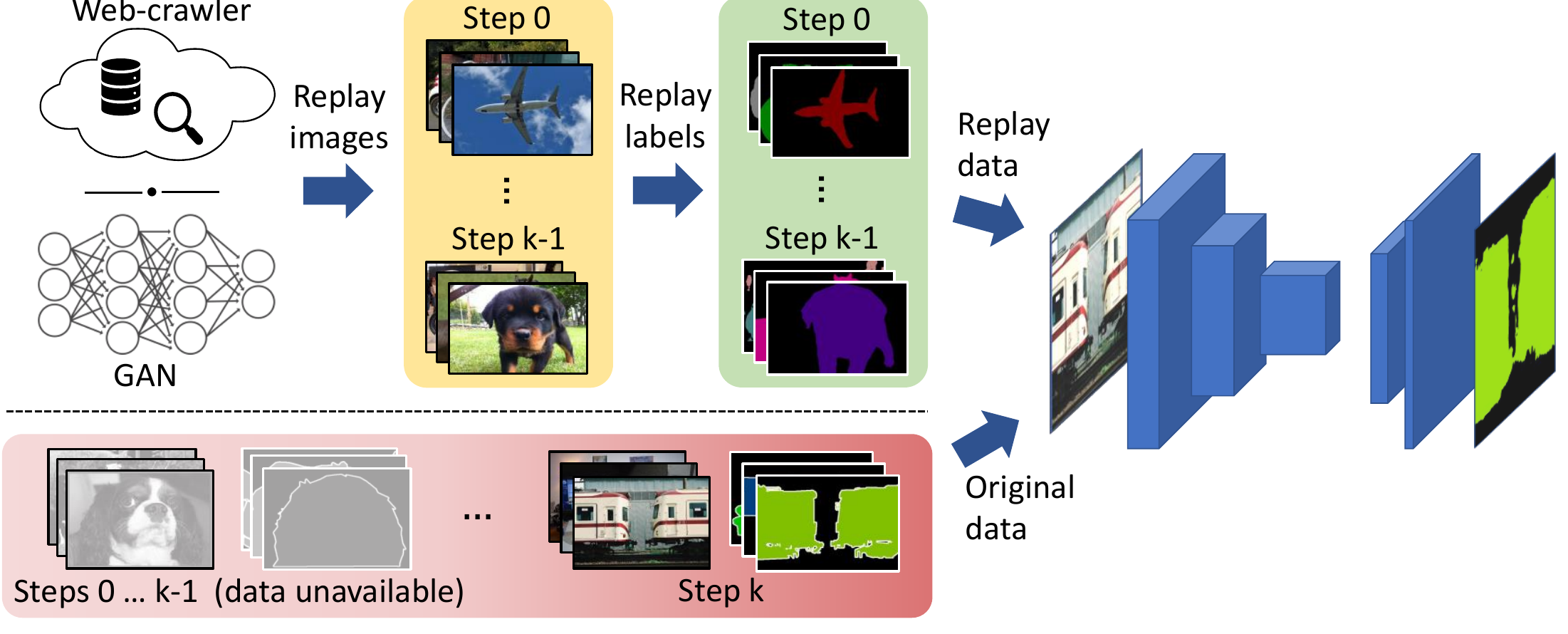}
\caption{ %
Replay images of previously seen classes are retrieved by a web crawler or a generative network and further labeled. 
Then, the network is incrementally trained with a mixture of new and replay data. %
}\vspace{-0.35cm}
\label{fig:GA}
\end{figure}

In this paper, we follow a completely different strategy and, instead of distilling knowledge from a teacher model (\ie, the old one) to avoid forgetting, we propose to generate samples of old classes by using replay strategies. %
We propose RECALL (REplay in ContinuAL Learning), a method that re-creates representations of old classes and mixes them with the available training data, \ie, containing novel classes being learned (see Fig.~\ref{fig:GA}). 
To reduce background shift we introduce a self-inpainting strategy that re-assigns the background region %
 according to predictions of the previous model. %

To generate representations of past classes we pursue two possible directions. The first is based on %
a pre-trained generative model, \ie, a Generative Adversarial Network (GAN) \cite{goodfellow2014} conditioned to produce samples of an input class. The GAN has been trained beforehand on a \revisedcr{dataset different} than the target one (we chose ImageNet as it comprehends a wide variety of classes and domains), thus requiring a Class Mapping Module to perform the translation %
between the two label spaces.
The second strategy, instead, is based on crawling images from the web, querying the class names to drive the search. 
Both approaches allow to retrieve a large amount of weakly labeled data. %
Finally, we generate pseudo-labels for semantic segmentation using a side labeling module, which requires only minimal extra storage. %

Our main contributions are:  1) we propose RECALL, which is the first approach to use replay data in continual semantic segmentation; 2) to the best of our knowledge, we are the first to introduce the webly-supervised paradigm in continual learning, showing how we can extract useful clues from extremely weakly supervised and noisy samples; 3) we devise a background inpainting strategy to generate pseudo-labels and overcome the background shift; 4) we achieve state-of-the-art results on a wide range of scenarios, especially when performing multiple incremental steps.

\section{Related Works}
\label{sec:related}

\textbf{Continual Learning (CL).}
Deep neural networks witnessed remarkable improvements in many fields; however, such models are prone to catastrophic forgetting when they are trained to continuously improve the learned knowledge (\eg, new categories) from progressively provided data %
\cite{goodfellow2013empirical}.
Catastrophic forgetting is a long-standing problem \cite{robins1995catastrophic,french1999catastrophic} which has been recently tackled in a variety of visual tasks such as image classification \cite{kirkpatrick2017overcoming,rebuffi2017icarl,li2018learning,wu2018incremental,ostapenko2019learning}, object detection \cite{shmelkov2017incremental,li2019efficient} and semantic segmentation \cite{michieli2019incremental,michieli2020knowledge,cermelli2020modeling,klingner2020class}.

Current techniques can be grouped into four main (non mutually exclusive) categories \cite{lesort2019continual}: namely, dynamic architectures, regularization-based, rehearsal and generative replays. Dynamic architectures can be explicit \cite{wang2017growing,li2018learning}, if new network branches are grown, or implicit \cite{fernando2017pathnet,serra2018overcoming}, if some network weights are available for certain tasks only.
Regularization-based approaches mainly propose to compute some penalty terms to regularize training (\eg, based on the importance of the weights for a specific task) \cite{kirkpatrick2017overcoming,zenke2017continual} or to distill knowledge from the old model \cite{shmelkov2017incremental,li2018learning,michieli2020knowledge}. Rehearsal approaches store a set of raw samples of past tasks into memory, which are then used while training for the new task \cite{rebuffi2017icarl,lopez2017gradient}. Finally, generative replays approaches \cite{shin2017continual,wu2018incremental,kamra2017deep} rely on generative models typically trained on the same data distribution, which are later used to generate artificial samples to preserve previous knowledge. Generative models are usually GANs \cite{shin2017continual,wu2018incremental,he2018exemplar} or auto-encoders \cite{kamra2017deep}.
In this work, we employ two kinds of generative replays: either resorting to a standard pre-trained GAN %
or to web-crawled images to avoid forgetting, without storing any of the samples related to previous tasks. When using the generative model, differently from previous works on continual image classification, we do not select real exemplars as anchor to support the learned distribution \cite{he2018exemplar} nor we train or fine-tune the GAN architecture on the current data distribution \cite{shin2017continual,he2018exemplar,wu2018incremental}, thus reducing  memory and  computation time.

\textbf{CL in Semantic Segmentation.}
Semantic segmentation has experienced a wide research interest in the past few years and deep networks have achieved noticeable results on this task. Current techniques are based on   the auto-encoder structure firstly employed by FCN \cite{long2015} and subsequently improved by many approaches \cite{chen2017rethinking,chen2018deeplab,zhao2017,yu2016}. Recently, increasing attention has been devoted to class-incremental semantic segmentation \cite{michieli2019incremental,cermelli2020modeling,michieli2021continual,douillard2021plop} to learn new categories from new data. In \cite{michieli2019incremental,michieli2020knowledge} the problem is first introduced and tackled with regularization approaches such as parameters freezing \revised{(\eg, fixing the encoder after the initial training stage)} and knowledge distillation. In \cite{klingner2020class} knowledge distillation is coupled with a class importance weighting scheme to emphasize gradients on difficult classes. Cermelli \etal \cite{cermelli2020modeling} study the distribution shift of the background class. In \cite{douillard2021plop} long- and short-range spatial relationships at feature level are preserved. In \cite{michieli2021continual} the latent space is regularized to improve class-conditional features separation.

\textbf{Webly-Supervised Learning} 
is an emerging paradigm in which large amounts of web data is exploited for learning CNNs \cite{chen2015webly,divvala2014learning,niu2018webly}. Recently, it was employed also in semantic segmentation to provide a plentiful source of both images \cite{jin2017webly,shen2018bootstrapping} and videos \cite{hong2017weakly} with weak image-level class labels during training. The most active research directions are devoted toward understanding how to query images, how to filter and exploit them (\eg, assigning pseudo-labels). To our knowledge, however, webly-supervised learning has not yet been explored in continual learning as a replay strategy.

\section{Problem Formulation}
\label{sec:problem}

The semantic segmentation task consists in labeling each  pixel in an image by assigning it to a class %
from a collection of possible semantic classes %
$\mathcal{C}$, which typically also comprises a special background category that we denote as $b$. 
More formally, given an image $\mathbf{X} \in \mathcal{X} \subset \mathbb{R}^{H \times W \times 3}$,  we aim at producing a map $\mathbf{\hat{Y}} \in \mathcal{Y}\subset \mathcal{C}^{H \times W}$ that is a prediction of the ground truth map $\mathbf{Y}$. This is nowadays usually achieved  by using a suitable deep learning model $M: \mathcal{X} \mapsto \mathbb{R}^{H \times W \times |\mathcal{C}|}$, commonly made by a feature extractor $E$ followed by a decoding module $D$, \ie, \revised{$M = D \circ E$}.

In  standard supervised learning, the model is learned in a single shot over a training set $\mathcal{T} \subset \mathcal{X} \times \mathcal{Y}$,  available in its complete form to the training algorithm. In  class-incremental learning, instead, we assume that the training is performed in multiple steps and only a subset of  training data is made available to the algorithm at each  step ${k=0,...,K}$.
More in detail, we start from an initial step ${k=0}$ where only training data concerning a subset of all the classes $\mathcal{C}_0 \subset \mathcal{C}$ is available (we assume that $b \in \mathcal{C}_0$). %
We denote with ${M_0: \mathcal{X} \mapsto \mathbb{R}^{H \times W \times |\mathcal{C}_0|} }$, $M_0 = D_0 \circ E_0$ the model trained after this initial step.
Moving to a generic step $k$, a new set of classes $\mathcal{C}_k$ is added to the class collection $\mathcal{C}_{0\rightarrow (k-1)}$ learned up to that point, %
resulting in an expanded set of learnable classes $\mathcal{C}_{0\rightarrow k}=\mathcal{C}_{0\rightarrow (k-1)} \cup \mathcal{C}_k$ (we assume ${\mathcal{C}_{0\rightarrow (k-1)} \cap \mathcal{C}_k = \varnothing}$).
The model after the $k$-th step of training is  ${M_k: \mathcal{X} \mapsto \mathbb{R}^{H \times W \times |\mathcal{C}_{0 \rightarrow k}|} }$, where $M_k = D_k \circ E_0 $, since in our approach the encoder $E_0$ is not trained during the incremental steps and only the decoder is updated \cite{michieli2019incremental}.

Two main continual scenarios \revisedcr{have been proposed (see \cite{michieli2019incremental,michieli2021continual,cermelli2020modeling} for a more detailed description)} and we tackle both in a unified framework. 

\noindent \textbf{Disjoint setup}: 
in the initial step all the images in the training set with at least one pixel belonging to a class of $\mathcal{C}_0$ (except for $b$) are assumed to be available. We denote with $\mathcal{Y}_{\mathcal{C}_0 \cup \{ b \}} \subset \mathcal{C}_0^{H \times W}$ the corresponding output space where labels can only belong to  $\mathcal{C}_0$, while all the pixels not pertaining to  these classes are assigned to $b$. 
The incremental partitions are built as \textit{disjoint} subsets of the whole training set.
The training data associated to $k$-th step, $\mathcal{T}_k \subset \mathcal{X} \times \mathcal{Y}_{\notation{\mathcal{C}_k} \cup \{ b \}}$, contains only images corresponding to classes in $\mathcal{C}_k$ with just classes of step $k$ annotated (possible old classes are labeled as $b$), and is disjoint w.r.t.\ previous and past partitions. 

\noindent \textbf{Overlapped setup}: 
in the first phase we select the subset of training images having only $\mathcal{C}_0$-labeled pixels.
Then, the training set  at each  incremental step contains \textit{all} the images with labeled pixels from $\mathcal{C}_k$
\ie ${ \mathcal{T}_k \subset \mathcal{X} \times \mathcal{Y}_{\notation{\mathcal{C}_k} \cup \{ b \}} }$.
Similarly to the initial step, labels are limited to semantic classes in  $\mathcal{C}_k$, while remaining pixels are assigned to $b$. 

In both setups, $b$ undergoes a semantic shift at each step, %
as pixels of ever changing class sets are assigned to it.

\section{General Architecture}
\label{sec:gen_arc}

\begin{figure*}[h]
\includegraphics[width=\linewidth]{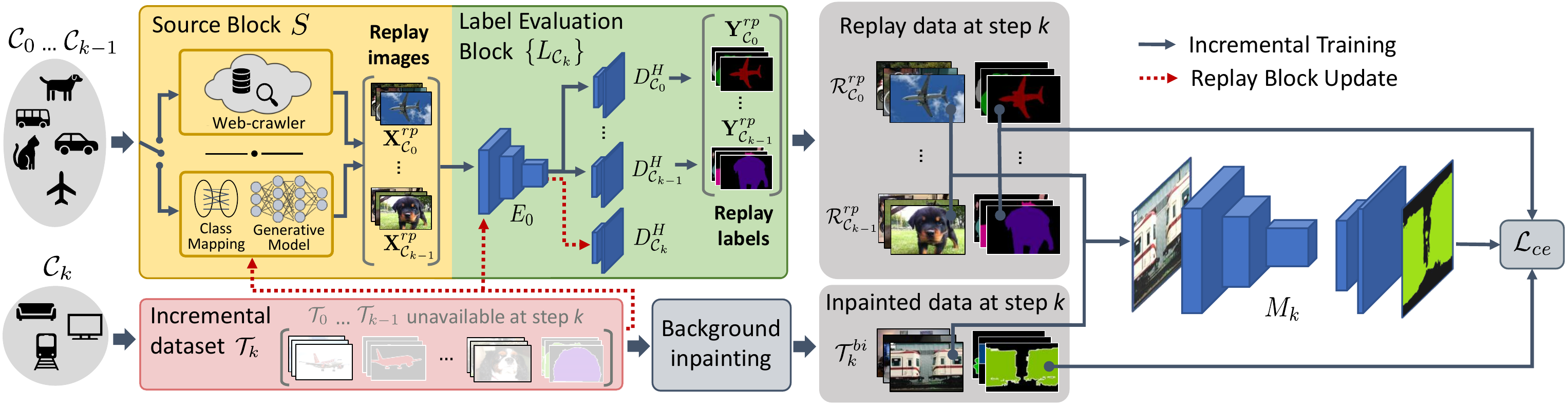}
\vspace{-0.2cm}
\caption{ Overview of the proposed RECALL: 
class labels from past incremental steps are provided to a Source Block, either a web crawler %
or a pre-trained conditional GAN, %
which retrieves a set of unlabeled replay images for the past semantic classes. 
Then, a Label Evaluation Block produces the missing annotations.  
Finally, the segmentation network is incrementally trained with a replay-augmented dataset, composed of both new classes data and replay data. 
}
\label{fig:architecture}
\end{figure*}

In the standard setup, the  segmentation model $M$ is trained with annotated samples from a training set $\mathcal{T}$. 
Data should be representative of the task we would like to solve, meaning that multiple instances of all the considered semantic classes $\mathcal{C}$ should be available in the provided dataset for the segmentation network to properly learn them. %
Once $\mathcal{T}$ has been assembled, the cross-entropy objective is commonly employed to optimize the weights of $M$:
\vspace{-0.2cm}
\begin{equation}
\! \! \mathcal{L}_{ce}(M;\mathcal{C},\mathcal{T}) \! = \! - \frac{1}{|\mathcal{T}|} \!
\sum_{\mathbf{X}, \mathbf{Y} \in \mathcal{T}} %
\sum_{c \in \mathcal{C}} 
\mathbf{Y}[c] \cdot \log\big(M(\mathbf{X})[c]\big)
\label{eq:L_CE}
\vspace{-0.2cm}
\end{equation}
In the incremental learning setting, when performing an incremental training step $k$ only samples related to new classes $\mathcal{C}_k$ are assumed to be at our disposal. 
Following the simplest approach, we could initialize our model's weights from the previous step ($M_{k-1}$, $k \geq 1$) and learn the segmentation task over classes $\mathcal{C}_{0 \rightarrow k}$ by optimizing the standard objective $\mathcal{L}_{ce}(M_k; \mathcal{C}_{0 \rightarrow k}, \mathcal{T}_k)$ with data from the current training partition $\mathcal{T}_k$. 
However, simple fine-tuning leads to catastrophic forgetting, being unable to preserve previous knowledge.

\noindent
\textbf{Architecture of Replay Block.}
To cope with this issue, we opt for a replay strategy. 
Our goal is to retrieve task-related knowledge of past classes to be blended into the ongoing incremental step, all without accessing training data of previous iterations. 
To this end, we introduce a Replay Block, whose target is twofold.
First, it has to provide images resembling instances of classes from previous steps, whether generating them from scratch or retrieving them from an available alternative source (\eg, a web database). 
Second, it has to obtain reliable semantic labels of those images, by resorting to learned knowledge from past steps. 

The Replay Block's image retrieval task is executed by what we call Source Block:
\vspace{-0.2cm}
\begin{equation}
S : \: \mathcal{C}_k \mapsto \mathcal{X}^{rp}_{\mathcal{C}_k}
\vspace{-0.2cm}
\end{equation}
This module takes in input a set of classes $\mathcal{C}_k$ (background excluded) and provides images 
whose semantic content can be ascribed to those categories 
(\eg, $\mathbf{X}^{rp} \in \mathcal{X}^{rp}_{\mathcal{C}_k}$). 
We adopt two different solutions for the Source Block, namely GAN and web-based techniques,  both detailed in Sec.~\ref{sec:replay}. %

The Source Block provides unlabeled image data (if we exclude the weak image-level classification labels), and for this reason we introduce an additional %
Label Evaluation Block $\{ L_{\mathcal{C}_k} \}_{\mathcal{C}_k \subset \mathcal{C}}$, %
which aims at annotating examples provided by the replay module.  
This block is made of separate instances $L_{\mathcal{C}_k} = D_{\mathcal{C}_k}^H \circ E_0$, each denoting a segmentation model to classify a specific set of semantic categories $\mathcal{C}_k \cup \{b\}$ (\ie, the classes in $\mathcal{C}_k$ plus the background) :
\vspace{-0.15cm}
\begin{equation}
L_{\mathcal{C}_k} : \mathcal{X}_{\mathcal{C}_k} \mapsto \mathbb{R}^{H \times W \times (\notation{|{\mathcal{C}_k} \cup \{b\}}|)}
\vspace{-0.2cm}
\end{equation}
All $L_{\mathcal{C}_k}$ modules share the encoder section $E_0$ from the initial training step, so that only a minimal portion of the segmentation network (\ie, $D^{H}_{\mathcal{C}_k}$, which accounts for only few parameters, see Sec.~\ref{subsec:ablation}) is stored for each block's instance.
 Notice that a single instance recognizing all classes could be used,  leading to an even more compact representation, but it experimentally led to worse performance.

Provided that $S$ and $L_{\mathcal{C}_k}$ are available, replay training data can be collected for classes in $\mathcal{C}_k$. 
A query to $S$ outputs a generic image example $\mathbf{X}^{rp}_{\mathcal{C}_k} = S (\mathcal{C}_k) $, 
which is then associated to its prediction 
$\mathbf{Y}^{rp}_{\mathcal{C}_k} = \argmax \limits_{c \in \mathcal{C}_k \cup \{b\}}  L_{\mathcal{C}_k}(\mathbf{X}^{rp}_{\mathcal{C}_k}) [ c ] $.
By retrieving multiple replay examples, \revisedcr{we build a replay dataset $\mathcal{R}_{\mathcal{C}_k} = \{ ( \mathbf{X}^{rp}_{\mathcal{C}_k} , \mathbf{Y}^{rp}_{\mathcal{C}_k} )_n \}_{n=1}^{N_{r}} $, where $N_{r}$ is a fixed hyperparameter empirically set (see section \ref{sec:implementation})}. %

\begin{figure}[tbp]
\includegraphics[trim=2cm 8cm 15.1cm 2.2cm, clip,width=0.96\linewidth]{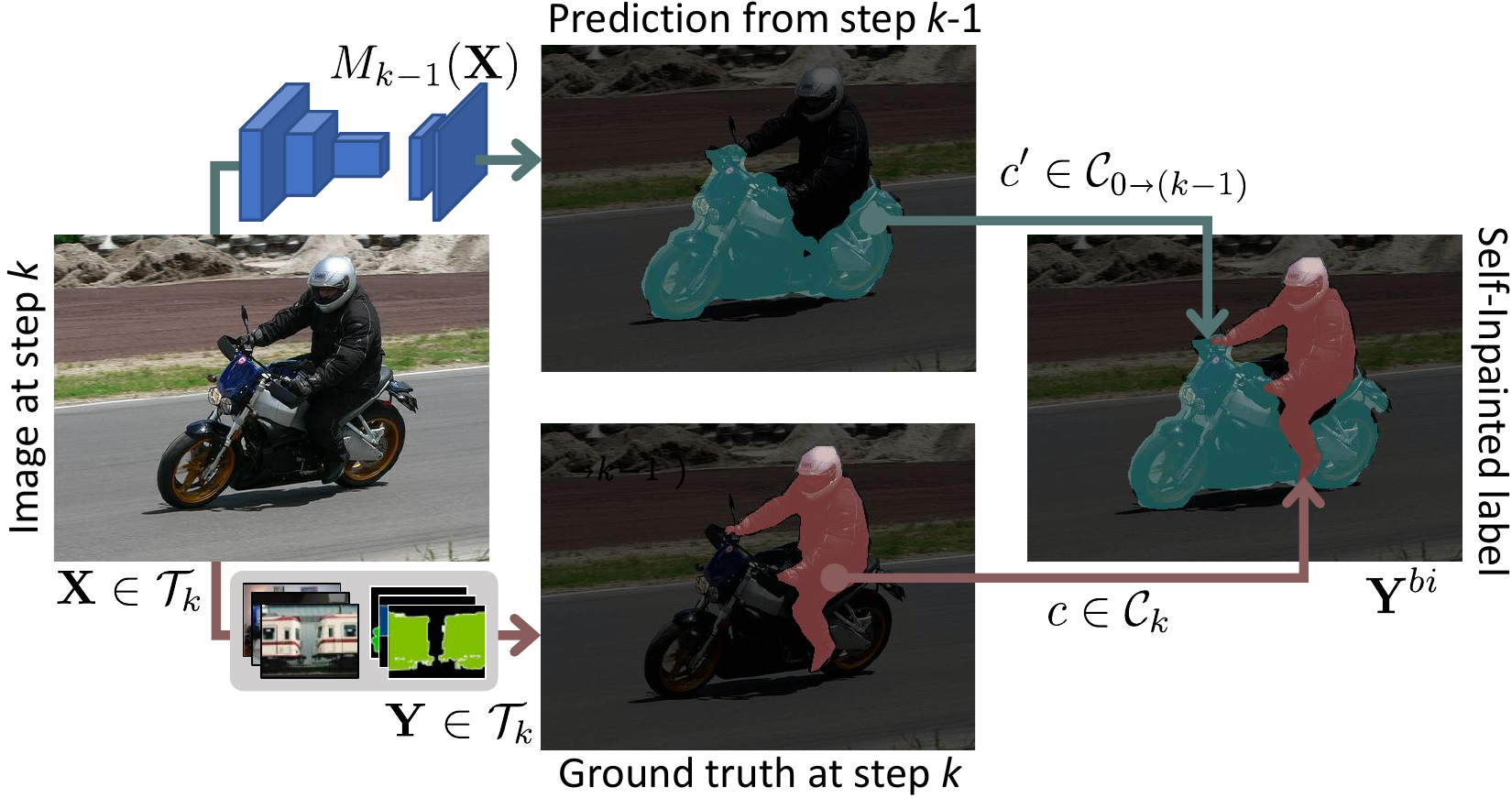}
\vspace{-0.2cm}
\caption{ \revised{Background self-inpainting process.}
}
\label{fig:BGR}
\end{figure}

\noindent
\textbf{Background Self-Inpainting.} 
To deal with the background shift phenomenon, we propose a simple yet effective inpainting mechanism to transfer knowledge from the previous model into the current one. 
While the replay block re-creates samples of previously seen classes, background inpainting acts on background regions of current samples reducing the background shift and at the same time bringing a regularization effect similar to knowledge distillation \cite{michieli2019incremental,cermelli2020modeling}, although its implementation is quite different.
At every  step $k$ with training set  $\mathcal{T}_k$, we take the background region of each ground truth map and we label it with the associated prediction from the previous model $M_{k-1}$ \revised{(see Fig.~\ref{fig:BGR})}. We call it \textit{background inpainting} since the background regions in label maps are changed according to a self-teaching scheme based on the prediction of the old model.
\revised{More formally, we replace each original label map $\mathbf{Y}$ available at step $k>0$ with its inpainted version $\mathbf{Y}^{bi}$:} %
\vspace{-0.1cm}
\begin{equation}
\mathbf{Y}^{bi} [h,\!w]  \!=\!  
\begin{cases}
    \mathbf{Y} [h,\!w]           &  \text{if} \: \mathbf{Y} [h,\!w]\! \in\! \mathcal{C}_k \\
     \argmax \limits_{c \in \mathcal{C}_{0 \rightarrow k-1}}  M_{k - 1} (\mathbf{X})   [h,\!w][c]  &  \text{otherwise}
\end{cases}
\end{equation}
\revised{where $(\mathbf{X},\mathbf{Y}) \in \mathcal{T}_{k}$, while $[h, w]$ denotes the pixel coordinates. Labels at step $k=0$ are not inpainted, as at that stage we lack any prior knowledge of past classes. When background inpainting is performed, each set $\mathcal{T}_{k}^{bi} \subset \mathcal{X}\times \mathcal{Y}_{\mathcal{C}_{0 \rightarrow k}}$ ($k>0$) contains all samples of $\mathcal{T}_k$ after being inpainted.}

\noindent
\textbf{Incremental Training with Replay Block.}
The training procedure of RECALL is detailed and summarized in Algorithm\ \ref{algorithm} and the process is described in Fig.~\ref{fig:architecture}. 
Suppose we are at the incremental step $k$, with only training data of classes in $\mathcal{C}_k$ from partition $\mathcal{T}_k$ available. 
In a first stage, the Replay Block is fixed and it is used to retrieve annotated data for steps from $0$ to $k-1$ uniformly distributed among all the past classes. 
Following the described pipeline,  the generative and labeling models are applied independently over each incremental class set  $ \mathcal{C}_i, i=0,...,k-1$. 
The replay training dataset for step $k$ is the union of the single replay sets for each previous step: $\mathcal{R}_{\mathcal{C}_{0 \rightarrow (k-1)}} = \bigcup\limits_{i=0}^{k-1} 
\mathcal{R}_{\mathcal{C}_i }$. 
Once we have assembled $\mathcal{R}_{\mathcal{C}_{0 \rightarrow (k-1)}}$, by merging it with $\mathcal{T}_k^{bi}$ we get an augmented step-$k$ training partition $\mathcal{T}_k^{rp} = \mathcal{T}_k^{bi} \cup \mathcal{R}_{\mathcal{C}_{0 \rightarrow (k-1)}}$.   
This new set, in principle, is complete of annotated samples containing both old and new classes, thanks to replay data. 
Therefore, we effectively learn the segmentation model $M_k$ through the cross-entropy objective {$\mathcal{L}_{ce}(M_k; \mathcal{C}_{0 \rightarrow k}, \mathcal{T}_k^{rp})$} on replay-augmented training data.
This mitigates the bias toward new classes, thus
preventing %
 forgetting. %

\begin{algorithm}[tbp]
\caption{RECALL: incremental training procedure.}
\vspace{1mm}
\textbf{Input:} $\{ \mathcal{T}_k \}_{k=0}^K$ and $\{ \mathcal{C}_k \}_{k=0}^K$
\\
\textbf{Output:} $M_K$ \\[0.3ex]
train $M_0 = E_0 \circ D_0$ with $\mathcal{L}_{ce}(M_0; \mathcal{C}_{0}, \mathcal{T}_0)$ \\
train $S$ on $(\mathcal{C}_0 , \mathcal{T}_0 )$ \\
train $D^H_{\mathcal{C}_0}$ with $\mathcal{L}_{ce}(L_{\mathcal{C}_0}; \mathcal{C}_{0}, \mathcal{T}_0)$ \\
\textbf{for} $k \leftarrow 1$ to $K$ \textbf{do} \\
\hspace*{4mm} background inpainting on $\mathcal{T}_k$ to obtain $\mathcal{T}_k^{bi}$ \\
\hspace*{4mm} train $S$ on $(\mathcal{C}_k , \mathcal{T}_k )$ \\
\hspace*{4mm} train $D^H_{\mathcal{C}_k}$ with $\mathcal{L}_{ce}(L_{\mathcal{C}_k}; \mathcal{C}_{k} \cup \{ b \}, \mathcal{T}_k)$ \\
\hspace*{4mm} generate $\mathcal{T}_k^{rp} = \mathcal{T}_k^{bi} \cup \mathcal{R}_{\mathcal{C}_{0 \rightarrow (k-1)}}$ \\
\hspace*{4mm} train $D_k$ with $\mathcal{L}_{ce}(M_k; \mathcal{C}_{0 \rightarrow k}, \mathcal{T}^{rp}_k)$ \\
\textbf{end for}
\vspace{1mm}
\label{algorithm}
\end{algorithm}

In a second stage, we exploit $\mathcal{T}_k$ to train the Class Mapping Module if needed (see Sec.~\ref{sec:replay}). 
In particular, we teach the Source Block $S$ to produce samples of $\mathcal{C}_k$, and we optimize the  decoder $D^H_{\mathcal{C}_k}$ to correctly segment, in conjunction with $E_0$, images from $\mathcal{T}_k$ by minimizing $\mathcal{L}_{ce}(L_{\mathcal{C}_k}; \mathcal{C}_k \cup \{b\}, \mathcal{T}_k)$. This stage is not exploited in the current step, but will be necessary in future ones. %

During a standard incremental training stage, %
we follow a mini-batch gradient descent scheme, where batches of annotated training data are sampled from $\mathcal{T}_k^{rp}$.
However, to guarantee a proper stream of information, we opt for an interleaving sampling policy, rather than a random one. 
In particular, at a generic iteration of training, a batch of data $\mathcal{B}^{rp}$ supplied to the network is made of $r_{new}$ samples from the current training partition $\mathcal{T}_k^{bi}$ and $r_{old}$ replay samples from $\mathcal{R}_{\mathcal{C}_{0 \rightarrow (k-1)}}$. 
The ratio between $r_{new}$ and $r_{old}$ controls the proportion of replay and new data (see also Sec.~\ref{subsec:ablation}). 
We need, in fact, to carefully balance how new data is dosed with respect to replay one, so that enough information about new classes is provided within the learning process, 
while concurrently we assist the network in recalling knowledge acquired in past steps to prevent catastrophic forgetting.

\section{Replay Strategies}
\label{sec:replay}

In this section we describe more in detail the replay strategies employed for the image generation task of the Source Block $S$. 
As mentioned previously, we opt for a generative approach based on a GAN framework and for an online retrieval solution, where images are collected by a web crawler.

\noindent
\textbf{Replay by GAN. } 
The GAN-based strategy exploits a deep generative adversarial framework to re-create the no longer available samples for previously seen classes.
We use a conditional GAN, $G$, pre-trained on a generic large-scale visual dataset with data from a wide set of semantic classes $\mathcal{C}^{G}$ and different domains. \revised{For the experiments, we choose %
an ImageNet \cite{deng2009imagenet} based pre-training. On this regard, we remark that classes and domains are not required to be completely coherent: for instance \textit{person} does not exist in ImageNet, but related classes (\eg, \textit{hat}) 
still allow to preserve its knowledge (further considerations on this are reported in Suppl. Mat.).}
When performing the $k$-th incremental step, we retrieve images containing previously seen classes by sampling the GAN's generator output, \ie, $\mathbf{X}^{rp}=G(\mathbf{n},c^{G})$ conditioned on GAN's classes $c^{G} \in \mathcal{C}^{G}$ corresponding to the target ones from the original training data ($\mathbf{n}$ is a generic noise input). 

Since the GAN is pre-trained on a separate dataset, typically it inherits a different label set.
For this reason, the Source Block with GAN is composed of two main modules, namely the actual GAN for image generation and a Class Mapping Module to translate each class of the semantic segmentation incremental dataset %
to the most similar class of the GAN's training dataset. 
Provided that we have trained both the GAN and class mapping modules, first we use the latter %
to translate the class set $\mathcal{C}_k$ to the matching set $\mathcal{C}^{G}_k$.
Then, a set of queries to the conditioned GAN's generator:  
\vspace{-0.2cm}
\begin{equation}
 \mathbf{X}^{rp}_{\mathcal{C}_k}=G(\mathbf{n},c^{G}), \quad c^{G} \in \mathcal{C}^{G}_k
 \label{eq:gan}
 \vspace{-0.2cm}
 \end{equation}
provides samples resembling the ones in $\mathcal{C}_k$, as long as the mapping is able to 
properly associate each original class to 
a statistically similar counterpart in the GAN's label space.

At each incremental step $k$, the Source Block with GAN goes through two separate training and inference stages. 
In a first training phase, samples from $\mathcal{T}_k$ are fed to an Image Classifier $I$, which is pre-trained to solve an image classification task on the GAN's dataset. 
In particular, for each class $c \in \mathcal{C}_k$ we select the corresponding training subset $\mathcal{T}^c_k \subset \mathcal{T}_k$, \ie, all the samples of set $\mathcal{T}_k$ associated to class $c$, and we sum the resulting class probability vectors from the classification output. 
Then, the GAN's class $c^{G}$ with the highest probability score is identified by:
\vspace{-0.1cm}
\begin{equation}
    c^{G}=\argmax_{j \in \mathcal{C}^{G}}  \sum_{\textbf{X} \leftarrow \mathcal{T}_k^c}   I(\textbf{X})  \left[  j \right]
\vspace{-0.1cm}
 \label{eq:mapper}
 \end{equation}
 where $\mathbf{X}$ is extracted from $\mathcal{T}_k^c$ (labels are not used) and $I(\textbf{X})$ denotes the vector output of the last softmax layer of $I$, whose $j$-th entry corresponds to the $j$-th GAN's class. %
By repeating this procedure for every class in $\mathcal{C}_k$, we build the mapped set $\mathcal{C}^{G}_k$. 
Class correspondence is stored, so that at each step we have access to class mappings of past iterations.  

In a second evaluation phase, classes in $\mathcal{C}_{0 \rightarrow (k-1)}$ are given as input to the Source Block.
Thanks to the class correspondences saved in previous steps, $\mathcal{C}_{0 \rightarrow (k-1)}$ are mapped to $\mathcal{C}_{0 \rightarrow (k-1)}^{G}$. 
Next, image generation conditioned on each class of $\mathcal{C}_{0 \rightarrow (k-1)}^{GAN}$ is performed, and the resulting replay images are fed to the Label Evaluation Block 
to be associated to their corresponding semantic labels.
By following this procedure, we end up with self-annotated data of past classes suitable %
to support the supervised training at the current step, which otherwise would be limited to new classes.

\noindent
\textbf{Replay by Web Crawler. }
As an alternative we propose to retrieve training examples from an online source.  
For the evaluation, we searched images from  the $Flickr$ website, %
but any other online database or search engine can be used. 

Assuming we are at the incremental step $k$ and we have access to the names of every class in the past iterations (\eg, $\forall c \in \mathcal{C}_{0 \rightarrow (k-1)}$), we download images whose tag and description happen to both contain the class name through the $Flickr$'s web crawler.
Then, the web-crawled images are fed to the Label Evaluation Block for their annotation.

Compared to the GAN-based approach, the online retrieval solution is  simpler as no learnable modules are introduced. 
In addition, we completely avoid %
\revisedcr{to assume that a larger dataset is available},
whose class range should be sufficiently ample and diverse to cope with the continuous stream of novel classes incrementally introduced. 
On the other side, this approach requires the availability of \revised{an} internet connection \revisedcr{and in some way exploits additional training data even if almost unsupervised. Plus,} we lack  control over the weak labeling performed by the web source.

\section{Implementation Details}
\label{sec:implementation}

We use the DeepLab-V2 \cite{chen2018deeplab} as segmentation architecture with ResNet-101 \cite{he2016deep} as backbone. 
Nonetheless, RECALL is independent of the specific network architecture. %
Encoder's weights are pre-trained on ImageNet \cite{deng2009imagenet} and all network's weights are trained in the initial step $0$. 
In the following steps, only the main decoder is trained, together with the additional $\{ D^{H}_{\mathcal{C}_{k}} \}_k$ helper decoders, which are needed to annotate replay samples (as discussed in Sec.~\ref{sec:gen_arc}). 
\revised{For fair comparison, all competing approaches are trained with the same backbone.}
SGD with momentum is used for weights optimization, with initial learning rate set to $5 \times 10^{-4}$ and decreased to $5 \times 10^{-6}$ according to polynomial decay of power $0.9$. %
Following previous works \cite{michieli2019incremental,michieli2020knowledge}, we train the model for $|\mathcal{C}_k| \times 1000$ learning steps in the disjoint setup and for $|\mathcal{C}_k| \times 1500$ steps in the overlapped setup.
Each helper decoder $D^{H}_{C_{k}}$ is trained with a polynomially decaying learning rate starting from $2 \times 10^{-4}$ and ending at $2 \times 10^{-6}$ for $|\mathcal{C}_k| \times 1000$ steps. %
As Source Block, we use BigGAN-deep \cite{brock2018large} pre-trained \cite{biggan_weights} on ImageNet. %
\revisedcr{At each incremental step $k$, we generate 500 replay samples per old class, \ie $N_r=500$%
.} 
To map classes from the segmentation dataset to the GAN's one, we use the EfficientNet-B2 \cite{tan2019efficientnet} classifier implemented at \cite{efficientnet_weights} and pre-trained on ImageNet.  The interleaving ratio $r_{old}/r_{new}$ is set to $1$.

\revisedcr{As input pre-processing, random scaling and mirroring are followed by random padding and cropping to $321 \times 321$ px. The entire framework is developed in TensorFlow \cite{abadi2016tensorflow} and trained on a single NVIDIA RTX 2070 Super. Training time varies depending on the setup, with the longest run taking about $5$ hours. Code and replay data are available at \url{https://github.com/LTTM/RECALL}.}

\section{Experimental Results}
\label{sec:results}

\begin{figure}[htbp]
\centering
\vspace{-0.15cm}
\includegraphics[trim=1.1cm 0.9cm 1.8cm 2.5cm, clip, width=0.82\linewidth]{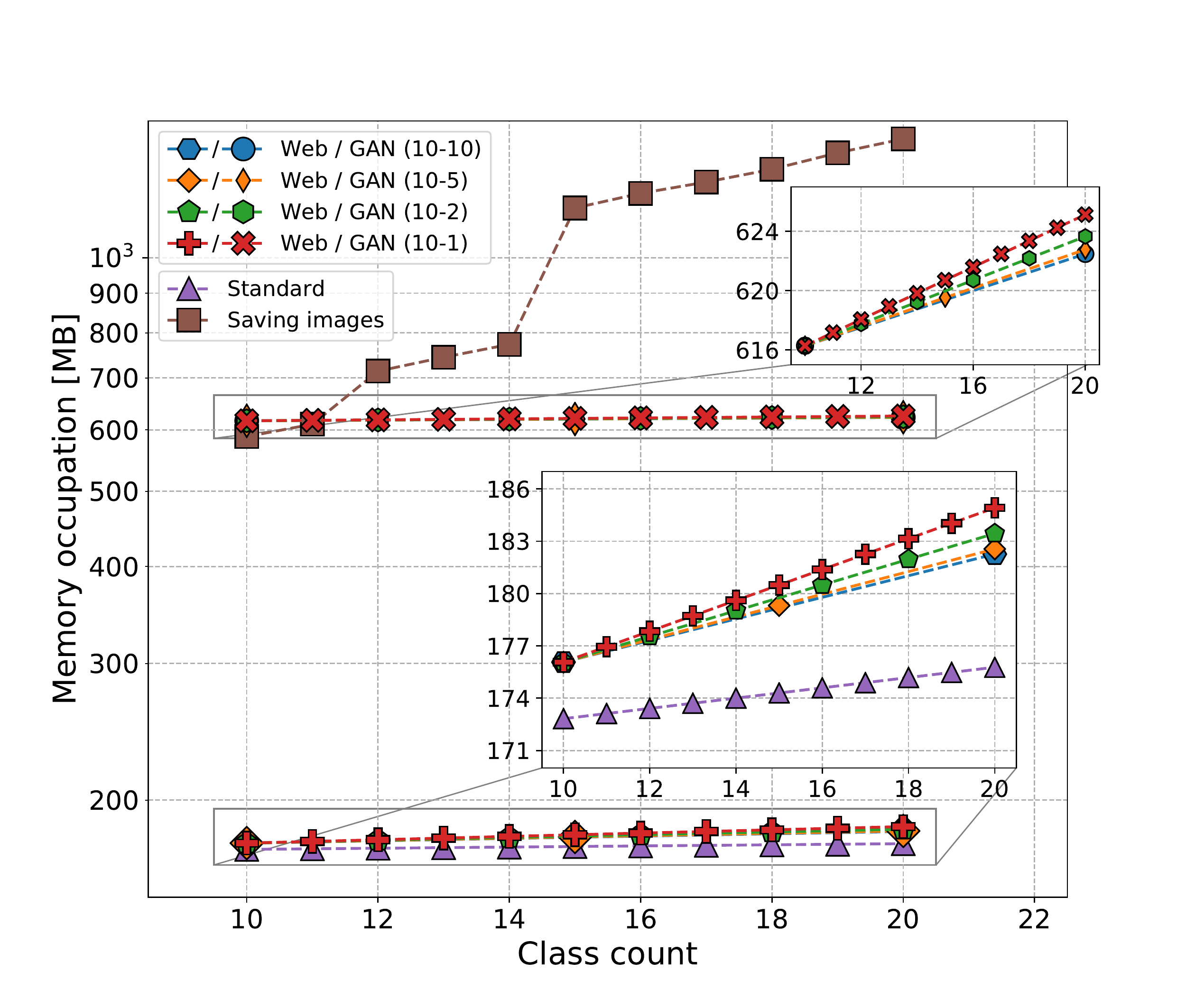}
\vspace{-0.2cm}
\caption{Memory occupation in the disjoint scenario.} %
\label{fig:memory}
\vspace{-0.1cm}
\end{figure}

In this section we present the experimental evaluation on the Pascal VOC 2012 dataset \cite{everingham2012pascal}. 
Following previous works on this topic \cite{shmelkov2017incremental,michieli2019incremental,michieli2020knowledge,cermelli2020modeling}, we start by analyzing the performance on three widely used incremental scenarios: \ie, addition of the last class (19-1), addition of the last 5 classes at once (15-5) and addition of the last 5 classes sequentially (15-1). 
Moreover, we report the performance on three more challenging scenarios in which 10 classes are added sequentially one by one (10-1), in 2 batches of 5 elements each (10-5) and all at once (10-10).
Classes for the incremental steps are selected according to the alphabetical order. 
We compare with the na\"ive fine-tuning approach (FT), which defines the lower limit to the accuracy of an incremental model, and with the joint training on the complete dataset in one step, which serves as upper bound. \revised{We also report the results %
of a simple Store and Replay (S\&R) method, where at each incremental step we store a certain number of true samples for newly added classes, such that the respective size  in average matches the size of the helper decoders needed by RECALL %
(see Fig.~\ref{fig:memory})}.
As comparison, we include $2$ methods extended from classification (\ie, LwF \cite{li2018learning} and its single-headed version LwF-MC \cite{rebuffi2017icarl}) %
and the most relevant methods 
designed for continual segmentation (\ie, ILT \cite{michieli2019incremental}, CIL \cite{klingner2020class}, MiB \cite{cermelli2020modeling} and SDR \cite{michieli2021continual}). 
Exhaustive quantitative results in terms of mIoU %
are shown in Table \ref{tab:voc_1}. 
\revisedcr{For each setup we report the mean accuracy for the initial set of classes, for the classes in the incremental steps and for all classes, computed after the overall training.}

\noindent
\textbf{Addition of the last class.} 
First, we train over the first 19 classes during step 0. 
Then, we perform a single incremental step %
to learn  \textit{tv/monitor}. 
Looking at %
Table~\ref{tab:voc_1} (upper-left section), we notice that FT results in a drastic performance degradation w.r.t.\ joint training, due to catastrophic forgetting. RECALL, instead, shows higher overall mIoU than competitors and it is especially effective on the last class, 
whilst still retaining high accuracy on the past ones
thanks to the regularization brought in by background inpainting and replay strategies. 
S\&R, instead, heavily forgets previous classes, thus confirming the usefulness of replay data.

\begin{table*}[h]
\definecolor{forestgreen}{rgb}{0.13, 0.55, 0.13}
\centering
\footnotesize
\setlength{\tabcolsep}{3.8pt}
\renewcommand{\arraystretch}{0.61}

\caption{mIoU on Pascal VOC2012 for different incremental setups. Results of competitors in the upper part come from \cite{michieli2021continual,cermelli2020modeling}, while we run their implementations for the new scenarios in the bottom part.}
\vspace{-0.2cm}%
\begin{tabular}{l V{3} cc|c | cc|c V{3} cc|c | cc|c V{3} cc|c | cc|c}
\multicolumn{1}{c}{} & \multicolumn{6}{c}{\textbf{19-1}} & \multicolumn{6}{c}{\textbf{15-5}} & \multicolumn{6}{c}{\textbf{15-1}} \\
& \multicolumn{3}{c|}{\textbf{Disjoint}} & \multicolumn{3}{c V{3}}{\textbf{Overlapped}}   
& \multicolumn{3}{c|}{\textbf{Disjoint}} & \multicolumn{3}{c V{3}}{\textbf{Overlapped}} 
& \multicolumn{3}{c|}{\textbf{Disjoint}} & \multicolumn{3}{c}{\textbf{Overlapped}} \\[0.5ex]
\textbf{Method} & 1-19 & \phantom{6}20\phantom{6} & all         & 1-19 & \phantom{6}20\phantom{6} & all  
                & 1-15 & 16-20 & all      & 1-15 & 16-20 & all 
                & 1-15 & 16-20 & all      & 1-15 & 16-20 & all    \\
\midrule
\revised{FT}   & 35.2 & 13.2 & 34.2  & 34.7 & 14.9 & 33.8    
     & 8.4 & 33.5 & 14.4   & 12.5 & 36.9 & 18.3   
     & 5.8 & 4.9 & 5.6     & 4.9 & 3.2 & 4.5  \\
S\&R  & 55.3 & 43.2 & 56.2    & 54.0 & 48.0 & 55.1
      & 38.5 & 43.1 & 41.6    & 36.3 & 44.2 & 40.3
      & 41.0 & 31.8 & 40.7    & 38.6 & 31.2 & 38.9\\
\midrule

\revised{LwF \cite{li2018learning}} & 65.8 & 28.3 & 64.0     & 62.6 & 23.4 & 60.8   
                                    & 39.7 & 33.3 & 38.2     & 67.0 & 41.8 & 61.0   
                                    & 26.2 & 15.1 & 23.6     & 24.0 & 15.0 & 21.9   \\

\revised{LwF-MC \cite{rebuffi2017icarl}} & 38.5 & 1.0 & 36.7     & 37.1 & 2.3 & 35.4  
                                         & 41.5 & 25.4 & 37.6    & 59.8 & 22.6 & 51.0  
                                         & 6.9 & 2.1 & 5.7       & 6.9 & 2.3 & 5.8   \\

\revised{ILT \cite{michieli2019incremental}} & 66.9 & 23.4 & 64.8    & 50.2 & 29.2 & 49.2 
                                             & 31.5 & 25.1 & 30.0    & 69.0 & 46.4 & 63.6  
                                             & 6.7 & 1.2 & 5.4       & 5.7 & 1.0 & 4.6  \\
\revised{CIL \cite{klingner2020class}} & 62.6 & 18.1 & 60.5    & 35.1 & 13.8 & 34.0   
                                       & 42.6 & 35.0 & 40.8    & 14.9 & 37.3 & 20.2   
                                       & 33,3 & 15.9 & 29.1    & 6.3 & 4.5 & 5.9   \\

MiB \cite{cermelli2020modeling} & 69.6 & 25.6 & 67.4   & \textbf{70.2} & 22.1 & 67.8 
                                          & 71.8 & 43.3 & 64.7   & \textbf{75.5} & 49.4 & 69.0 
                                          & 46.2 & 12.9 & 37.9   & 35.1 & 13.5 & 29.7   \\                      

\revised{SDR \cite{michieli2021continual}} & \textbf{69.9} & 37.3 & \textbf{68.4} & 69.1 & 32.6 & 67.4  
                                           & \textbf{73.5} & 47.3 & \textbf{67.2} & 75.4 & 52.6 & \textbf{69.9}  
                                           & 59.2 & 12.9 & 48.1 & 44.7 & 21.8 & 39.2   \\                          

\midrule 

RECALL (GAN) & 65.2 & \textbf{50.1} & 65.8    & 67.9 & 53.5 & 68.4   
           & 66.3 & 49.8 & 63.5    & 66.6 & 50.9 & 64.0   
           & 66.0 & 44.9 & 62.1    & 65.7 & 47.8 & 62.7 \\ 
       
RECALL (Web) & 65.0 & 47.1 & 65.4   & 68.1 & \textbf{55.3} & \textbf{68.6} 
           & 69.2 & \textbf{52.9} & 66.3   & 67.7 & \textbf{54.3} & 65.6 
           & \textbf{67.6} & \textbf{49.2} & \textbf{64.3}   & \textbf{67.8} & \textbf{50.9} & \textbf{64.8} \\
  
\midrule
Joint   & 75.5 & 73.5 & 75.4    & 75.5 & 73.5 & 75.4    
        & 77.5 & 68.5 & 75.4    & 77.5 & 68.5 & 75.4    
        & 77.5 & 68.5 & 75.4    & 77.5 & 68.5 & 75.4\\

\multicolumn{10}{c}{}\\[-1ex]
\multicolumn{1}{c}{} & \multicolumn{6}{c}{\textbf{10-10}} & \multicolumn{6}{c}{\textbf{10-5}} & \multicolumn{6}{c}{\textbf{10-1}} \\
& \multicolumn{3}{c|}{\textbf{Disjoint}} & \multicolumn{3}{c V{3}}{\textbf{Overlapped}}   
& \multicolumn{3}{c|}{\textbf{Disjoint}} & \multicolumn{3}{c V{3}}{\textbf{Overlapped}} 
& \multicolumn{3}{c|}{\textbf{Disjoint}} & \multicolumn{3}{c}{\textbf{Overlapped}} \\[0.5ex]
\textbf{Method} & 1-10 & 11-20 & all      & 1-10 & 11-20 & all  
                & 1-10 & 11-20 & all      & 1-10 & 11-20 & all 
                & 1-10 & 11-20 & all      & 1-10 & 11-20 & all    \\

\midrule
\revised{FT} & 7.7 & 60.8 & 33.0   & 7.8 & 58.9 & 32.1 
   & 7.2 & 41.9 & 23.7   & 7.4 & 37.5 & 21.7 
   & 6.3 & 2.0 & 4.3     & 6.3 & 2.8 & 4.7 \\

S\&R   & 25.1 & 53.9 & 41.7   & 18.4 & 53.3 & 38.2  
       & 26.0 & 28.5 & 29.7   & 22.2 & 28.5 & 27.9
       & 30.2 & 19.3 & 27.3   & 28.3 & 20.8 & 27.1\\
\midrule 

\revised{LwF \cite{li2018learning}}  & 63.1 & 61.1 & 62.2   & \textbf{70.7} & 63.4 & \textbf{67.2} 
                           & 52.7 & 47.9 & 50.4   & 55.5 & 47.6 & 51.7 
                           & 6.7 & 6.5 & 6.6      & 16.6 & 14.9 & 15.8 \\

\revised{LwF-MC \cite{rebuffi2017icarl}} & 52.4 & 42.5 & 47.7   & 53.9 & 43.0 & 48.7 
                               & 44.6 & 43.0 & 43.8   & 44.3 & 42.0 & 43.2 
                               & 6.9 & 1.7 & 4.4      & 11.2 & 2.5 & 7.1 \\

\revised{ILT \cite{michieli2019incremental}} & \textbf{67.7} & \textbf{61.3} & \textbf{64.7}   & 70.3 & 61.9 & 66.3 
                                   & 53.4 & 48.1 & 50.9   & 55.0 & 44.8 & 51.7 
                                   & 14.1 & 0.6 & 7.5     & 16.5 & 1.0 & 9.1 \\

\revised{CIL \cite{klingner2020class}} & 37.4 & 60.6 & 48.4   & 38.4 & 60.0 & 48.7 
                             & 27.5 & 41.4 & 34.1   & 28.8 & 41.7 & 34.9 
                             & 7.1 & 2.4 & 4.9      & 6.3 & 0.8 & 3.6 \\

\revised{MiB \cite{cermelli2020modeling}} & 66.9 & 57.5 & 62.4    & 70.4 & 63.7 & 67.2 
                                          & 54.3 & 47.6 & 51.1    & 55.2 & 49.9 & 52.7 
                                          & 14.9 & 9.5 & 12.3     & 15.1 & 14.8 & 15.0 \\
                                
\revised{SDR \cite{michieli2021continual}} & 67.5 & 57.9 & 62.9   & 70.5 & \textbf{63.9} & \textbf{67.4}  
                                           & 55.5 & 48.2 & 52.0   & 56.9 & 51.3 & 54.2  
                                           & 25.5 & 15.7 & 20.8   & 26.3 & 19.7 & 23.2   \\

\midrule

RECALL (GAN) & 62.6 & 56.1 & 60.8    & 65.0 & 58.4 & 63.1
           & 60.0 & 52.5 & 57.8    & 60.8 & 52.9 & 58.4 
           & 58.3 & 46.0 & 53.9    & 59.5 & 46.7 & 54.8  \\

RECALL (Web) & 64.1 & 56.9 & 61.9    & 66.0 & 58.8 & 63.7
           & \textbf{63.2} & \textbf{55.1} & \textbf{60.6}    & \textbf{64.8} & \textbf{57.0} & \textbf{62.3}  
           & \textbf{62.3} & \textbf{50.0} & \textbf{57.8}    & \textbf{65.0} & \textbf{53.7} & \textbf{60.7}  \\
\midrule 

Joint   & 76.6 & 74.0 & 75.4    & 76.6 & 74.0 & 75.4    
        & 76.6 & 74.0 & 75.4    & 76.6 & 74.0 & 75.4    
        & 76.6 & 74.0 & 75.4    & 76.6 & 74.0 & 75.4\\

\end{tabular}

\label{tab:voc_1}
\end{table*}%

\noindent
\textbf{Addition of last 5 classes.}
In this setup, 15 classes are learned in the initial step, while the remaining 5 are added in one shot (15-5) or sequentially one at a time (15-1). 
Compared to the 19-1 setup, the addition of multiple classes in the incremental iterations makes  catastrophic forgetting even more severe. %
\revised{The accuracy gap between FT and joint training, in fact, raises from about  $41\%$ of the 19-1 case to more than $70\%$ of mIoU in the 15-1 scenario.} 
Taking a closer look at the results in Table~\ref{tab:voc_1} (upper mid and right sections), our replay approaches strongly limit the degradation caused by catastrophic forgetting. 
This trend can be observed in  the 15-5 setup and more evidently in the 15-1 one, %
both in the disjoint and overlapped settings: 
exploiting \revised{generated or web-derived} replay samples proves to effectively restore knowledge of past classes, leading to a final mIoU approaching that of the joint training. 
Storing and replaying \revised{original} samples, instead, improves the performance w.r.t.\ FT, %
but ultimately leads to a mIoU lower of more than $20\%$ if compared to our approaches.
This is due to the limited number of samples %
to be stored in order to match the helper decoder size: %
their sole addition is, in fact, insufficient to adequately preserve learned knowledge.
Finally, we observe that RECALL can scale much better than competitors when multiple incremental steps are performed (scenario 15-1), as typically encountered in real-world applications.

\begin{figure}[tbp]
\centering
\includegraphics[trim=0.2cm 0.25cm 0.2cm 0.2cm, clip, width=\linewidth]{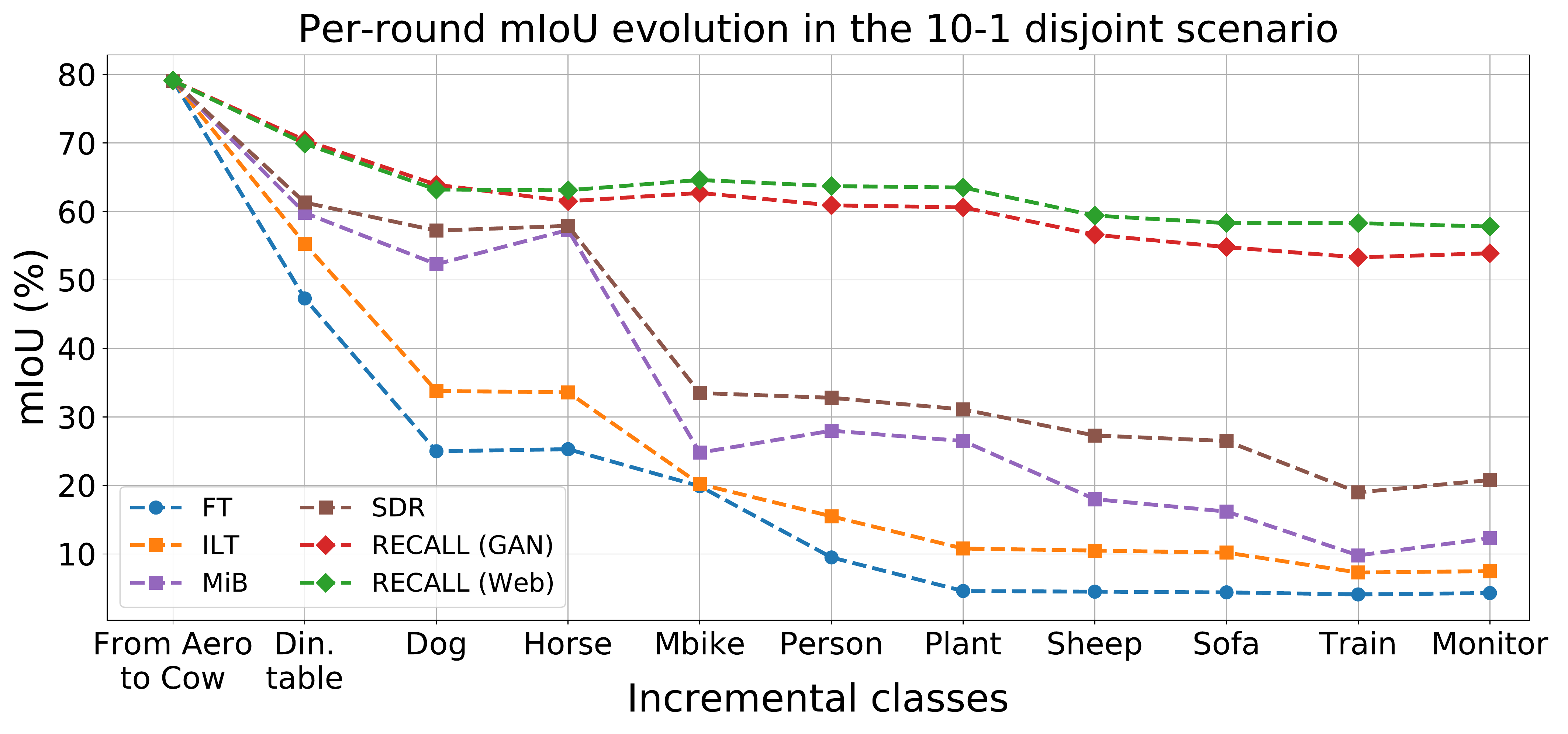}
\vspace{-0.45cm}
\caption{\revisedcr{Evolution of mIoU on the 10 tasks of 10-1 disjoint.}} %
\vspace{-0.3cm}
\label{fig:perround_mIoU}
\end{figure}

\noindent
\textbf{Addition of last 10 classes.}
To analyze the previous claim, we introduce some new challenging experiments, not evaluated in previous works. In these tests only 10 classes are observed in the initial step, while the remaining ones are added in a single batch (10-10), in 2 steps of 5 classes each (10-5), or individually (10-1). 
Again, FT is heavily affected by the information loss that occurs when performing incremental training without regularization, leading to  performance drops up to about $71\%$ of mIoU w.r.t.\ the joint training in the most challenging 10-1 setting. 
Thanks to the introduction of replay data, RECALL brings a remarkable performance boost to the segmentation accuracy and
becomes more and more valuable as the difficulty of the settings increases. In the 10-10 case, our method achieves slightly lower mIoU results than competitors (although comparable).
As we increase complexity, our approach is able to outperform competitors by about $8\%$ of mIoU in 10-5 and by $37\%$ of mIoU in 10-1. %
We remark how RECALL shows a convincing capability of %
providing a rather steady accuracy in different setups, regardless of the number of incremental steps used to introduce new classes.
For example, in the disjoint scenario, when moving from simpler to more challenging setups (\ie, from 10-10 to 10-1, passing through 10-5), the mIoU of FT drops as $33.0\%\!\rightarrow\!23.7\%\!\rightarrow\!4.3\%$ 
and the one of SDR (\ie, the best compared approach) as $62.9\%\!\rightarrow\!52.0\%\!\rightarrow\!20.8\%$,  while our approach maintains a stable mIoU trend of $61.9\%\!\rightarrow\!60.6\%\!\rightarrow\!57.8\%$. Finally, we report the mIoU after each incremental step on the 10-1 disjoint scenario in Fig.~\ref{fig:perround_mIoU}, where our approaches show much higher mIoU at every learning step, indicating improved resilience to forgetting and background shift, than competitors.

\newcommand{\imgsize}{20.5mm}
\begin{figure*}[htbp]
\centering

\setlength{\tabcolsep}{0.7pt} %
\renewcommand{\arraystretch}{0.6}
\centering
\footnotesize
\begin{subfigure}[htbp]{2\textwidth}
\begin{tabular}{c<{\hspace{1mm}}cccccccc}
  
   \rotatebox{90}{\hspace{+4ex}15-1} &  
   \includegraphics[width=\imgsize]{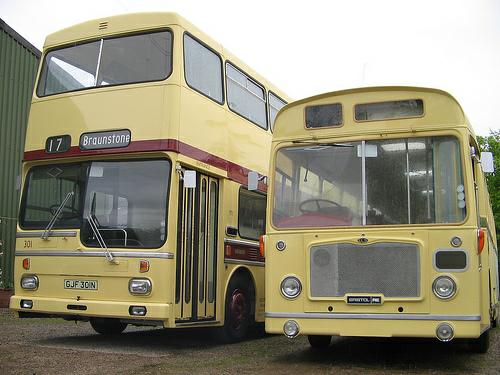} &
   \includegraphics[width=\imgsize]{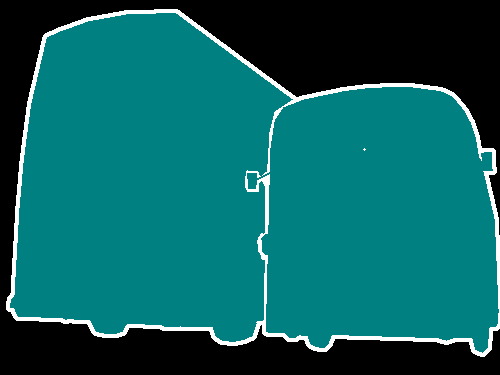} & 
   \includegraphics[width=\imgsize]{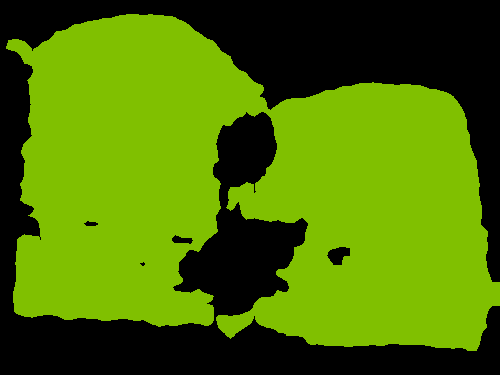} & 
   \includegraphics[width=\imgsize]{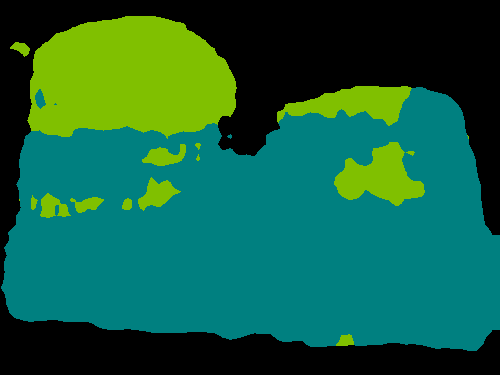} & 
   \includegraphics[width=\imgsize]{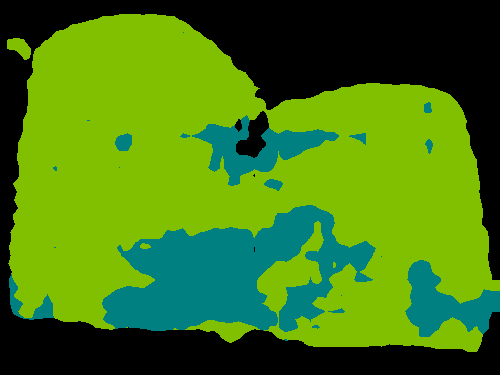} &
   \includegraphics[width=\imgsize]{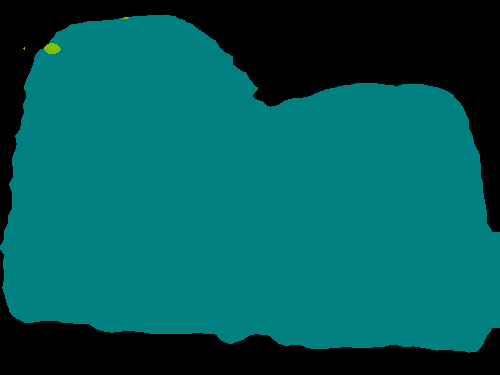} & 
   \includegraphics[width=\imgsize]{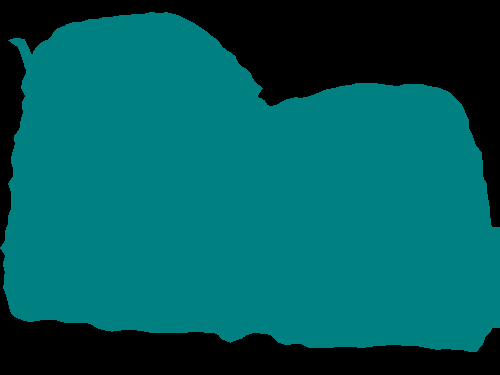} &
   \includegraphics[width=\imgsize]{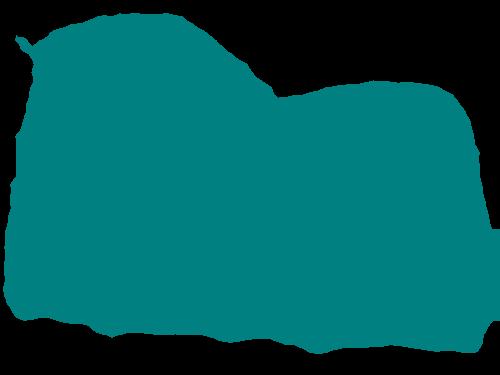} \\ 
   
    \rotatebox{90}{\hspace{+4ex}15-5} & 
    \includegraphics[width=\imgsize]{img/T3/0026.jpg} &
   \includegraphics[width=\imgsize]{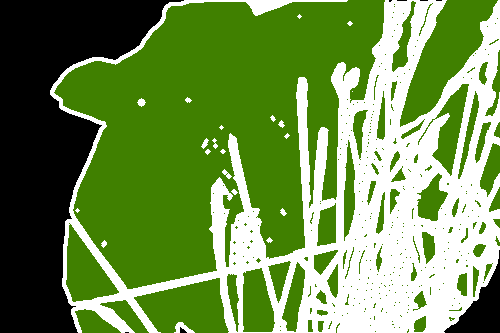} & 
   \includegraphics[width=\imgsize]{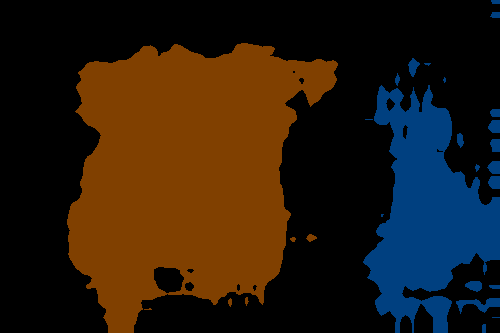} & 
   \includegraphics[width=\imgsize]{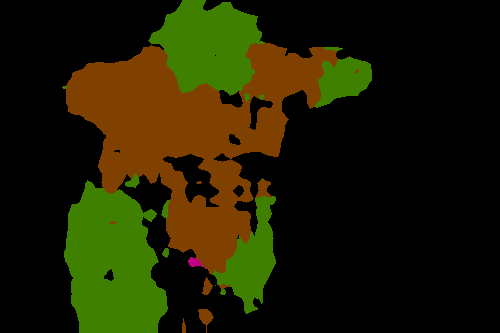} &
   \includegraphics[width=\imgsize]{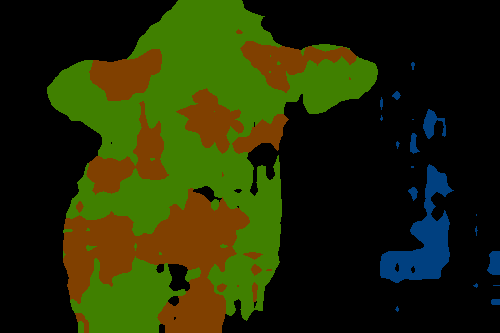} &
   \includegraphics[width=\imgsize]{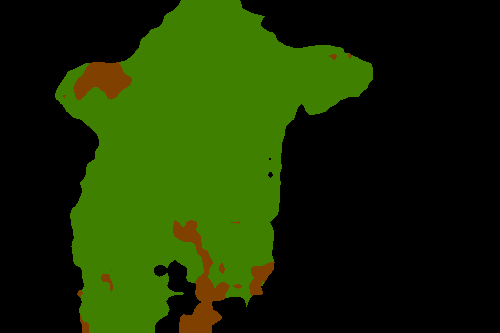} & 
   \includegraphics[width=\imgsize]{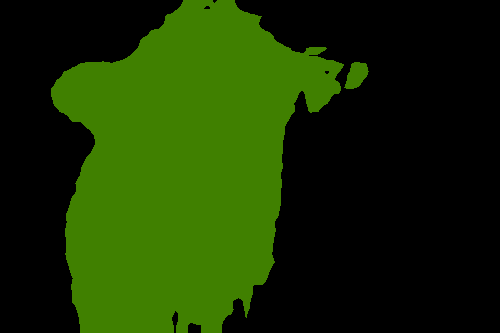} &
   \includegraphics[width=\imgsize]{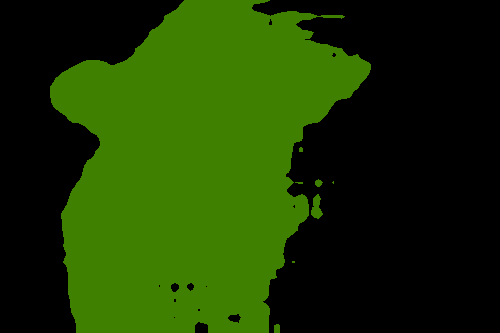} \\ 
   
   \rotatebox{90}{\hspace{+4ex}10-1} & 
    \includegraphics[width=\imgsize]{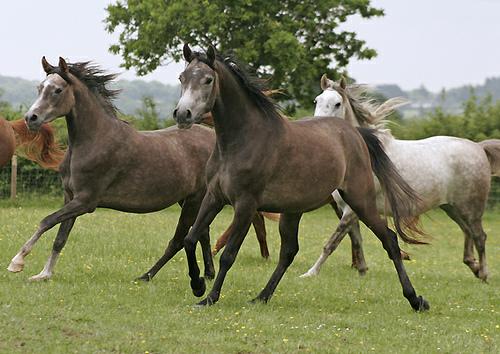} &
   \includegraphics[width=\imgsize]{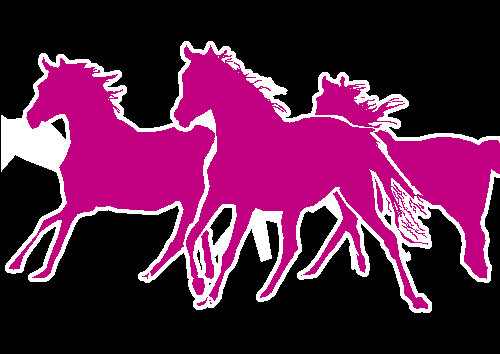} & 
   \includegraphics[width=\imgsize]{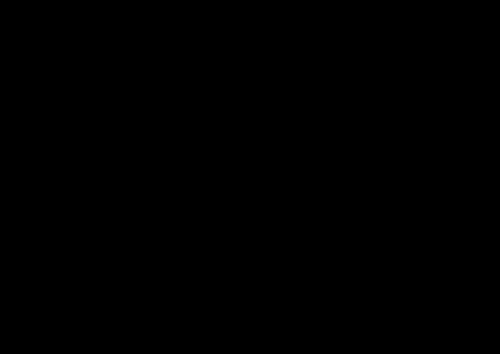} & 
   \includegraphics[width=\imgsize]{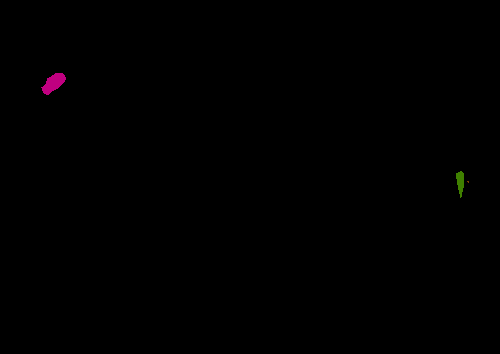} & 
   \includegraphics[width=\imgsize]{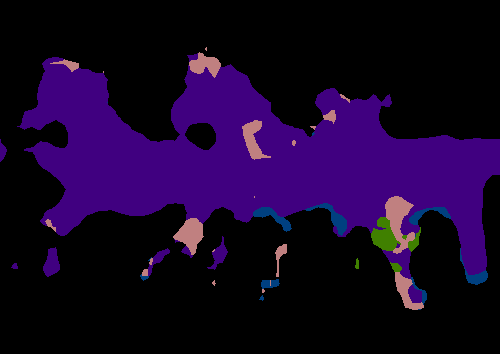} &
   \includegraphics[width=\imgsize]{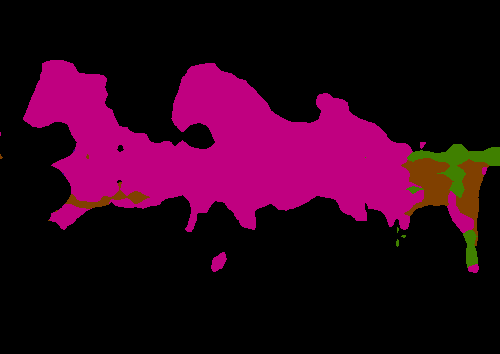} & 
   \includegraphics[width=\imgsize]{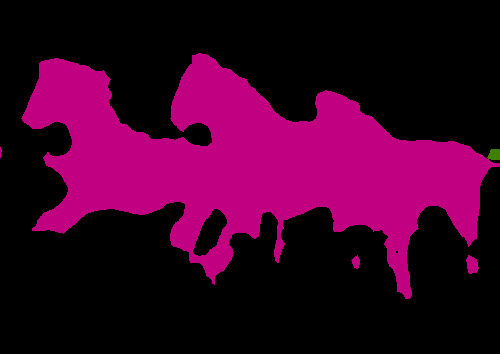} &
   \includegraphics[width=\imgsize]{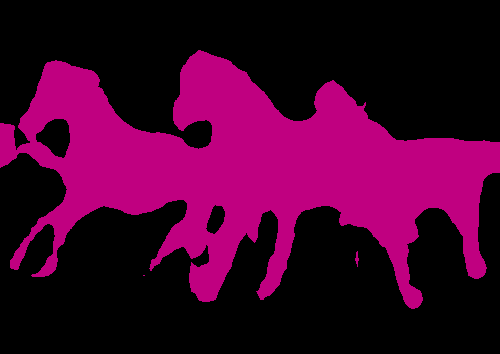} \\

  &  RGB  &  GT  & FT & S\&R & Inpainting  &  RECALL (GAN)  & RECALL (Web) &  Joint
  
\end{tabular}
\end{subfigure}
\vspace{-0.2cm}
\caption{Qualitative results on disjoint incremental setups: from top to bottom 15-1, 15-5 and 10-1.
(\textit{best viewed in colors}).}
\label{fig:qual_res}
\end{figure*}

\revisedcr{In the qualitative results in Fig.~\ref{fig:qual_res} we observe that RECALL effectively alleviates forgetting and reduces the bias towards novel classes.}
\revisedcr{In the first row, the \textit{bus} is correctly preserved while FT, S\&R and inpainting wrongly classify it as \textit{train} (\ie, one of the novel classes); in the second row, FT places \textit{sheep} and \textit{tv} (newly added classes) in place of \textit{cow}; in the third row, some \textit{horse}'s features are either mixed with those of \textit{person} and \textit{cat} or completely destroyed, while they are preserved by our methods (the web scheme shows higher accuracy than the GAN here).}
\revisedcr{Additional results, a few tests combining RECALL with competitors and a preliminary evaluation on ADE20K \cite{zhou2017scene} are presented in Suppl.\ Mat.}

\begin{figure}[htbp]
\centering
\includegraphics[trim=0.2cm 0.25cm 0.2cm 0.2cm, clip, width=0.82\linewidth]{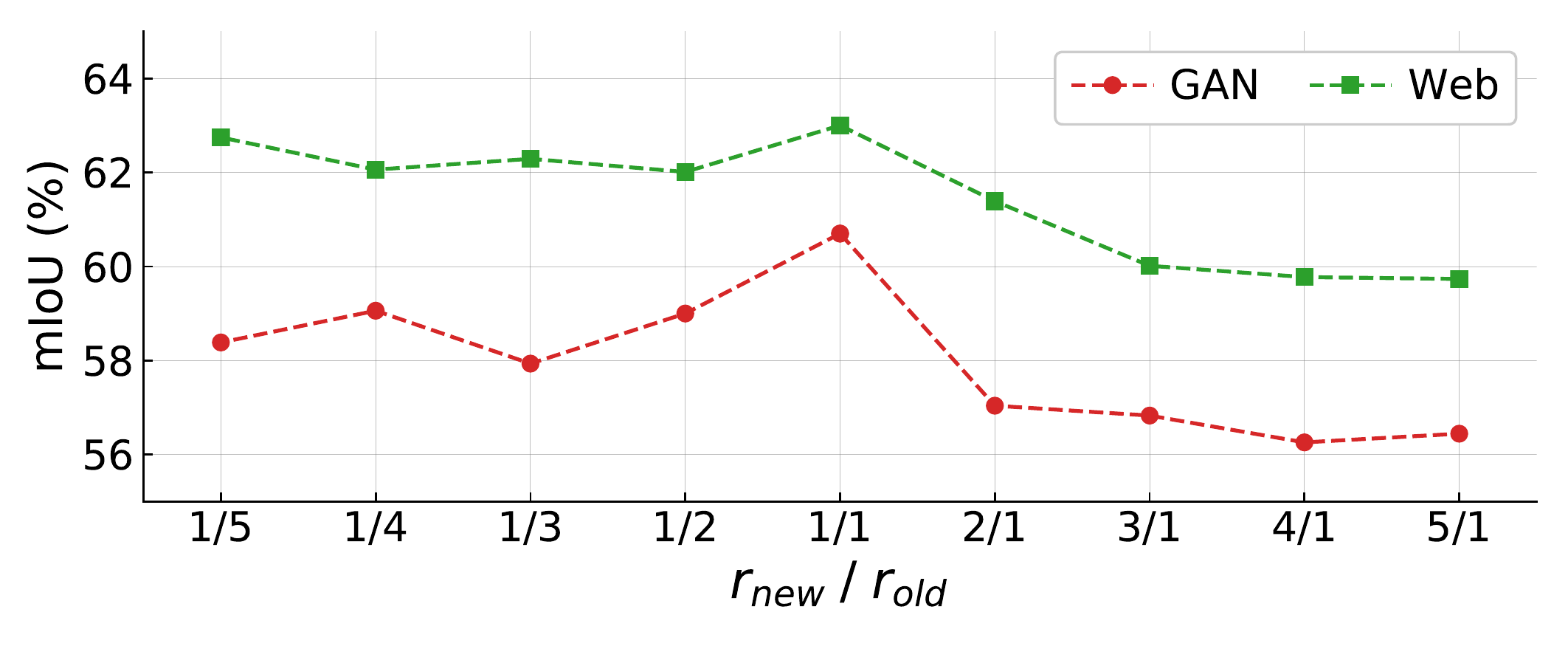}
\vspace{-0.25cm}
\caption{Distinct interleaving policies in 15-1 disjoint.} %
\label{fig:interleave}
\vspace{-0.2cm}
\end{figure}

\subsection{Ablation Study}
\label{subsec:ablation}

To further validate the robustness of our approach, we perform some ablation studies.
First of all, we analyze the memory requirements. %
The plot in Fig.~\ref{fig:memory} shows in semi-log scale the memory occupation (expressed in MB) of the data to be stored at the end of each incremental step, as a function of  the number of classes learned up to that point. 
We denote with \textit{standard} an incremental approach which do not store any sample (\eg, FT, LwF, ILT, MiB, SDR). %
The saved model generally corresponds to a fixed size encoder and a decoder, whose dimension slightly increases at each step to account for additional output channels for the new learnable classes. \textit{Saving images}, instead, refers to the extreme scenario where training images of past steps are stored, thus being available throughout the entire incremental process. 
As concerns our approach, to annotate originally weakly-labeled replay images, we devise a specific module (Sec.~\ref{sec:gen_arc}), which  requires to save a set of helper decoders $\{ D^{H}_{\mathcal{C}_{i}} \}_{i=0}^{k}$, one for each past step. %
Finally, for the GAN-based approach we add the storage required for the generative model. %
Fig.~\ref{fig:memory} shows that our web-based solution is very close to the \textit{standard} ones in terms of memory occupation. 
The space required to store the GAN is comparable to that needed to save images in the very initial steps, but then remains constant %
while the space for saving all training data quickly grows. %

\revised{
We further analyze the contribution of the background inpainting and replay techniques in %
Table \ref{tab:abl}. %
While  inpainting alone provides a solid contribution in terms of knowledge preservation acting similarly to knowledge distillation, we observe that its effect tends to attenuate with multiple incremental steps.
For example, moving from 10-10 to 10-1 overlapped setups, the mIoU drops more than 20\%. 
On the other hand, the proposed replay techniques prove to be beneficial when multiple training stages are involved. %
On the same setting, replay techniques alone limit the degradation to only 8\%. 
Yet, jointly employing replay and inpainting further boosts the final results in all setups (up to $15\%$), %
proving that they can be effectively combined.
}

\begin{table}[tbp]
\vspace{-0.15cm}
\caption{mIoU results showing the contribution of each module, D: Disjoint, O: Overlapped.}
\vspace{-0.2cm}
\centering
\fontsize{8}{10}\selectfont
\setlength{\tabcolsep}{1.3pt}
\renewcommand{\arraystretch}{1}
\begin{tabular}{c|cc|cc|cc|cc|cc|cc}
\multicolumn{1}{c}{} & \multicolumn{2}{c}{19-1} & \multicolumn{2}{c}{15-5} & \multicolumn{2}{c}{15-1} 
& \multicolumn{2}{c}{10-10} & \multicolumn{2}{c}{10-5} & \multicolumn{2}{c}{10-1} \\
Method & D & O & D & O & D & O  
       & D & O & D & O & D & O \\
\midrule
Bgr inp.  & 65.6  & 66.7  & 52.2 & 52.5  & 49.7  & 49.9  
            & 58.8   & 60.7   & 47.5   & 47.1   & 34.0   & 39.0 \\
GAN    & 54.5   & 56.2   & 49.8   & 49.1   & 47.9   & 48.2   
            & 45.8   & 48.8   & 38.1   & 43.7   & 36.6   & 40.8 \\
Web    & 57.3   & 57.4   & 55.2   & 54.7  & 55.0   & 53.7   
            & 55.2   & 58.2   & 47.9   & 52.1   & 45.4   & 50.1 \\
\midrule
GAN+inp.   & 65.8 & 68.4 & 63.5   & 64.0 & 62.1 & 62.7 
             & 60.8    & 63.1   & 57.8    & 58.4   & 53.9    & 54.8 \\
Web+inp.   & 65.4 & 68.6 & 66.3   & 65.6 & 64.3 & 64.8
             & 61.9    & 63.7   & 60.6    & 62.3   & 57.8    & 60.7 \\
\end{tabular}
\label{tab:abl}
\vspace{-0.29cm}
\end{table}

Finally, we analyze how results vary w.r.t.\ the proportion of 
new ($r_{new}$) and replay ($r_{old}$) samples seen during training (Fig.~\ref{fig:interleave}): the mIoU is quite stable w.r.t. this ratio, however the maximum value is reached when the same number of old and replay samples is used, \ie, $r_{new}/r_{old}=1$.

\section{Conclusions}
\label{sec:conclusions}

\revisedcr{In this paper we introduced RECALL, which targets continual semantic segmentation by means of replay strategies to alleviate catastrophic forgetting and background inpainting to mitigate background shift. 
Two replay schemes are proposed to retrieve data related to former training stages, either reproducing it via a conditional GAN or crawling it from the web. 
The experimental analyses proved the efficacy of our framework in improving accuracy and robustness to multiple incremental steps compared to competitors. New research will improve the generative model, coupling it more strictly with the incremental setup, and explore how to control and refine weak supervision during web-crawling. Evaluation on different datasets, such as ADE20K, will also be performed.}

\newpage
{\small
\bibliographystyle{ieee_fullname}
\bibliography{strings,refs}
}

\clearpage

\setcounter{equation}{0}
\setcounter{figure}{0}
\setcounter{table}{0}
\setcounter{section}{0}
\setcounter{page}{1}
\renewcommand{\thesection}{S\arabic{section}}
\renewcommand{\theequation}{S\arabic{equation}}
\renewcommand{\thefigure}{S\arabic{figure}}
\renewcommand{\thetable}{S\arabic{table}}

\begin{center}
  \textbf{\large RECALL: Replay-based Continual Learning in Semantic Segmentation \\ \vspace{0.1cm} \textit{Supplementary Material}}\\[.2cm]
\end{center}

In this document, we present some additional material to better motivate the design choices behind our method, RECALL, along with some \revisedcr{additional} experiments. More in detail, we start by discussing the impact of the pre-training dataset used for the initialization of the ResNet101 backbone on the performance of continual semantic segmentation algorithms. Then, we further comment on the Class Mapping Module needed to perform the conversion from the class space of the GAN to the class space of the considered segmentation dataset. Then, we report some additional insights on the experimental results 
\revisedcr{on the Pascal VOC2012 benchmark and include some preliminary analyses on the ADE20K dataset.} Finally, some preliminary results combining RECALL with competitors are reported.

\section{Analyses on Pre-Training}

The weights of semantic segmentation deep learning architectures are typically initialized with pre-trained values computed on a large dataset for a related (but usually different) task. The most common pre-training strategy consists in using weights computed on image classification large-scale datasets, such as ImageNet \cite{deng2009imagenet}. On the other hand, it has been shown \cite{michieli2020knowledge,chen2018deeplab} that pre-training weights on a related segmentation dataset, such as MS COCO \cite{lin2014microsoft}, could further boost  results on semantic segmentation benchmarks. However, using another semantic segmentation dataset raises some concerns about the fact that the pre-training data could contain information about the tasks to be learned in the incremental steps, thus following previous works \cite{michieli2019incremental,michieli2021continual,cermelli2020modeling} we decided for a more conservative approach in the main paper using ImageNet pre-training.

To further investigate this aspect we show extensive results in %
Table~\ref{tab:voc_IN} comparing ImageNet and MS COCO pre-training strategies for all the considered incremental setups. Here we can see that, as expected, pre-training on MS COCO always helps incremental semantic segmentation. We argue that the motivation is at least two-fold: first, a better initialized model on the same target task could converge to a better solution, and second, a model pre-trained on MS COCO could have already learned some spatial and semantic information of classes added to the model in incremental steps (this second point is the reason why we decided to avoid using this strategy even if it leads to better results).
More in detail, we can observe that %
incremental approaches show significant improvements of up to $15\%$.
This quite large gap could be due also to the encoder freezing procedure we employ in our our work (similarly to \cite{michieli2019incremental,michieli2020knowledge}) that reduces catastrophic forgetting, but at the same time does not allow the network to update the feature extraction module according to the information in the samples of the new classes at each incremental step, which can be used only to update the decoder. Indeed, a model initialized with pre-trained weights on MS COCO (which has enough variability and semantic content information) needs less training steps to adapt to a new (related) semantic segmentation dataset (\ie, different domain but same task). On the other side, pre-training on a different task and different domain (\eg, on ImageNet) requires more training steps to adapt to the new scenario. We can verify this claim looking at Table~\ref{tab:voc_IN}, where we can observe how the mIoU gap between the two pre-training strategies is %
larger when the initial step has fewer classes (\eg, 10).

\section{Class Mapping Module}
\label{sec:cmm}

Here we provide some further analysis and insights on the Class Mapping Module (introduced in Section 5 of the main paper), which is used to translate each class of the semantic segmentation incremental dataset (\eg, Pascal VOC2012 \cite{everingham2012pascal}) to the most similar class of the GAN’s training dataset (\eg, ImageNet \cite{deng2009imagenet}). 
Notice that properly mapping the labels between the different domains is an important step, since incorrect pairings may easily harm the accuracy of the final model.

To solve the task, we took an Image Classifier $I$ pre-trained to address an image classification task on the GAN’s dataset. Then, for each class $c$ in the current label set we select the corresponding training subset (\ie, all the samples of the current training set associated to class $c$), and we sum the resulting class probability vectors from the classification output  (according to $I$). An $\argmax$ operation is then performed, to identify the GAN’s class $c_{G}$ with the highest probability score.
To show the effectiveness of the proposed classification, we report in Table~\ref{tab:CMM} the $3$ classes from the GAN's dataset with the highest score 
for each class of the  Pascal VOC2012 dataset. 
We can see that for all the classes the top selected pairings appear reasonable at first (notice that only the best matching class is selected in the proposed approach). 
At a closer look, we find that the classifier selects an unexpected label only in a single case, 
that is the \textit{person} class being  translated into \textit{cowboy hat}; however, we remark that the ImageNet dataset does not contain the \textit{person} class, thus inherently lacking a close match for that category. %
In light of this, we believe that the chosen class (\ie, \textit{cowboy hat}) is a reasonable choice and may still help in retaining high accuracy on the \textit{person} class, being the \textit{cowboy hat} always shown on top of people's heads. %
This situation is %
interesting as it shows the robustness of  our approach not only to different domains with different statistical distributions (ImageNet domain versus Pascal VOC2012 one), but also to different labeling domains (the label set of VOC2012 is not a subset of the ImageNet one).

The effectiveness of the mapping can also be visually appreciated from the sample generated images shown in Figure \ref{fig:generated} of Section \ref{sec:visual}. In particular, notice how even for the \textit{person} class mapped to the \textit{cowboy hat} the images look reasonable, even if the variability in this case is much smaller if compared to the original VOC class data.

\begin{table*}[htbp]
\centering
\caption{%
Mean IoU achieved by the proposed approach on the Pascal-VOC 2012 dataset for different incremental setups and pre-training strategies.}
\label{tab:voc_IN}
\footnotesize
\setlength{\tabcolsep}{2.5pt}
\renewcommand{\arraystretch}{1.2}
\begin{tabular}{c c V{3} cc|c|cc|c  
                    V{3} cc|c|cc|c
                    V{3} cc|c|cc|c}

\multicolumn{1}{c}{} & \multicolumn{1}{c}{} & \multicolumn{6}{c}{\textbf{19-1}} & \multicolumn{6}{c}{\textbf{15-5}} & \multicolumn{6}{c}{\textbf{15-1}} \\
& & \multicolumn{3}{c|}{\textbf{Disjoint}} & \multicolumn{3}{c V{3}}{\textbf{Overlapped}}   
& \multicolumn{3}{c|}{\textbf{Disjoint}} & \multicolumn{3}{c V{3}}{\textbf{Overlapped}} 
& \multicolumn{3}{c|}{\textbf{Disjoint}} & \multicolumn{3}{c}{\textbf{Overlapped}} \\
\textbf{Method} & \textbf{Init} & 1-19 & 20 & all      & 1-19 & 20 & all 
                                & 1-15 & 16-20 & all   & 1-15 & 16-20 & all 
                                & 1-15 & 16-20 & all   & 1-15 & 16-20 & all  \\

\hline

\multirow{2}{*}{\makecell{GAN}} 
& \revised{ImageNet}
& 65.2 & 50.1 & 65.8    & 67.9 & 53.5 & 68.4
& 66.3 & 49.8 & 63.5    & 66.6 & 50.9 & 64.0
& 66.0 & 44.9 & 62.1    & 65.7 & 47.8 & 62.7 \\
& \revised{MSCOCO}
& 68.7 & 58.4 & 69.3    & 68.6 & 59.6 & 69.3
& 70.3 & 58.4 & 68.5    & 70.3 & 59.5 & 68.7
& 70.4 & 55.5 & 67.8    & 70.8 & 57.5 & 68.6 \\

\hline

\multirow{2}{*}{\makecell{Web}} 
& \revised{ImageNet}
& 65.0 & 47.1 & 65.4   & 68.1 & 55.3 & 68.6 
& 69.2 & 52.9 & 66.3   & 67.7 & 54.3 & 65.6 
& 67.6 & 49.2 & 64.3   & 67.8 & 50.9 & 64.9 \\
& \revised{MSCOCO} 
& 69.7 & 55.3 & 70.1   & 68.8 & 60.7 & 69.5
& 70.7 & 59.2 & 69.0   & 70.7 & 59.9 & 69.1
& 70.9 & 57.4 & 68.7   & 71.2 & 55.7 & 68.5 \\
\multicolumn{1}{c}{}\\[-1.5ex]
\multicolumn{1}{c}{} & \multicolumn{1}{c}{} & \multicolumn{6}{c}{\textbf{10-10}} & \multicolumn{6}{c}{\textbf{10-5}} & \multicolumn{6}{c}{\textbf{10-1}} \\
& & \multicolumn{3}{c|}{\textbf{Disjoint}} & \multicolumn{3}{c V{3}}{\textbf{Overlapped}}   
& \multicolumn{3}{c|}{\textbf{Disjoint}} & \multicolumn{3}{c V{3}}{\textbf{Overlapped}} 
& \multicolumn{3}{c|}{\textbf{Disjoint}} & \multicolumn{3}{c}{\textbf{Overlapped}} \\
\textbf{Method} & \textbf{Init} & 1-10 & 11-20 & all   & 1-10 & 11-20 & all 
                                        & 1-10 & 11-20 & all   & 1-10 & 11-20 & all 
                                        & 1-10 & 11-20 & all   & 1-10 & 11-20 & all  \\
   
\hline

\multirow{2}{*}{\makecell{GAN}} 
&  \revised{ImageNet}
& 62.6 & 56.1 & 60.8   & 65.0 & 58.4 & 63.1  
& 60.0 & 52.5 & 57.8   & 60.8 & 52.9 & 58.4  
& 58.3 & 46.0 & 53.9   & 59.5 & 46.7 & 54.8  \\
& \revised{MSCOCO} 
& 68.2 & 64.5 & 67.6   & 68.3 & 66.1 & 68.4
& 68.2 & 62.1 & 66.4   & 67.3 & 61.9 & 65.8
& 68.2 & 58.3 & 64.6   & 67.8 & 60.5 & 65.4  \\

\hline

\multirow{2}{*}{\makecell{Web}} 
&  \revised{ImageNet}
& 64.1 & 56.9 & 61.9   & 66.0 & 58.8 & 63.7
& 63.2 & 55.1 & 60.6   & 64.8 & 57.0 & 62.3   
& 62.3 & 50.0 & 57.8   & 65.0 & 53.7 & 60.7   \\
& \revised{MSCOCO} 
& 68.0 & 64.9 & 67.7   & 68.2 & 66.4 & 68.4
& 68.6 & 63.9 & 67.4   & 67.6 & 64.6 & 67.3
& 68.0 & 58.0 & 64.3   & 68.5 & 62.5 & 66.7 \\

\end{tabular}

\end{table*}

\begin{table*}[htbp]
  \centering
\setlength{\tabcolsep}{5.5pt}
  \caption{mIoU on VOC2012 disjoint 15-1 with replay data. G: GAN, F: Flickr. Na\"ive: only decoder of last step is used for pseudo-labeling, ours: our complete approach (RECALL) is used.}
    \begin{tabular}{l|c|c|c|c|c}
          & none  & + na\"ive (G) & + na\"ive (F) & + ours (G) & + ours (F) \\\cline{1-6}
    ILT   & 5.4 &    37.8 (\textcolor{myblue}{+32.4})    &   39.9 (\textcolor{myblue}{+34.5})     & 49.6 (\textcolor{mygreen}{+44.2})  & 51.5 (\textcolor{mygreen}{+46.1}) \\
    MiB   & 37.9 &  49.3  (\textcolor{myblue}{+11.4})    &   50.5 (\textcolor{myblue}{+12.6})     & 63.5 (\textcolor{mygreen}{+25.6})  & 65.7  (\textcolor{mygreen}{+27.8}) \\
    SDR   & 48.1 &   53.1 (\textcolor{myblue}{+05.0})     &  55.8  (\textcolor{myblue}{+07.7})    & 65.5 (\textcolor{mygreen}{+17.4})  & 66.5 (\textcolor{mygreen}{+18.4}) \\
    \end{tabular}%
  \label{tab:rebuttal_results}%
\end{table*}%

\section{Per-Class Quantitative Results}

For a more detailed evaluation, we present the per-class IoU values for some of the proposed approaches and scenarios. %
We considered the following methods in the disjoint scenario on all the experimental protocols: fine-tuning (FT), background inpainting, RECALL (GAN), RECALL (Web) and joint training. The results are summarized in Table~\ref{tab:voc_per_class}.
From here, we can appreciate how fine-tuning always catastrophically forgets previous classes when learning new ones. The simple background inpainting strategy allows to largely alleviate such phenomenon bringing a similar effect to recent knowledge distillation approaches \cite{michieli2020knowledge,cermelli2020modeling}. On top of this, we apply GAN or Web-based replay strategies to regularize training and background content inpainting scheme to reduce bias toward the background. While these strategies are specifically designed to preserve old knowledge, they also allow to achieve large mIoU gains on new classes reducing the false positive rate (\ie, the detection of new classes in locations containing  the old ones).

In order to better understand the effect of our proposed modules, we report in Table~\ref{tab:perround_tables} the Pixel Accuracy (PA) and the IoU for the class being added at each step of the disjoint 10-1 scenario. The results demonstrate that, on the newly introduced class, FT generally achieves a very high PA (top-left) and a per-class IoU (top-right) comparable to the other approaches. 
Yet, FT concurrently shows very low mIoU %
over all classes learned up to the current step (bottom-right), as well as over only previously seen categories (bottom-left). 
All combined, this is indicative of an overestimation of the new class. In other words, FT progressively forgets foregoing semantic information, while predicting more often the newly seen class (which experiences high PA but low IoU, due to many false positive predictions). Our approach, instead, can effectively improve knowledge preservation thanks to replay data and background inpainting,
providing steady mIoU results throughout the incremental steps.  

\section{Additional Qualitative Results}
\label{sec:visual}

We report some additional qualitative results: %
Figure \ref{fig:generated} show some examples of images produced by the GAN replay strategy.
Figure \ref{fig:qual_res} displays the segmentation output in the disjoint scenario for all the experimental incremental training protocols (\ie, FT, background inpainting, RECALL with GAN or Web and joint). In particular, we show the results for a couple of samples in each of the 6 considered setups (\ie, 19-1, 15-5, 15-1, 10-10, 10-5 and 10-1). 
Finally, Figure \ref{fig:incr_res}, instead, shows the evolution of the output maps across the incremental steps in the 15-1 scenario for a couple of sample images.

The sample generated images in Figure \ref{fig:generated} allow to verify the effectiveness of the conditioned image generation strategy. Notice how in most cases the images are very similar to the Pascal VOC ones, even for the \textit{person} class that does not have a direct mapping as discussed in Section \ref{sec:cmm}. 
For the sake of comparison, the figure also reports 2 randomly sampled images for each class taken either from the Pascal VOC2012 dataset (first 2 columns of Figure~\ref{fig:generated}) or from Flickr for the Web approach (last 2 columns of Figure~\ref{fig:generated}).

In Figure \ref{fig:qual_res} it is possible to see that the background inpainting strategy constitutes a clear improvement with respect to the simple fine-tuning approach, allowing to reduce catastrophic forgetting, which is very critical in FT. However, forgetting is still fairly noticeable with the sole inpainting strategy, where the output maps are  quite noisy and relevant parts of the objects get lost in many scenes, typically overestimating the background class. 
However, the addition of replay data in both the GAN and the Web-based solutions proves to be very effective in further reducing the forgetting phenomenon, thus providing a final segmentation performance very close to the joint-training reference except for some details, which are typically close to the boundaries of the objects. Furthermore, our approaches do not mislead previous classes with similar ones introduced in the incremental steps (\eg, FT and inpainting mislead the \textit{cow} with \textit{sheep} in row 3 and the \textit{bus} with \textit{train} in row 4).

The accuracy boost introduced by the proposed replay strategies can be further appreciated in Figure \ref{fig:incr_res}, where  we report the segmentation output computed after each incremental step of the 15-1 disjoint setup for a couple of image samples. 
The improvement can be noted by looking, for example,  at the images on the second and fourth incremental steps, where the new classes \textit{sheep} and \textit{train} are introduced respectively. When FT or background inpainting are adopted, the segmentation network tends to experience a severe forgetting of the old classes \textit{cow} and \textit{bus} (which are mistaken for visually similar novel ones). This behavior is corrected by providing replay training data to the network: %
 both GAN and Web-based strategies are able to preserve an accurate recognition of old classes, even when semantically similar ones are incrementally added.

\section{Combining RECALL with Competitors}

\revisedcr{To the best of our knowledge, no works on continual semantic segmentation using GAN-generated or web-crawled data exist. The aim of our work is to provide a general framework to retrieve and employ unlabeled replay data. In this section, we demonstrate that our framework can be applied on top of competing approaches to improve their performance: some experimental results are shown in Table~\ref{tab:rebuttal_results}. Adding replay data with na\"ive pseudo-labeling (\ie, using the decoder of the previous step) already leads to a performance improvement, but combining our method with previous approaches leads to much higher results with improvements ranging from $17\%$ to $46\%$, proving the effectiveness and general applicability of the modules introduced in RECALL. }

\section{Preliminary Analyses on ADE20K}

\revisedcr{In the main paper we reported the experiments on Pascal VOC2012, which contains object-level classes. However, 
when addressing real-world tasks, the complete understanding of the surroundings is usually required:
for instance, to distinguish a mixture of \textit{stuff} and object-level classes, as in the ADE20K dataset \cite{zhou2017scene}. Indeed, the ADE20k dataset poses a great challenge due to the vast class set, comprising \textit{stuff} categories not present in Pascal VOC2012. We remark that exact correspondence between GAN's conditioning class space and semantic segmentation category set is not required by our replay strategy.  For example, as we have already observed, the \textit{person} class is missing in the ImageNet dataset (used for pre-training the generative model), but images of people can still be retrieved from generated images of semantically related categories, such as \textit{hat} (see Table~\ref{tab:CMM}). Thus, even when the generative model cannot be explicitly conditioned to reproduce some specific segmentation classes (\eg, \textit{stuff} categories), it is possible to retrieve instances of them just relying on semantically correlated categories. This retrieval (\ie, mapping) operation is performed automatically by our approach. This is even more true for the web-replay strategy, where we have complete control over the keywords used for the search. %
}

\revisedcr{Hence, we argue that our approach is suitable for continual semantic segmentation even when non-object categories are present. To make our point, we run preliminary experiments on ADE20k. We consider the $100$-$10$ setting, where $100$ classes are learned in the first step and the others are added in batches of $10$ at each incremental step.  The FT baseline reaches a very low mIoU of $0.8\%$, while our RECALL with GAN-based replay samples improves the score up to $11.4\%$, showing that our methods mitigate catastrophic forgetting even in this challenging setup. To achieve these results, we did not perform any parameter tuning (\ie, we kept the same pre-processing, learning parameters, batch and image sizes used for VOC2012). Further experiments and tuning will allow us to provide a proper comparison with other works on this benchmark.}

\begin{table*}[h]
\centering
\caption{Class mapping between Pascal VOC and ImageNet datasets. The table shows the 3 best matching  ImageNet classes for each Pascal VOC 2012 class.  ($^*$): matching classes for \textit{tv/monitor} are not computed since replay data is not needed.}
\setlength{\tabcolsep}{3.5pt}
\renewcommand{\arraystretch}{1.2}

\setlength\arrayrulewidth{0.6pt}
\rowcolors{2}{white}{gray!25}
\begin{tabular}{cc|ccc}

\rowcolor{gray!50}
 \multicolumn{2}{c|}{\textbf{Pascal}} & \multicolumn{3}{c}{\textbf{ImageNet}} \\
\rowcolor{gray!50}
index & class & 1st class & 2nd class  & 3rd class  \\
\hline
1 & airplane   & airliner       & warplane        & wing  \\
2 & bicycle    & mountain bike    & tandem           & tricycle  \\
3 & bird       & kite (bird)      & dipper          & quail \\
4 & boat       & catamaran        & lakeside        & fireboat  \\
5 & bottle     & beer bottle      & soda bottle      & water bottle  \\
6 & bus        & trolleybus       & carriage       & minibus \\
7 & car        & racing car       & station wagon    & minivan  \\
8 & cat        & tabby cat       & Egyptian cat     & tiger cat \\
9 & chair      & rocking chair   & dining table     & folding chair\\
10 & cow       & ox               & oxcart         & water ox \\
11 & dining table & dining table  & china closet     & restaurant  \\
12 & dog       & Labrador retriever  & pit bull terrier  & beagle  \\
13 & horse     & sorrel          & ox              & fox squirrel  \\
14 & motorbike & moped         & scooter         & disc brake \\
15 & person    & cowboy hat       & crash helmet     & crutch \\
16 & potted plant  & pot          & pencil case    & greenhouse \\
17 & sheep     & ram           & llama           & bighorn sheep \\
18 & sofa      & studio couch    & quilt            & rocking chair \\
19 & train     & carriage         & electric locomotive  & freight car \\
20 & tv/monitor* & - & - & -  %

\end{tabular}

\hspace{5mm}

\label{tab:CMM}
\end{table*}

\newcommand{\rot}[1]{\rotatebox{50}{#1}}

\begin{table*}[htbp]
\centering
\caption{Per-round accuracy measures in the 10-1 disjoint scenario. 
In the top part we report the PA (left) and IoU (right) of the last class currently introduced. 
The bottom part, instead, shows the mean IoU over the old classes up to the ongoing step (left), as well as the overall mean IoU including the new classes (right). The classes added at each incremental step are: %
 1:\textit{dining table}, 2:\textit{dog}, 3:\textit{horse}, 4:\textit{motorbike}, 5:\textit{person}, 6:\textit{potted plant}, 7:\textit{sheep}, 8:\textit{sofa}, 9:\textit{train} and 10:\textit{tv/monitor}. Best in \textbf{bold}.}
\footnotesize
\setlength{\tabcolsep}{0.4pt} %
\renewcommand{\arraystretch}{1.1}
 
 \begin{tabular}{l|cccccccccc}
 \multicolumn{1}{c|}{PA (new)} 
 & \rot{step 1} & \rot{step 2} & \rot{step 3} & \rot{step 4} & \rot{step 5} & \rot{step 6} & \rot{step 7} & \rot{step 8} & \rot{step 9} & \rot{step 10} \\
 \hline
  FT &  69.2  & \textbf{90.9}  & \textbf{87.1}  & \textbf{79.2}  & 88.3  & 3.8   & 26.9  & \textbf{49.9}  & 55.0    & 57.5 \\
  ILT \cite{michieli2019incremental} &  70.2  & 88.3  & 44.3  & 52.8  & 86.9  & 58.7  & 22.5  & 43.6  & 48.8  & 56.3 \\
  MiB \cite{cermelli2020modeling} &  \textbf{75.5}  & 87.2  & 38.5  & 51.2  & \textbf{89.6}  & \textbf{61.0}    & 6.1   & 38.2  & 47.4  & \textbf{60.2} \\
  SDR \cite{michieli2021continual} &  68.7  & 73.2  & 34.0  & 31.0    & 84.5  & 55.6  & 15.7  & 39.0    & 45.1  & 55.8 \\
  RECALL (GAN) & 24.0  & 56.7  & 38.1  & 58.5  & 76.2  & 27.1  & 36.8  & 25.4  & 53.0  & 45.0 \\
  RECALL (Web) & 23.8  & 57.3  & 46.1  & 62.3  & 79.3  & 36.9  & \textbf{42.8}  & 26.9  & \textbf{64.2}  & 57.4 \\
 \end{tabular}%
 \quad
 \begin{tabular}{l|cccccccccc}
 \multicolumn{1}{c|}{IoU (new)} 
  & \rot{step 1} & \rot{step 2} & \rot{step 3} & \rot{step 4} & \rot{step 5} & \rot{step 6} & \rot{step 7} & \rot{step 8} & \rot{step 9} & \rot{step 10} \\
 \hline
  FT    & 16.2  & 37.0  & 15.8  & 12.6  & \textbf{78.4}  & 3.2   & 9.2   & 16.3  & 9.7   & 19.9 \\
  ILT \cite{michieli2019incremental}  & 14.6  & 37.2  & 14.5  & 11.5  & 68.6  & 5.5   & 4.9   & 10.3  & 11.1  & 13.8 \\
  MiB  \cite{cermelli2020modeling}  & 14.6  & 38.0  & 12.2  & 9.5   & 66.0  & 9.6   & 1.7   & 5.5   & 10.9  & 7.0 \\
  SDR \cite{michieli2021continual}   & \textbf{25.0}  & 54.0  & 13.3  & 10.9  & 69.2  & 10.9  & 5.4   & 8.9   & 15.4  & 17.8 \\
  RECALL (GAN) & 22.4  & 53.9  & 36.1  & 54.6  & 64.4  & 23.6  & 34.4  & 23.6  & 49.3  & 38.8 \\
  RECALL (Web) & 22.1  & \textbf{54.5}  & \textbf{43.7}  & \textbf{58.4}  & 67.9  & \textbf{30.5}  & \textbf{39.6}  & \textbf{24.7}  & \textbf{58.3}  & \textbf{45.0} \\

 \end{tabular}%
 \vspace{3mm}
 \begin{tabular}{l|cccccccccc}
 \multicolumn{1}{c|}{mIoU (old)} 
  & \rot{step 1} & \rot{step 2} & \rot{step 3} & \rot{step 4} & \rot{step 5} & \rot{step 6} & \rot{step 7} & \rot{step 8} & \rot{step 9} & \rot{step 10} \\
 \hline
  FT    & 50.4  & 23.9  & 26.1  & 20.5  & 4.6   & 4.7   & 4.2   & 3.7   & 3.8   & 3.5 \\
  ILT \cite{michieli2019incremental}   & 59.4  & 33.5  & 35.2  & 20.9  & 11.7  & 11.2  & 10.9  & 10.2  & 7.1   & 7.2 \\
  MiB \cite{cermelli2020modeling}  & 64.3  & 53.6  & 61.1  & 26.0  & 25.3  & 27.6  & 19.0  & 16.8  & 9.7   & 12.6 \\
  SDR \cite{michieli2021continual}   & 64.9  & 57.5  & 61.6  & 35.2  & 30.2  & 32.4  & 28.7  & 27.5  & 19.2  & 21.0 \\
  RECALL (GAN) & \textbf{74.7}  & \textbf{64.7}  & 63.4  & 63.2  & 60.7  & 62.9  & 57.9  & 56.5  & 53.5  & 54.6 \\
    RECALL (Web) & 74.3  & 63.9  & \textbf{64.6}  & \textbf{65.0}  & \textbf{63.5}  & \textbf{65.5}  & \textbf{60.6}  & \textbf{60.1}  & \textbf{58.3}  & \textbf{58.4} \\
 \end{tabular}%
 \quad
 \begin{tabular}{l|cccccccccc}
 \multicolumn{1}{c|}{mIoU (all)} 
  & \rot{step 1} & \rot{step 2} & \rot{step 3} & \rot{step 4} & \rot{step 5} & \rot{step 6} & \rot{step 7} & \rot{step 8} & \rot{step 9} & \rot{step 10} \\
 \hline
  FT  & 47.3  & 25.0  & 25.3  & 19.9  & 9.5   & 4.6   & 4.5   & 4.4   & 4.1   & 4.3 \\
  ILT \cite{michieli2019incremental} & 55.3  & 33.8  & 33.6  & 20.2  & 15.5  & 10.8  & 10.5  & 10.2  & 7.3   & 7.5 \\
  MiB \cite{cermelli2020modeling} & 59.8  & 52.3  & 57.3  & 24.8  & 28.0  & 26.5  & 18.0  & 16.2  & 9.8   & 12.3 \\
  SDR \cite{michieli2021continual} & 61.3  & 57.2  & 57.9  & 33.5  & 32.8  & 31.1  & 27.3  & 26.5  & 19.0  & 20.8 \\
  RECALL (GAN) & \textbf{70.4}  & \textbf{63.9}  & 61.5  & 62.7  & 60.9  & 60.6  & 56.6  & 54.8  & 53.3  & 53.9 \\
  RECALL (Web) & 69.9  & 63.2  & \textbf{63.1}  & \textbf{64.6}  & \textbf{63.7}  & \textbf{63.5}  & \textbf{59.4}  & \textbf{58.3}  & \textbf{58.3}  & \textbf{57.8} \\
\end{tabular}%
 
\label{tab:perround_tables}%
\end{table*}%

\begin{table*}[htbp]
\renewcommand{\arraystretch}{1.3}
  \centering
  \small
  \setlength{\tabcolsep}{1.1pt} %
   \caption{Per-class IoU of compared methods in disjoint experimental protocol on multiple scenarios of Pascal VOC 2012.}
    \begin{tabular}{c<{\hspace{1mm}}|l|>{\hspace{1mm}}cccccccccccccccccccc!{\color{lightgray}\vrule width 1pt}c<{\hspace{1mm}}|>{\hspace{1mm}}ccc|}
    \cline{2-26}
       & Method  & \rotatebox{90}{backgr.} 
       &  \rotatebox{90}{aero}  & \rotatebox{90}{bike}  & \rotatebox{90}{bird} 
               & \rotatebox{90}{boat}    & \rotatebox{90}{bottle} & \rotatebox{90}{bus}   &\rotatebox{90}{car} 
               & \rotatebox{90}{cat}     & \rotatebox{90}{chair}  & \rotatebox{90}{cow}   & \rotatebox{90}{din. table}
               & \rotatebox{90}{dog}     & \rotatebox{90}{horse}  & \rotatebox{90}{mbike} & \rotatebox{90}{person} 
               & \rotatebox{90}{plant}   &  \rotatebox{90}{sheep} & \rotatebox{90}{sofa}  & \multicolumn{1}{c}{\rotatebox{90}{train}} 
               & \rotatebox{90}{tv}      & old   & new   & all \\
     \cline{2-26}
    \cline{2-26}
     \multirow{5}{*}{\rotatebox{90}{19-1}} 
     & FT & 72.4 & 62.4 & 6.7 & 45.0 & 47.1 & 39.5 & 33.7 & 40.9 & 25.7 & 4.3 & 54.0 & 8.0 & 25.0 & 50.4 & 50.6 & 0.0 & 35.3 & 43.0 & 0.8 & 59.5 & 13.2 & 35.2 & 13.2 & 34.2 \\
     
     & Inpainting & 91.0 & 83.9 & 35.1 & 77.3 & 62.3 & 70.7 & 77.9 & 73.4 & 85.7 & 31.5 & 73.1 & 48.0 & 81.3 & 74.4 & 64.6 & 81.0 & 44.1 & 75.7 & 41.3 & 74.5 & 30.4  & 66.1 & 30.4 & 65.6 \\ 
     & GAN & 91.7 & 82.8 & 32.3 & 82.6 & 62.8 & 74.1 & 86.2 & 79.6 & 86.0 & 30.0 & 58.9 & 45.9 & 80.5 & 67.9 & 73.4 & 80.6 & 35.3 & 62.9 & 39.6 & 77.9 & 50.1  & 65.2 & 50.1 & 65.8 \\ 
     & Web & 91.4 & 82.8 & 35.9 & 83.4 & 59.9 & 73.5 & 85.3 & 73.7 & 85.7 & 31.3 & 59.4 & 40.9 & 81.1 & 67.1 & 73.4 & 80.5 & 43.1 & 61.5 & 42.6 & 74.4 & 47.1  & 65.0 & 47.1 & 65.4 \\ 
     
     & Joint & 92.5 & 89.9 & 39.2 & 87.6 & 65.2 & 77.3 & 91.1 & 88.5 & 92.9 & 34.8 & 84.0 & 53.7 & 88.9 & 85.0 & 85.1 & 84.9 & 60.0 & 79.7 & 47.0 & 82.2 & 73.5 & 75.5 & 73.5 & 75.4\\
     \cline{2-26}
     
    \end{tabular}%
    \vspace{0.5mm}
    
    \begin{tabular}{c<{\hspace{1mm}}|l|>{\hspace{1mm}}cccccccccccccccc!{\color{lightgray}\vrule width 1pt}ccccc<{\hspace{1mm}}|>{\hspace{1mm}}ccc|}
     \cline{2-26}
     \multirow{5}{*}{\rotatebox{90}{15-5}} 
     & FT & 72.4 & 62.4 & 6.7 & 45.0 & 47.1 & 39.5 & 33.7 & 40.9 & 25.7 & 4.3 & 54.0 & 8.0 & 25.0 & 50.4 & 50.6 & 0.0 & 35.3 & 43.0 & 0.8 & 59.5 & 13.2 & 35.2 & 13.2 & 34.2 \\
     & Inpainting & 89.0 & 68.7 & 36.0 & 68.2 & 48.4 & 71.4 & 12.8 & 77.3 & 85.6 & 26.7 & 8.1 & 48.8 & 80.3 & 61.6 & 68.8 & 78.7 & 20.1 & 29.0 & 26.3 & 38.4 & 51.8  & 56.1 & 33.1 & 52.2 \\  
     & GAN & 90.4 & 78.8 & 35.0 & 79.5 & 60.3 & 75.7 & 79.3 & 78.7 & 85.9 & 22.8 & 55.0 & 46.6 & 80.0 & 67.4 & 72.1 & 77.8 & 37.3 & 60.2 & 32.2 & 64.4 & 55.1  & 66.3 & 49.8 & 63.5 \\  
     & Web &90.8  & 82.2 & 35.5 & 81.7 & 63.9 & 75.3 & 85.0 & 77.8 & 86.3 & 28.0 & 67.5 & 48.7 & 81.0 & 72.7 & 73.8 & 78.0 & 40.4 & 65.7 & 31.9 & 69.1 & 57.6  & 69.2 & 52.9 & 66.3  \\  
     
     & Joint & 92.5 & 89.9 & 39.2 & 87.6 & 65.2 & 77.3 & 91.1 & 88.5 & 92.9 & 34.8 & 84.0 & 53.7 & 88.9 & 85.0 & 85.1 & 84.9 & 60.0 & 79.7 & 47.0 & 82.2 & 73.5 & 77.5 & 68.5 & 75.4\\
     \cline{2-26} 
    \end{tabular}%
    
    \vspace{0.5mm}
    
    \begin{tabular}{c<{\hspace{1mm}}|l|>{\hspace{1mm}}cccccccccccccccc!{\color{lightgray}\vrule width 1pt}ccccc<{\hspace{1mm}}|>{\hspace{1mm}}ccc|}
     \cline{2-26}
     \multirow{5}{*}{\rotatebox{90}{15-1}} 
     
     & FT & 74.2 & 27.2 & 0.0 & 1.6 & 15.1 & 11.3 & 0.0 & 4.1 & 0.5 & 0.0 & 0.0 & 0.0 & 0.0 & 0.2 & 0.2 & 0.0 & 27.0 & 25.6 & 28.9 & 33.5 & 52.2 & 8.4 & 33.5 & 14.4\\ 
 
     & Inpainting & 85.9 & 38.9 & 31.4 & 79.4 & 41.5 & 71.3 & 28.9 & 62.6 & 85.6 & 32.2 & 29.6 & 50.2 & 76.6 & 69.2 & 55.3 & 80.2 & 18.5 & 37.4 & 36.3 & 19.8 & 17.9  & 55.5 & 26.0 & 49.9 \\  
     & GAN & 90.5 & 80.7 & 34.5 & 79.5 & 59.1 & 75.5 & 72.7 & 78.2 & 85.3 & 25.3 & 59.0 & 39.9 & 79.9 & 68.8 & 72.5 & 78.6 & 23.2 & 58.0 & 39.2 & 60.1 & 43.8  & 66.0 & 44.9 & 62.1 \\  
     & Web & 90.5 & 82.1 & 34.4 & 81.5 & 62.6 & 76.0 & 82.3 & 77.0 & 85.1 & 27.4 & 63.6 & 39.4 & 80.3 & 71.9 & 72.2 & 78.4 & 35.4 & 64.4 & 35.7 & 61.9 & 48.7  & 67.6 & 49.2 & 64.3 \\  
     
     & Joint & 92.5 & 89.9 & 39.2 & 87.6 & 65.2 & 77.3 & 91.1 & 88.5 & 92.9 & 34.8 & 84.0 & 53.7 & 88.9 & 85.0 & 85.1 & 84.9 & 60.0 & 79.7 & 47.0 & 82.2 & 73.5 & 77.5 & 68.5 & 75.4 \\
     \cline{2-26}
    \end{tabular}%
    
    \vspace{0.5mm}
    
    \begin{tabular}{c<{\hspace{1mm}}|l|>{\hspace{1mm}}ccccccccccc!{\color{lightgray}\vrule width 1pt}cccccccccc<{\hspace{1mm}}|>{\hspace{1mm}}ccc|}
    \cline{2-26} 
     \multirow{5}{*}{\rotatebox{90}{10-10}}
      
     & FT & 82.1 & 0.2 & 0.0 & 1.2 & 0.0 & 1.4 & 0.0 & 0.0 & 0.0 & 0.0 & 0.0 & 52.3 & 73.2 & 49.8 & 73.1 & 81.8 & 41.4 & 49.7 & 49.1 & 76.1 & 62.2 & 7.7 & 60.9  & 33.0 \\
     
     & Inpainting & 90.9 & 81.8 & 34.1 & 73.1 & 58.6 & 73.3 & 85.6 & 78.8 & 78.2 & 29.0 & 29.1 & 43.7 & 66.6 & 47.7 & 73.0 & 74.2 & 29.6 & 57.3 & 38.8 & 70.9 & 61.4  & 62.2 & 56.3 & 60.7 \\  
     & GAN & 90.8 & 83.3 & 30.4 & 75.8 & 61.4 & 73.5 & 80.8 & 77.2 & 72.8 & 23.6 & 46.8 & 48.0 & 65.4 & 55.3 & 66.1 & 72.5 & 36.8 & 58.3 & 36.1 & 67.1 & 55.6  & 62.6 & 56.1 & 60.8 \\  
     & Web & 90.9 & 82.3 & 32.7 & 75.4 & 63.2 & 72.8 & 81.7 & 73.5 & 76.2 & 24.2 & 58.5 & 46.5 & 68.8 & 60.2 & 64.7 & 73.3 & 38.3 & 58.3 & 34.2 & 68.5 & 56.2  & 64.1 & 56.9 & 61.9 \\  
          
     & Joint & 92.5 & 89.9 & 39.2 & 87.6 & 65.2 & 77.3 & 91.1 & 88.5 & 92.9 & 34.8 & 84.0 & 53.7 & 88.9 & 85.0 & 85.1 & 84.9 & 60.0 & 79.7 & 47.0 & 82.2 & 73.5 & 76.6 & 74.0 & 75.4 \\
     \cline{2-26} 
    \end{tabular}%
    
    \vspace{0.5mm}
  
  \begin{tabular}{c<{\hspace{1mm}}|l|>{\hspace{1mm}}ccccccccccc!{\color{lightgray}\vrule width 1pt}cccccccccc<{\hspace{1mm}}|>{\hspace{1mm}}ccc|}
    \cline{2-26} 
     \multirow{5}{*}{\rotatebox{90}{10-5}}
     
     & FT & 78.2 & 0.0 & 0.0 & 0.0 & 0.0 & 1.0 & 0.0 & 0.0 & 0.0 & 0.0 & 0.0 & 16.4 & 13.7 & 23.4 & 55.7 & 46.7 & 37.4 & 39.8 & 47.9 & 75.6 & 62.3 & 7.2 & 41.9 & 23.7 \\
     
     & Inpainting & 88.5 & 58.8 & 31.9 & 55.4 & 58.2 & 69.2 & 0.2 & 78.3 & 83.2 & 28.2 & 5.0 & 36.4 & 71.6 & 34.7 & 61.4 & 74.1 & 20.4 & 26.2 & 25.5 & 34.0 & 47.9  & 46.8 & 43.2 & 47.1 \\  
     & GAN & 89.3 & 77.9 & 28.9 & 72.1 & 59.3 & 73.6 & 75.1 & 75.8 & 79.5 & 20.5 & 37.2 & 44.2 & 67.8 & 50.9 & 59.5 & 71.3 & 31.4 & 51.9 & 32.3 & 63.1 & 53.0  & 60.0 & 52.5 & 57.8 \\  
     & Web & 89.5 & 80.8 & 31.2 & 74.6 & 61.6 & 72.0 & 81.6 & 74.3 & 80.6 & 19.8 & 55.1 & 44.3 & 69.2 & 56.9 & 56.5 & 71.7 & 39.8 & 59.0 & 30.2 & 69.7 & 54.2  & 63.2 & 55.1 & 60.6  \\  
          
     & Joint & 92.5 & 89.9 & 39.2 & 87.6 & 65.2 & 77.3 & 91.1 & 88.5 & 92.9 & 34.8 & 84.0 & 53.7 & 88.9 & 85.0 & 85.1 & 84.9 & 60.0 & 79.7 & 47.0 & 82.2 & 73.5 & 76.6 & 74.0 & 75.4  \\
     \cline{2-26} 
    \end{tabular}%
    
    \vspace{0.5mm}
    
    \begin{tabular}{c<{\hspace{1mm}}|l|>{\hspace{1mm}}ccccccccccc!{\color{lightgray}\vrule width 1pt}cccccccccc<{\hspace{1mm}}|>{\hspace{1mm}}ccc|}
    \cline{2-26} 
     \multirow{5}{*}{\rotatebox{90}{10-1}} 
     & FT & 69.4 & 0.0 & 0.0 & 0.0 & 0.0 & 0.0 & 0.0 & 0.0 & 0.0 & 0.0 & 0.0 & 0.0 & 0.0 & 0.0 & 0.0 & 0.0 & 0.0 & 0.0 & 0.0 & 0.0 & 19.9 & 6.3 & 2.0 & 4.3  \\
     
     & Inpainting & 85.0 & 35.3 & 28.6 & 72.6 & 37.2 & 67.9 & 16.9 & 65.3 & 83.0 & 32.1 & 13.3 & 35.1 & 41.4 & 5.0 & 22.5 & 71.7 & 21.5 & 16.9 & 34.5 & 19.2 & 13.0  & 45.2 & 28.1 & 39.0  \\  
     & GAN & 88.8 & 77.9 & 26.6 & 71.8 & 58.6 & 73.2 & 63.5 & 74.0 & 75.7 & 20.5 & 41.3 & 34.2 & 60.5 & 43.6 & 58.6 & 66.2 & 15.8 & 51.7 & 37.4 & 53.0 & 38.8  & 58.3 & 46.0 & 53.9 \\  
     & Web & 89.1 & 79.1 & 31.0 & 74.4 & 62.2 & 66.5 & 81.7 & 74.1 & 78.7 & 19.4 & 56.2 & 41.8 & 62.7 & 58.5 & 62.1 & 66.6 & 8.5 & 59.3 & 36.8 & 59.2 & 45.0  & 62.3 & 50.0 & 57.8 \\  
          
     & Joint & 92.5 & 89.9 & 39.2 & 87.6 & 65.2 & 77.3 & 91.1 & 88.5 & 92.9 & 34.8 & 84.0 & 53.7 & 88.9 & 85.0 & 85.1 & 84.9 & 60.0 & 79.7 & 47.0 & 82.2 & 73.5 & 76.6 & 74.0 & 75.4 \\
     \cline{2-26} 
     
    \end{tabular}%
  \label{tab:voc_per_class}%
\end{table*}%

\newcommand{\imgsizee}{28mm}
\begin{figure*}[htbp]
\centering

\newcommand{\includefixed}[1]{\includegraphics[width=0.16\textwidth, height=0.1\textheight]{#1}}
\setlength{\tabcolsep}{0.7pt} %
\renewcommand{\arraystretch}{0.6}
\centering
\begin{subfigure}[htbp]{1\textwidth}
\centering
\begin{tabular}{cc<{\hspace{1mm}}|>{\hspace{1mm}}cc<{\hspace{1mm}}|>{\hspace{1mm}}cc}
    
    \multicolumn{2}{c|}{\large Pascal} & \multicolumn{2}{c|}{\large GAN} & \multicolumn{2}{c}{\large Flickr} \\[1ex]
   \includefixed{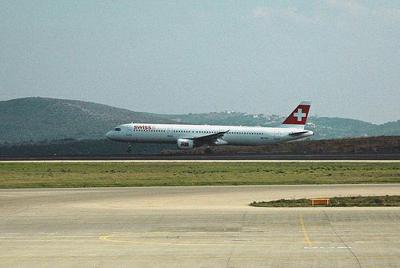} &
   \includefixed{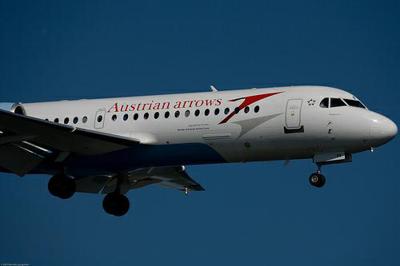} & 
   \includefixed{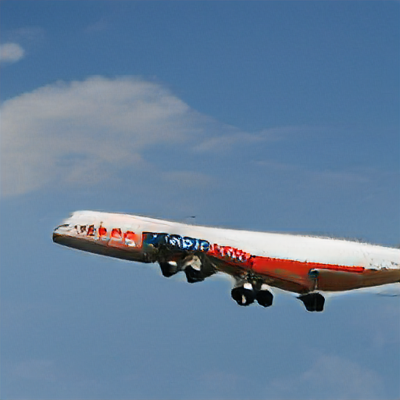} &
   \includefixed{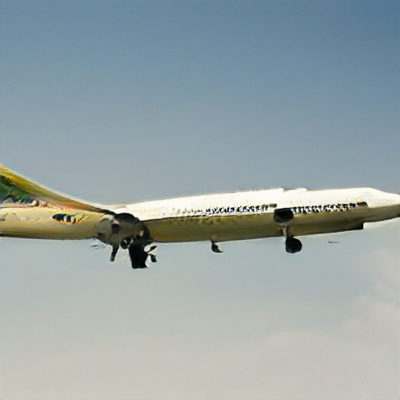} & 
   \includefixed{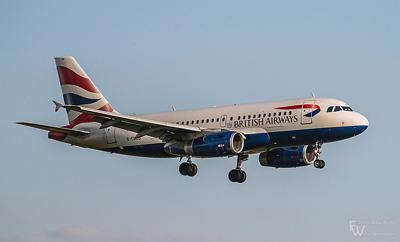} & 
   \includefixed{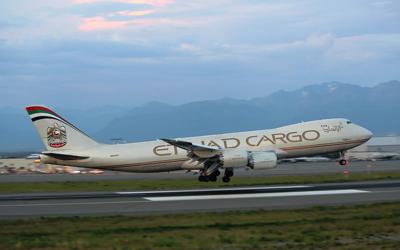}  \\ 
   
   \includefixed{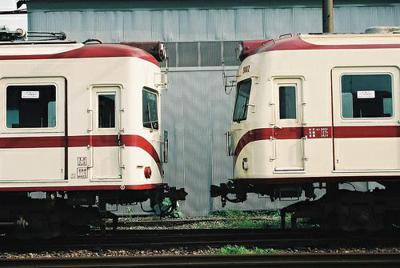} &
   \includefixed{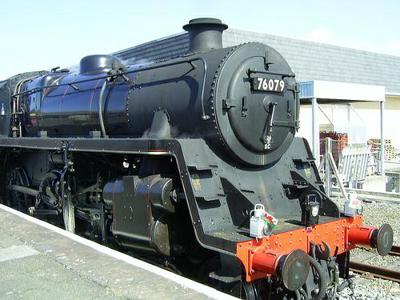} & 
   \includefixed{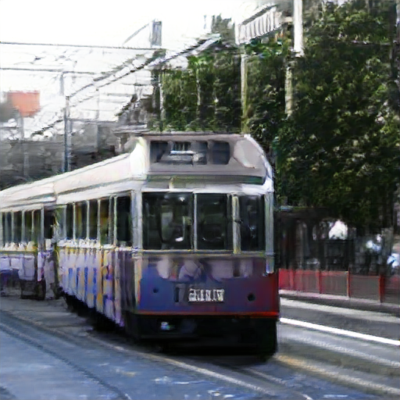} &
   \includefixed{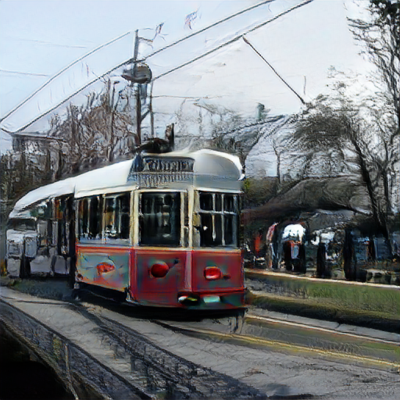} & 
   \includefixed{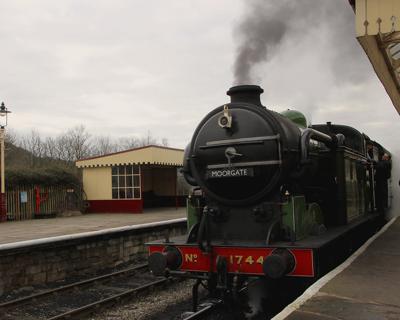} & 
   \includefixed{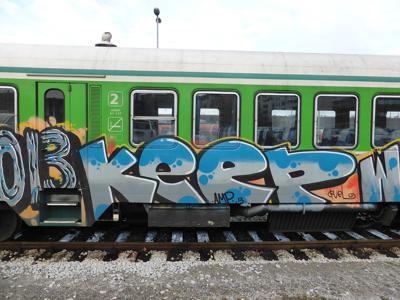}  \\ 
   
   \includefixed{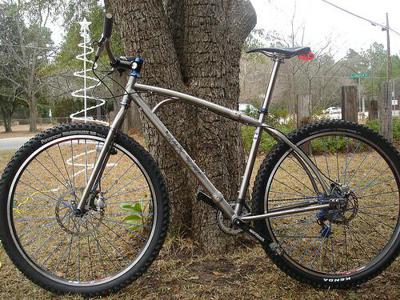} &
   \includefixed{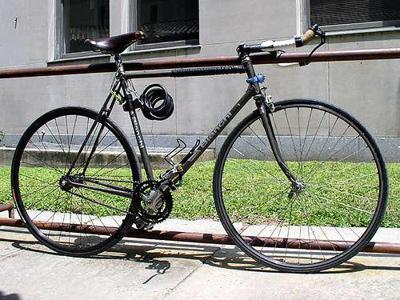} & 
   \includefixed{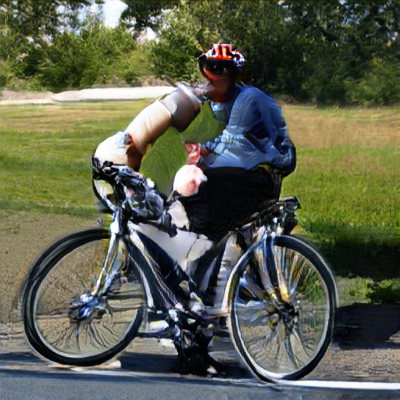} &
   \includefixed{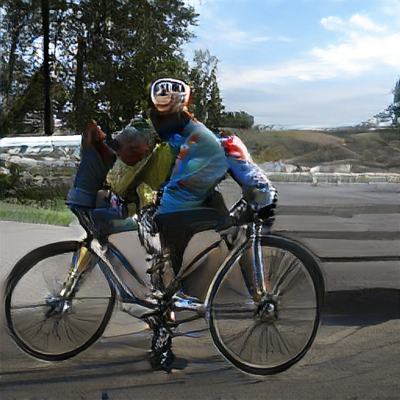} & 
   \includefixed{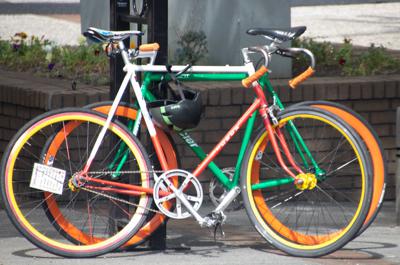} & 
   \includefixed{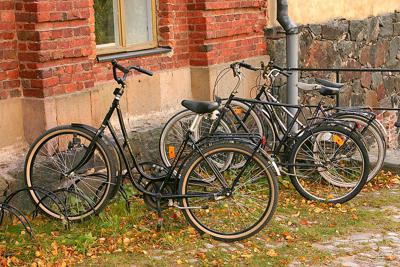}  \\ 
   
   \includefixed{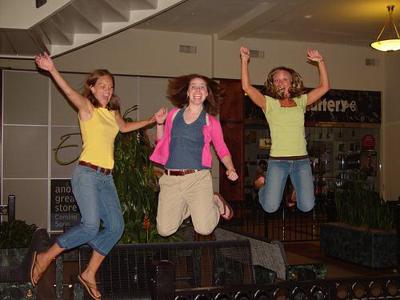} &
   \includefixed{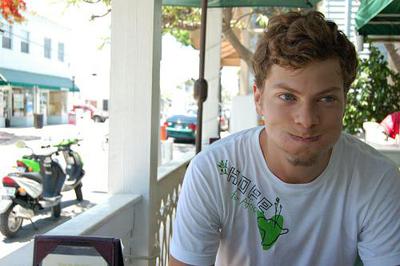} & 
   \includefixed{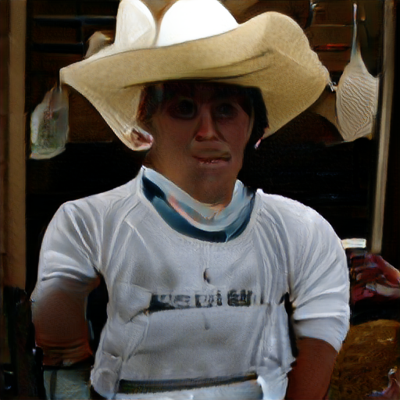} &
   \includefixed{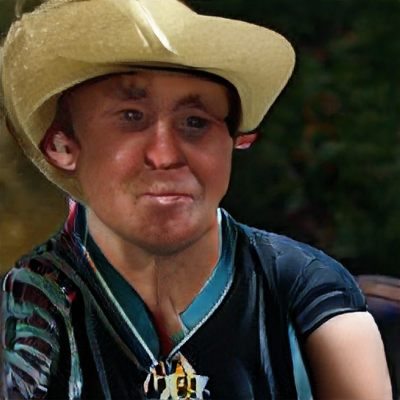} & 
   \includefixed{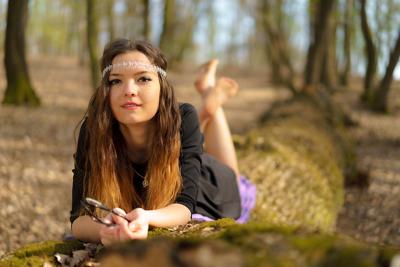} & 
   \includefixed{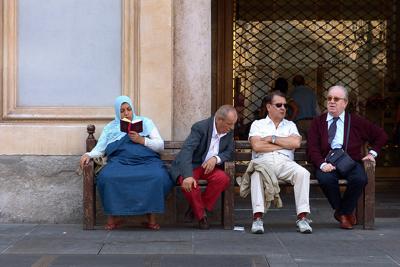}  \\ 
   
    \includefixed{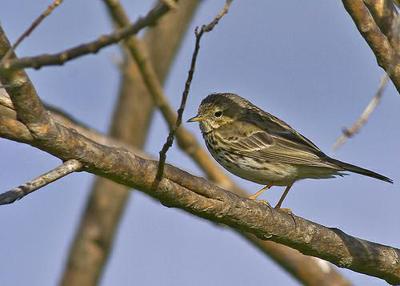} &
   \includefixed{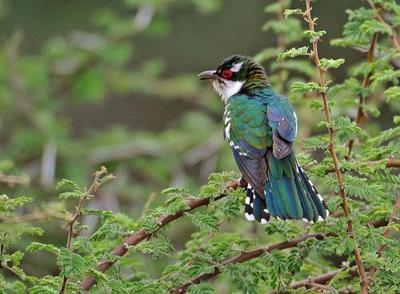} & 
   \includefixed{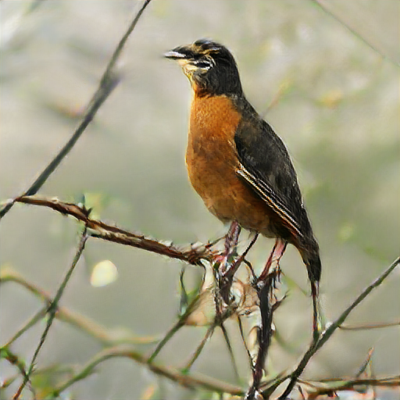} &
   \includefixed{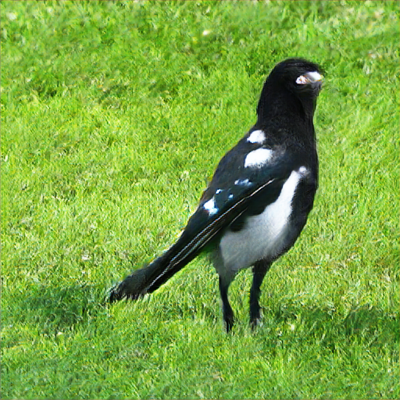} & 
   \includefixed{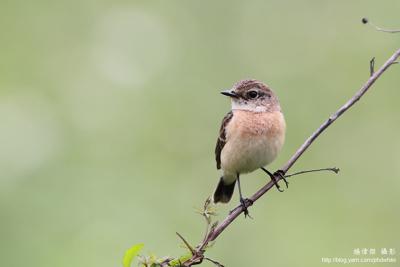} & 
   \includefixed{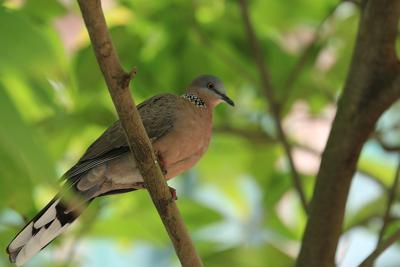}  \\

   \includefixed{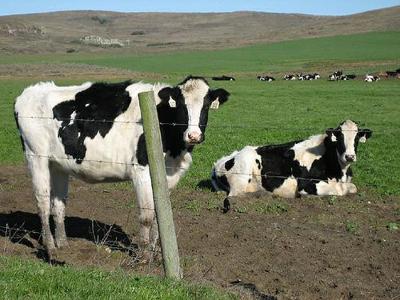} &
   \includefixed{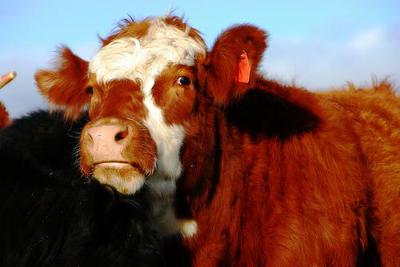} & 
   \includefixed{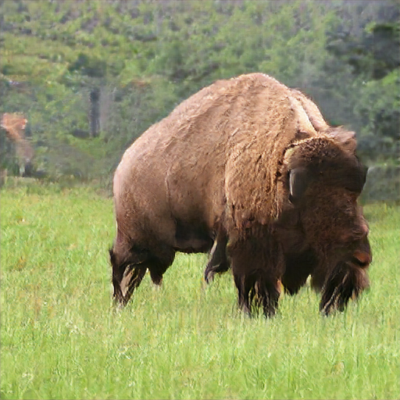} &
   \includefixed{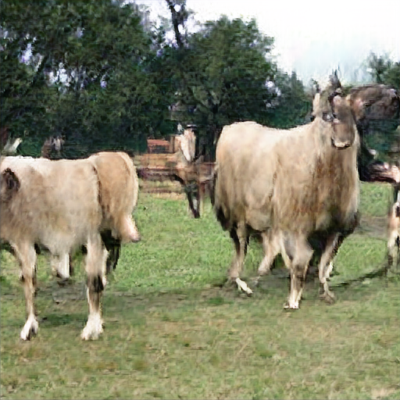} & 
   \includefixed{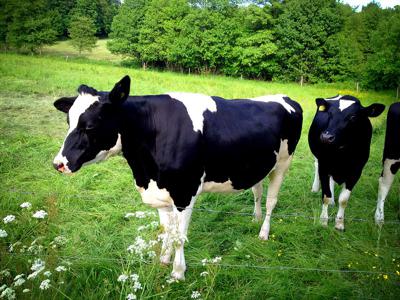} & 
   \includefixed{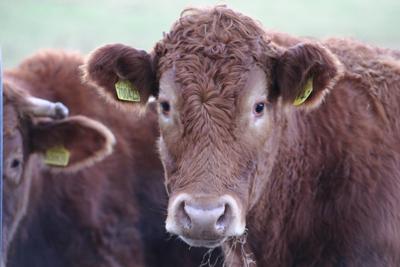}  \\ 
   
   \includefixed{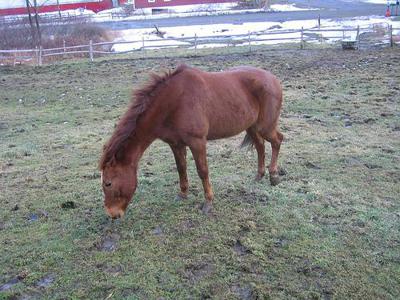} &
   \includefixed{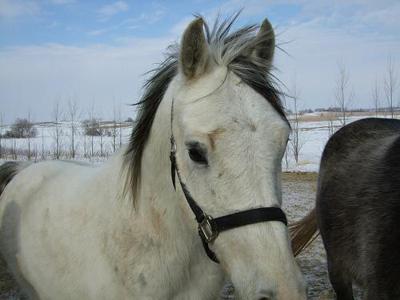} & 
   \includefixed{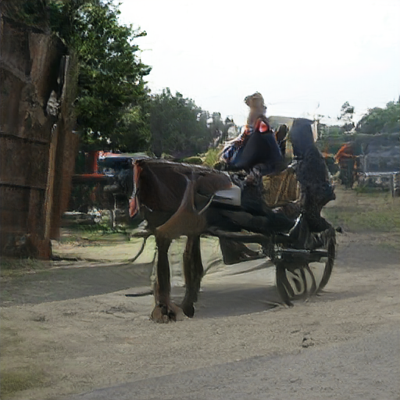} &
   \includefixed{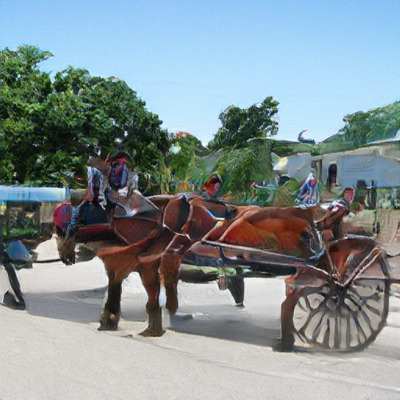} & 
   \includefixed{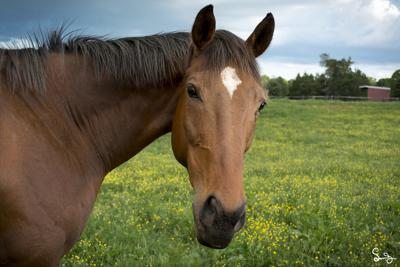} & 
   \includefixed{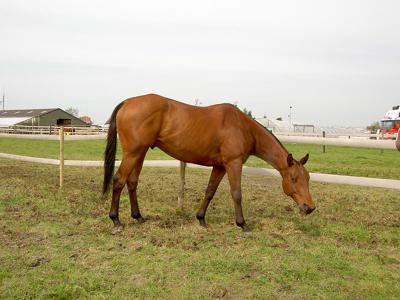}  \\ 
   
   \includefixed{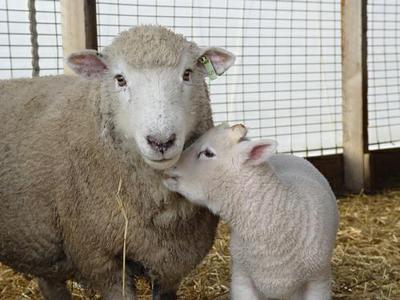} &
   \includefixed{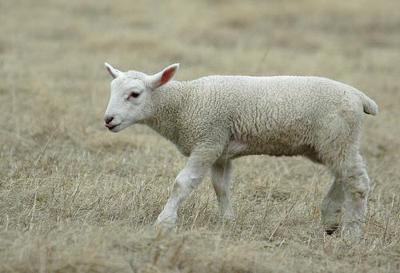} & 
   \includefixed{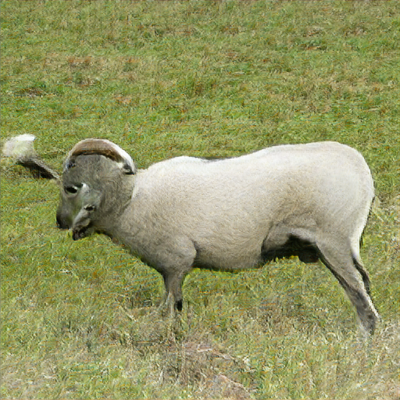} &
   \includefixed{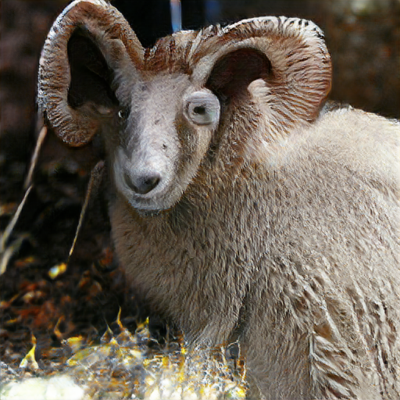} & 
   \includefixed{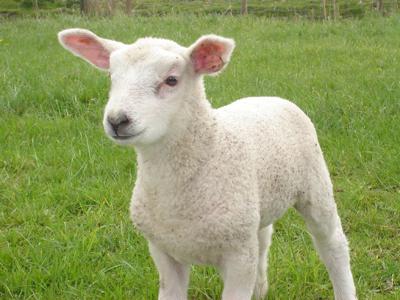} & 
   \includefixed{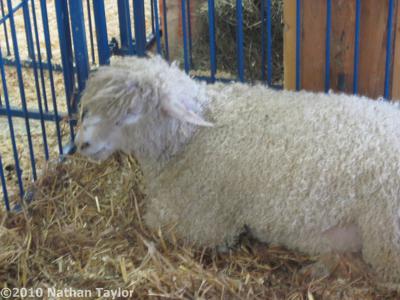}  \\

\end{tabular}
\end{subfigure}

\caption{Original images from the incremental Pascal VOC dataset, together with replay data generated by GAN or retrieved by Flickr's web crawler.}
\label{fig:generated}
\end{figure*}
\renewcommand{\imgsize}{21mm}
\begin{figure*}[htbp]
\centering

\setlength{\tabcolsep}{1pt} %
\renewcommand{\arraystretch}{0.6}
\centering
\begin{subfigure}[htbp]{1\textwidth}
\centering
\begin{tabular}{c<{\hspace{1mm}}ccccccc}
   
   &  RGB  &  GT  &  FT & Inpainting  &   Ours (GAN)  &  Ours (Web)  &  Joint \\[1ex]
   \multirow{2}{*}{\rotatebox{90}{\hspace{+30ex}19-1}} &  
   \includegraphics[width=\imgsize]{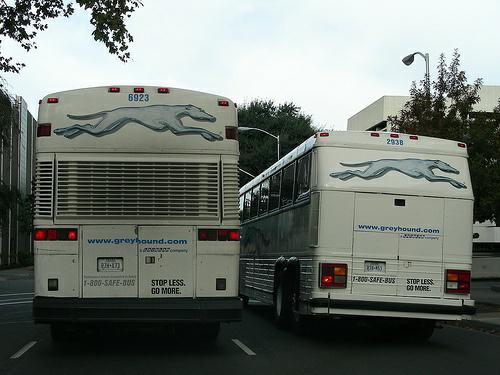} &
   \includegraphics[width=\imgsize]{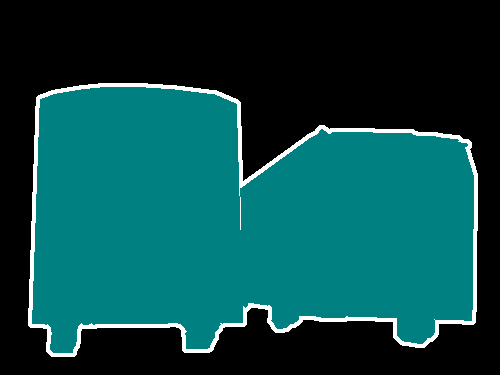} & 
   \includegraphics[width=\imgsize]{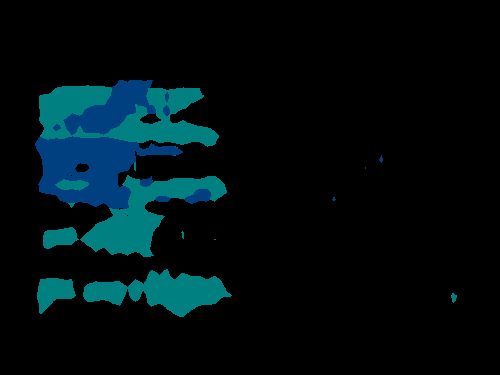} & 
   \includegraphics[width=\imgsize]{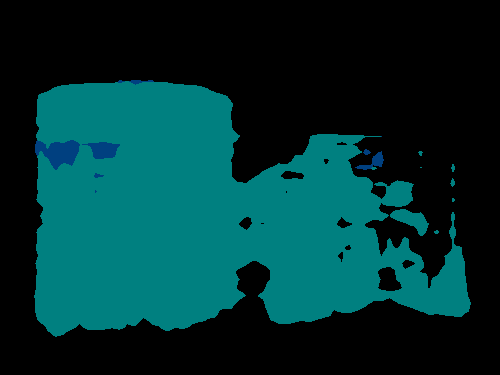} & 
   \includegraphics[width=\imgsize]{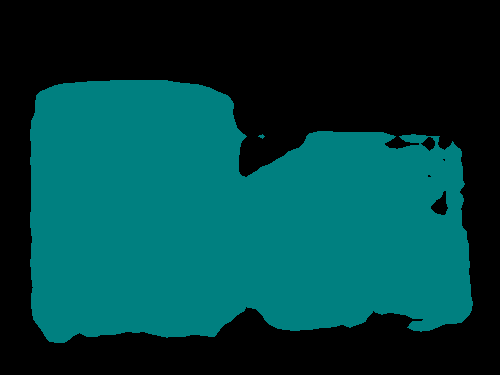} & 
   \includegraphics[width=\imgsize]{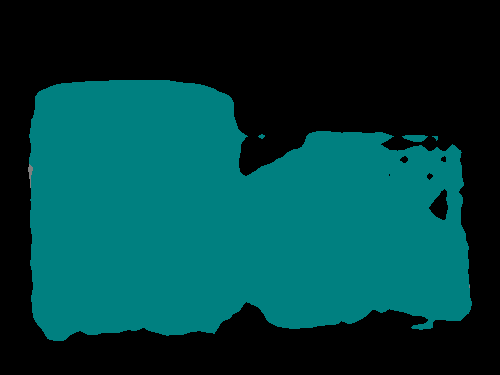} &
   \includegraphics[width=\imgsize]{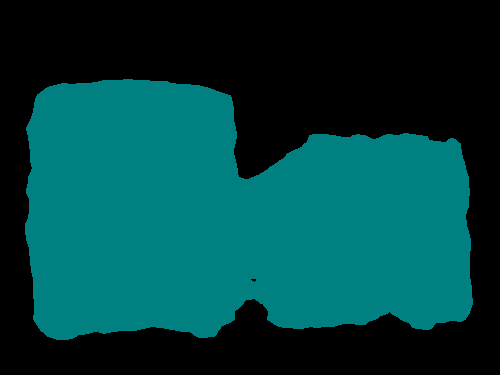} \\ 
    &  
   \includegraphics[width=\imgsize]{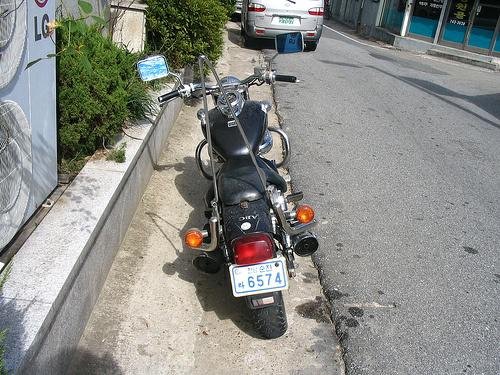} &
   \includegraphics[width=\imgsize]{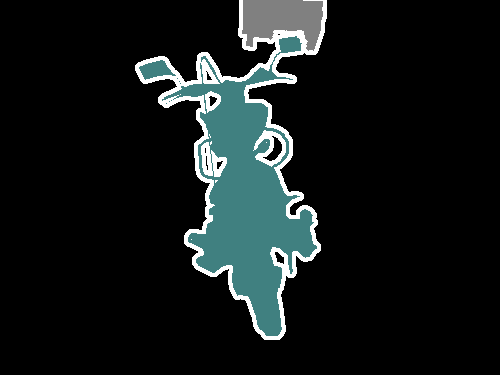} & 
   \includegraphics[width=\imgsize]{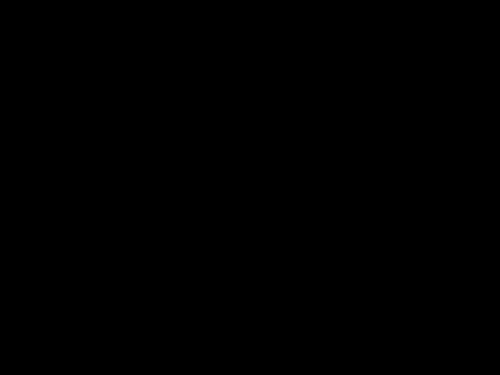} & 
   \includegraphics[width=\imgsize]{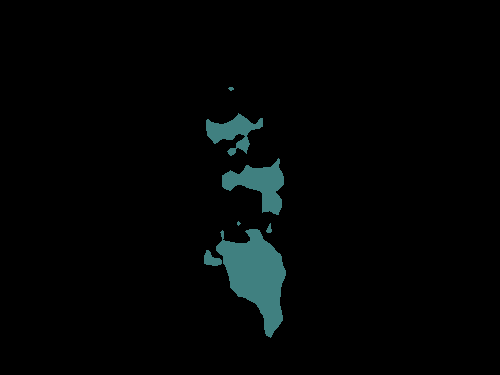} & 
   \includegraphics[width=\imgsize]{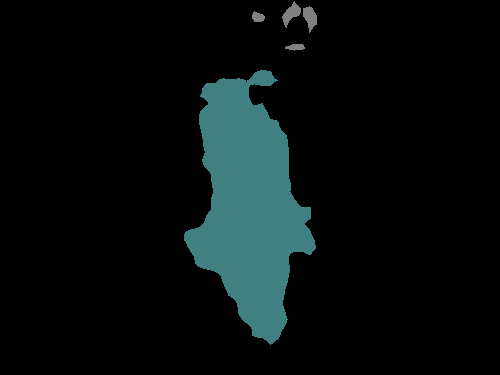} & 
   \includegraphics[width=\imgsize]{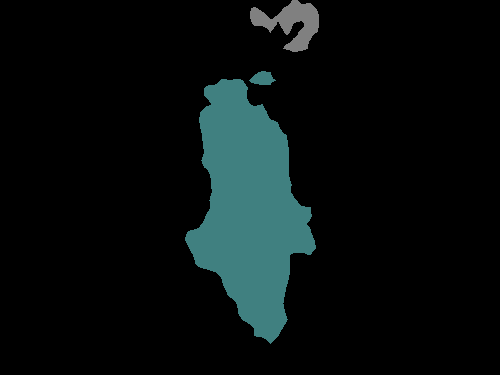} &
   \includegraphics[width=\imgsize]{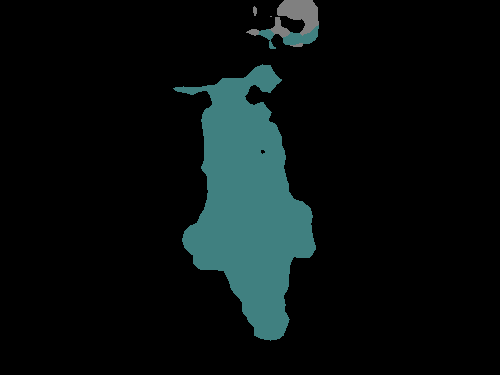} \\
   
   \hline
   
   \multirow{2}{*}{\rotatebox{90}{\hspace{+4ex}15-5}} & 
   \includegraphics[trim = 0 0 0 -6mm,clip, width=\imgsize]{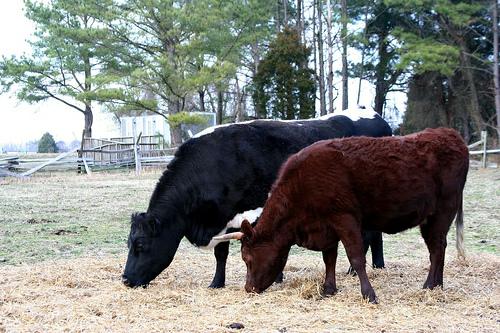} &
   \includegraphics[width=\imgsize]{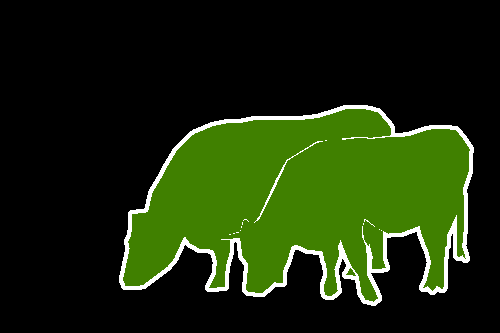} & 
   \includegraphics[width=\imgsize]{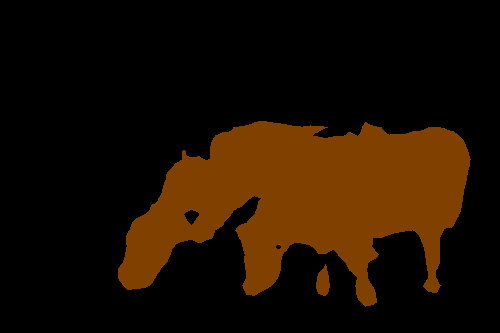} & 
   \includegraphics[width=\imgsize]{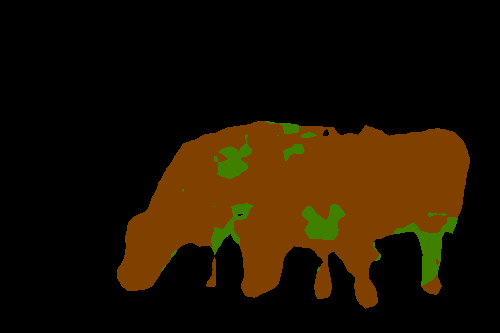} & 
   \includegraphics[width=\imgsize]{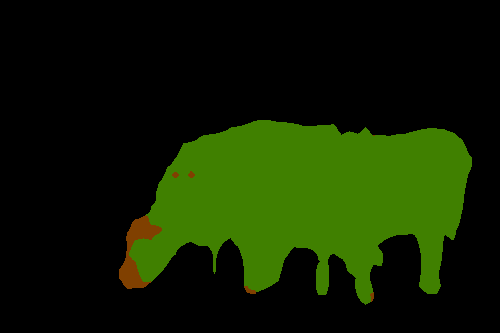} & 
   \includegraphics[width=\imgsize]{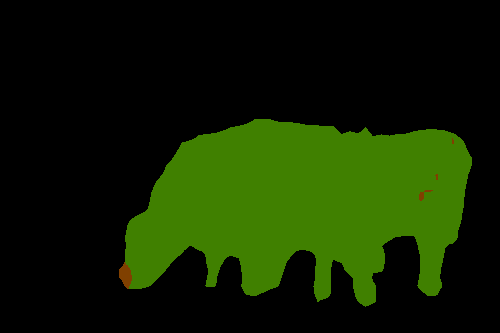} &
   \includegraphics[width=\imgsize]{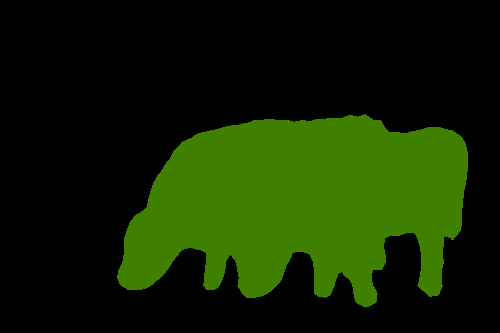}  \\ 
   & 
   \includegraphics[width=\imgsize]{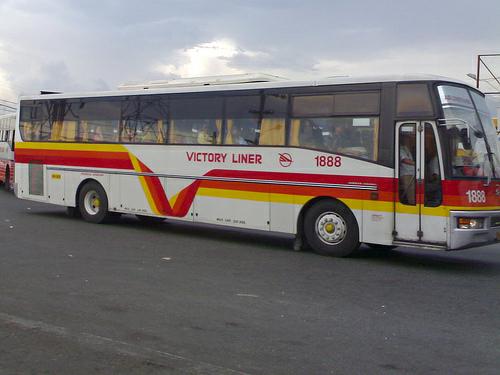} &
   \includegraphics[width=\imgsize]{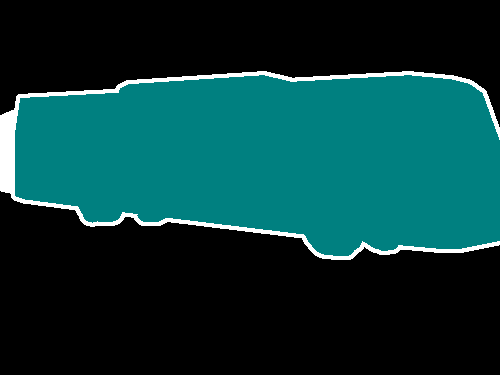} & 
   \includegraphics[width=\imgsize]{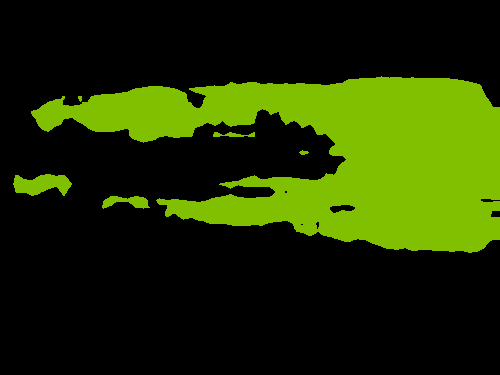} & 
   \includegraphics[width=\imgsize]{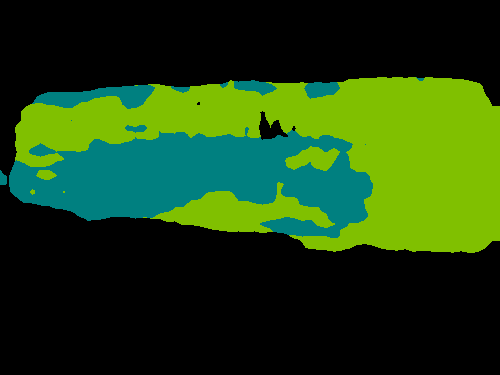} & 
   \includegraphics[width=\imgsize]{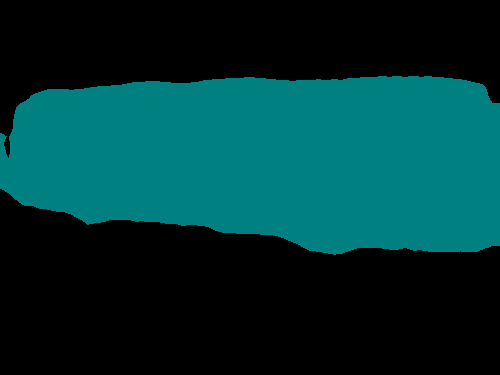} & 
   \includegraphics[width=\imgsize]{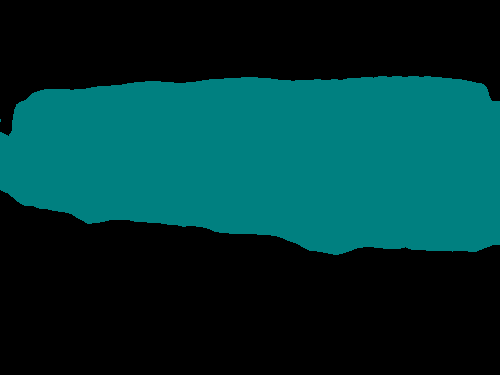} &
   \includegraphics[width=\imgsize]{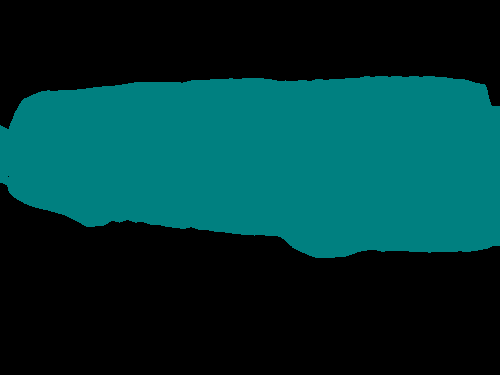}  \\ 
   \hline
   \multirow{2}{*}{\rotatebox{90}{\hspace{+4ex}15-1}} & 
   \includegraphics[trim = 0 0 0 -6mm,clip, width=\imgsize]{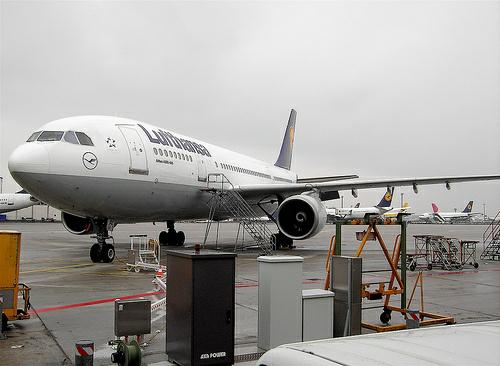} &
   \includegraphics[width=\imgsize]{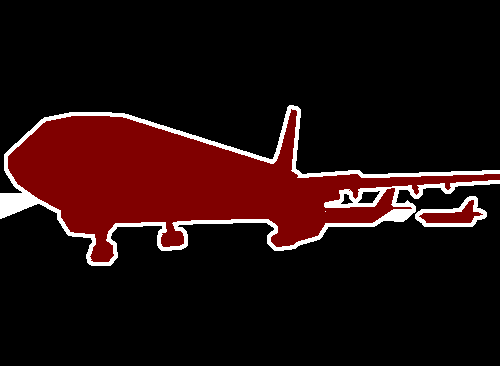} & 
   \includegraphics[width=\imgsize]{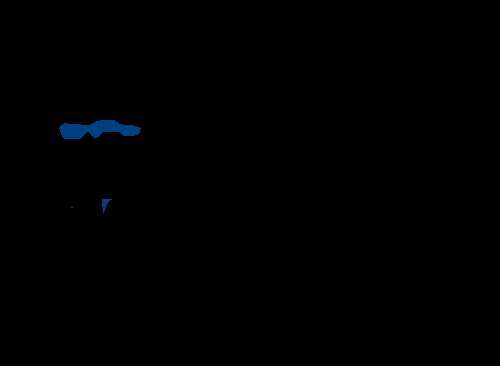} & 
   \includegraphics[width=\imgsize]{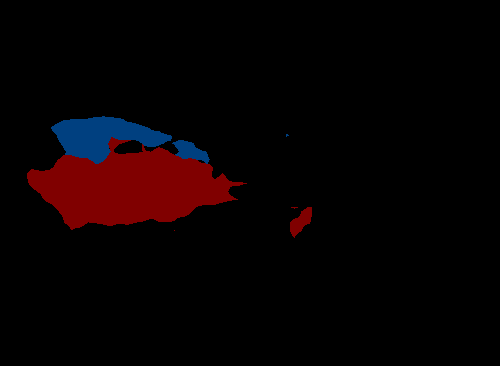} & 
   \includegraphics[width=\imgsize]{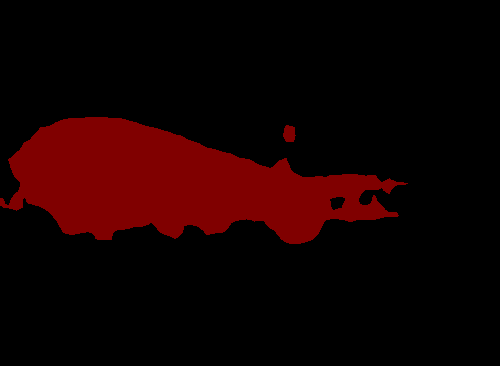} & 
   \includegraphics[width=\imgsize]{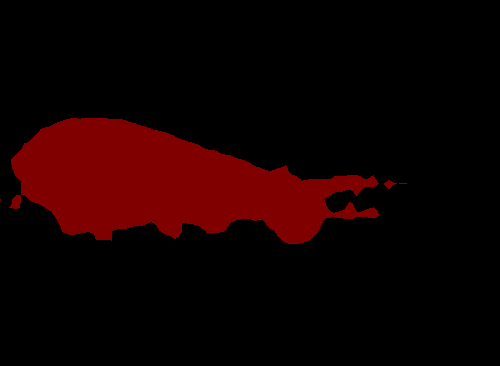} &
   \includegraphics[width=\imgsize]{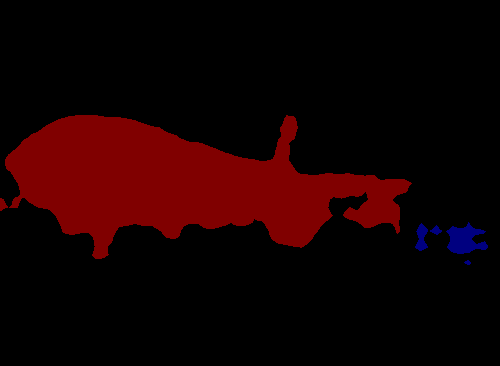}  \\ 
   & 
   \includegraphics[width=\imgsize]{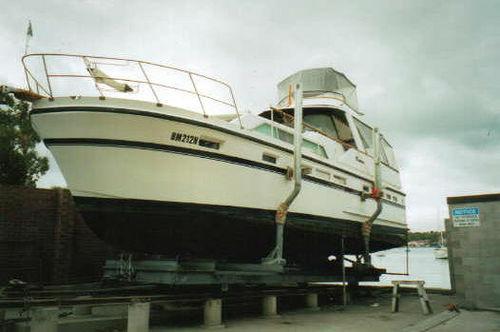} &
   \includegraphics[width=\imgsize]{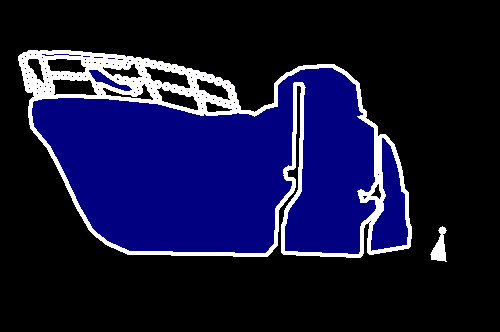} & 
   \includegraphics[width=\imgsize]{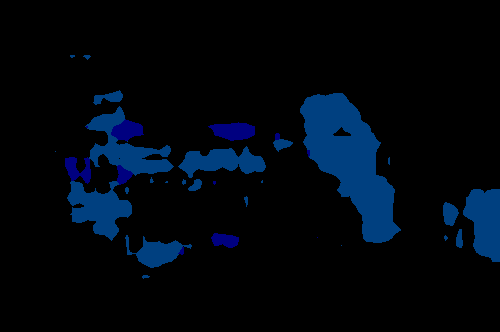} & 
   \includegraphics[width=\imgsize]{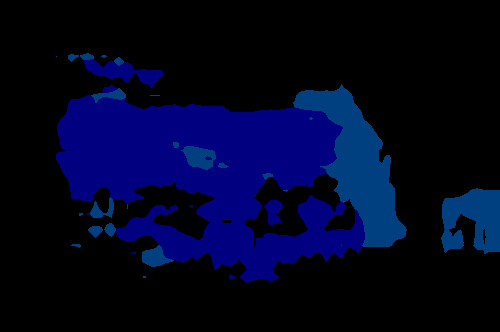} & 
   \includegraphics[width=\imgsize]{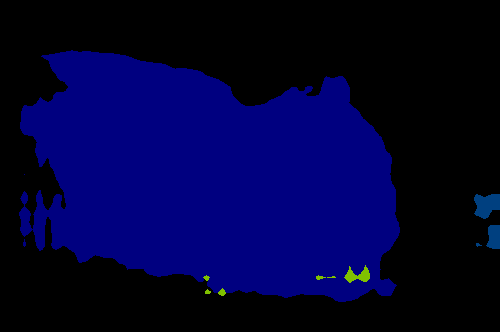} & 
   \includegraphics[width=\imgsize]{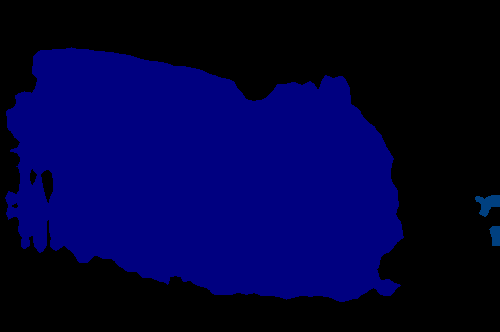} &
   \includegraphics[width=\imgsize]{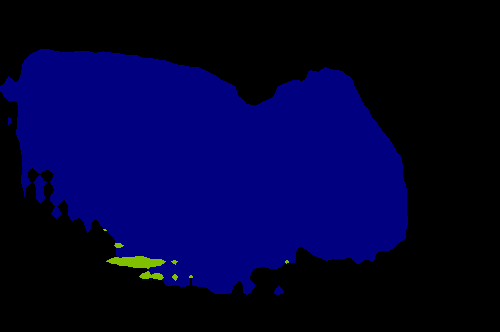}  \\ 
   \hline
   \multirow{2}{*}{\rotatebox{90}{\hspace{+4ex}10-10}} & 
   \includegraphics[trim = 0 0 0 -6mm,clip, width=\imgsize]{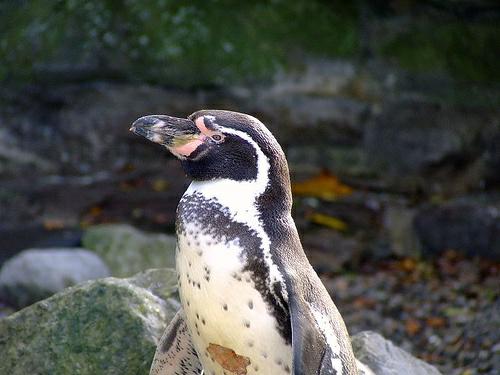} &
   \includegraphics[width=\imgsize]{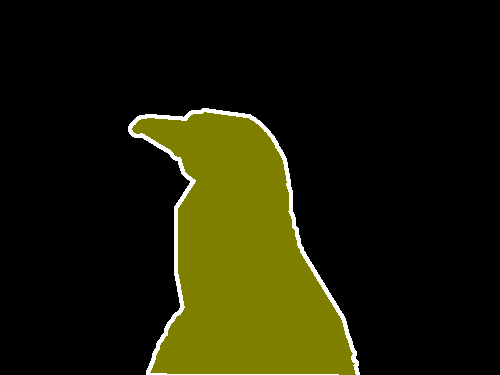} & 
   \includegraphics[width=\imgsize]{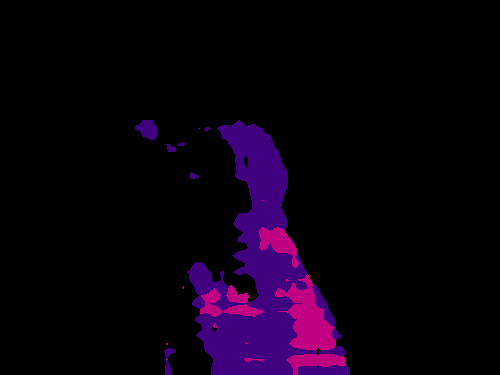} & 
   \includegraphics[width=\imgsize]{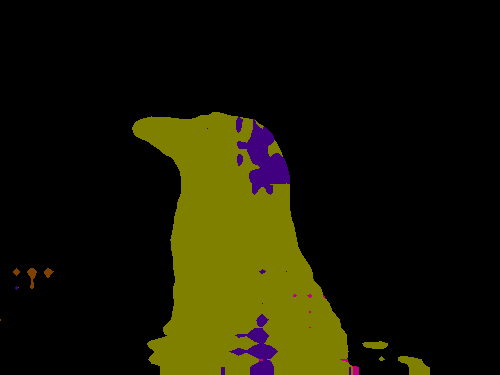} & 
   \includegraphics[width=\imgsize]{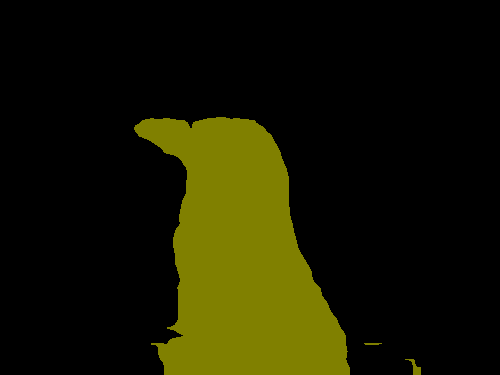} & 
   \includegraphics[width=\imgsize]{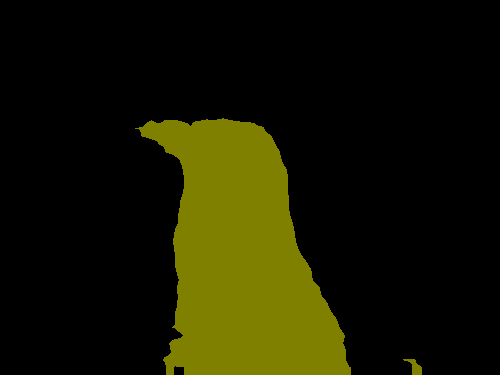} &
   \includegraphics[width=\imgsize]{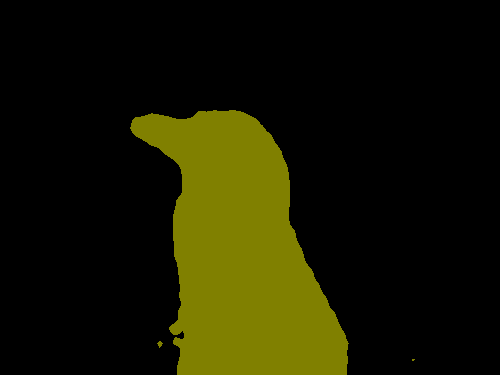}  \\ 
   & 
   \includegraphics[width=\imgsize]{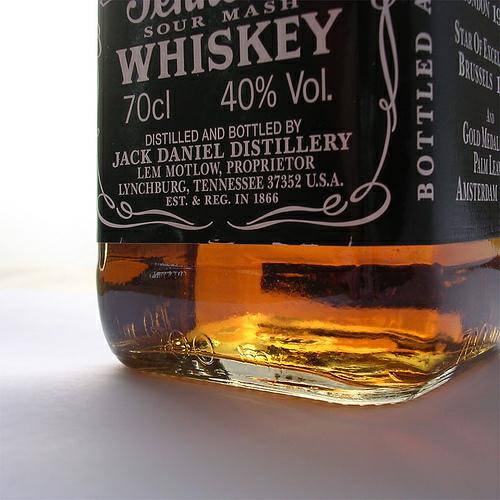} &
   \includegraphics[width=\imgsize]{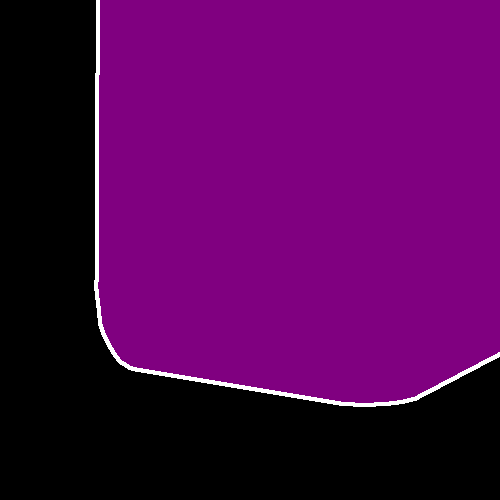} & 
   \includegraphics[width=\imgsize]{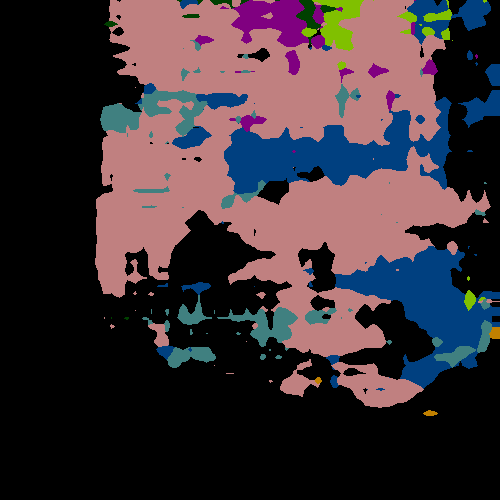} & 
   \includegraphics[width=\imgsize]{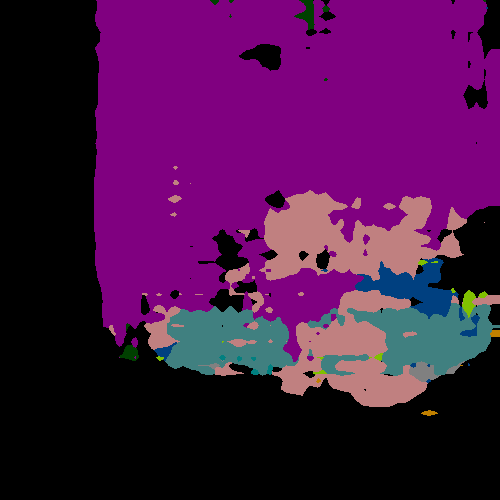} & 
   \includegraphics[width=\imgsize]{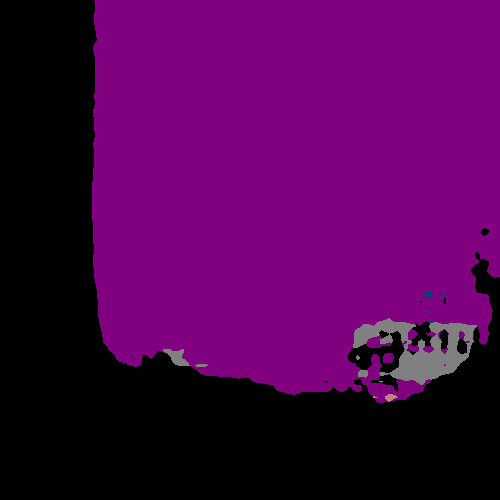} & 
   \includegraphics[width=\imgsize]{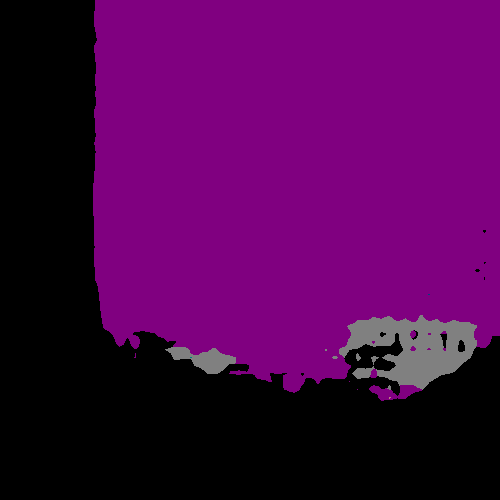} &
   \includegraphics[width=\imgsize]{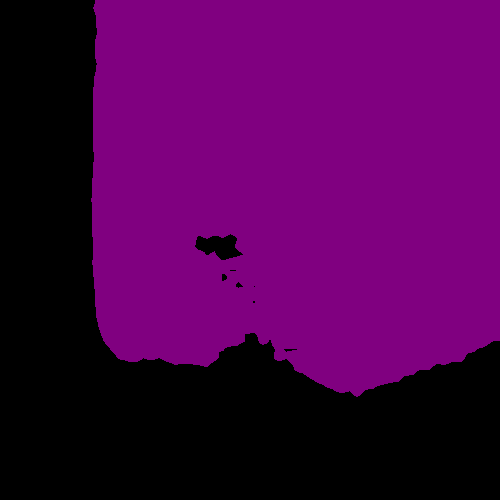}  \\ 
   \hline
   \multirow{2}{*}{\rotatebox{90}{\hspace{+4ex}10-5}} & 
   \includegraphics[trim = 0 0 0 -6mm,clip, width=\imgsize]{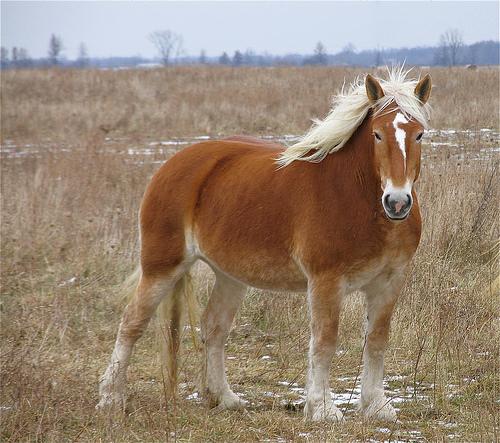} &
   \includegraphics[width=\imgsize]{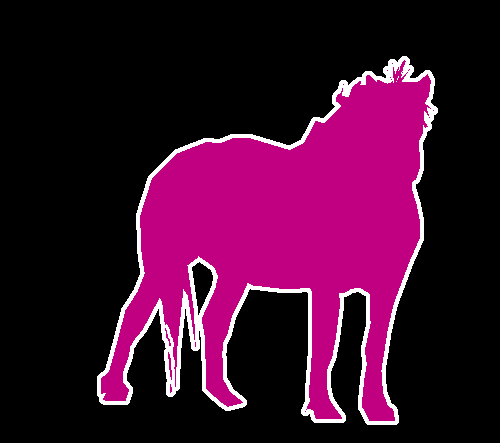} & 
   \includegraphics[width=\imgsize]{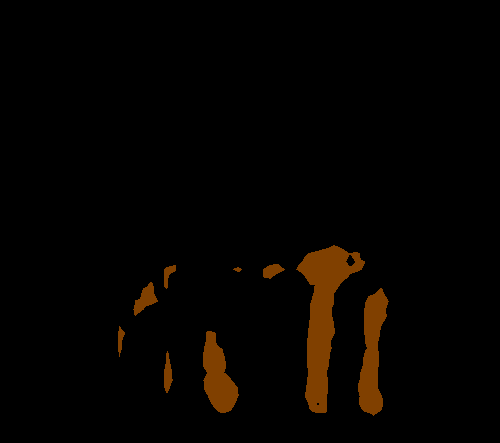} & 
   \includegraphics[width=\imgsize]{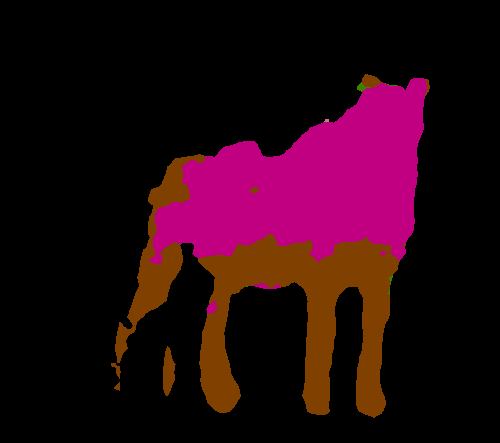} & 
   \includegraphics[width=\imgsize]{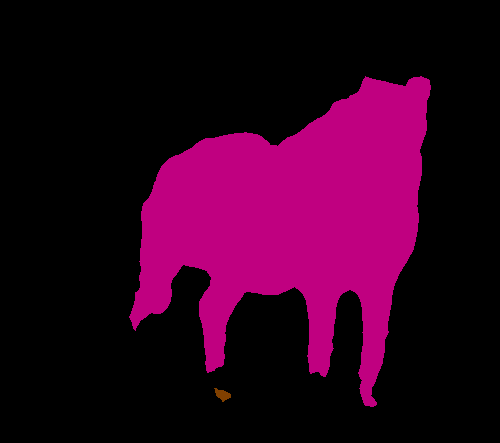} & 
   \includegraphics[width=\imgsize]{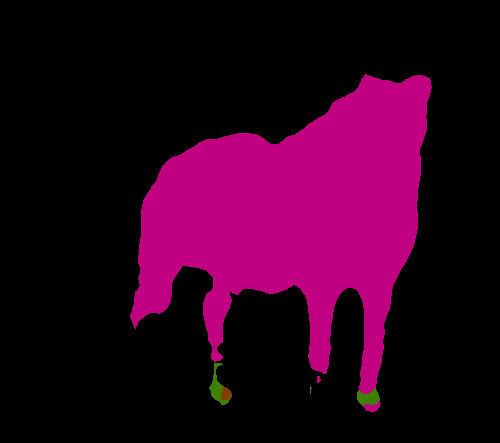} &
   \includegraphics[width=\imgsize]{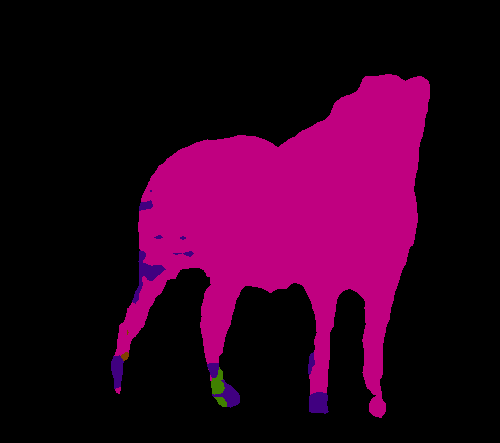}  \\ 
   & 
   \includegraphics[width=\imgsize]{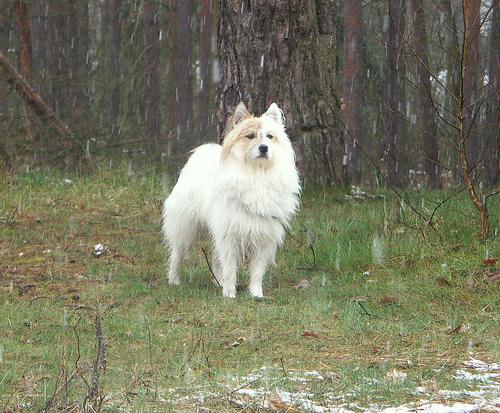} &
   \includegraphics[width=\imgsize]{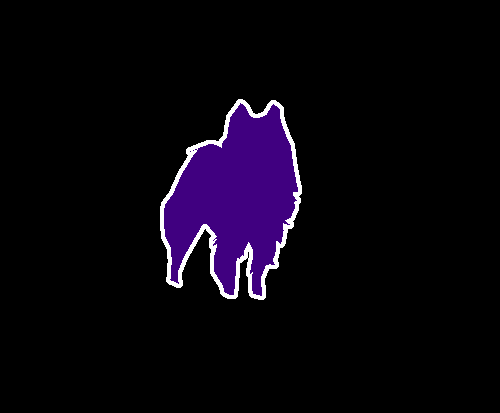} & 
   \includegraphics[width=\imgsize]{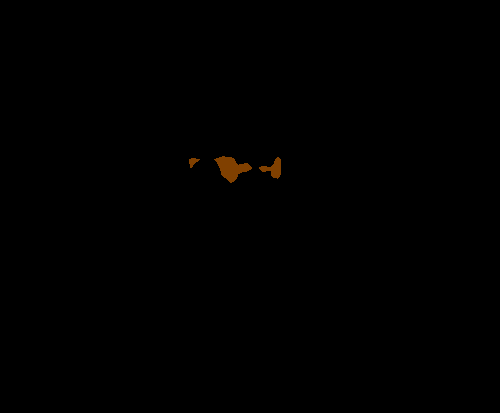} & 
   \includegraphics[width=\imgsize]{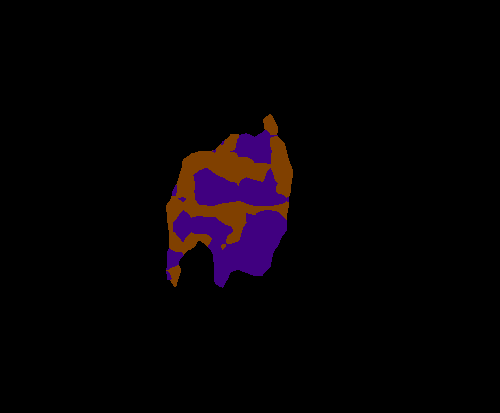} & 
   \includegraphics[width=\imgsize]{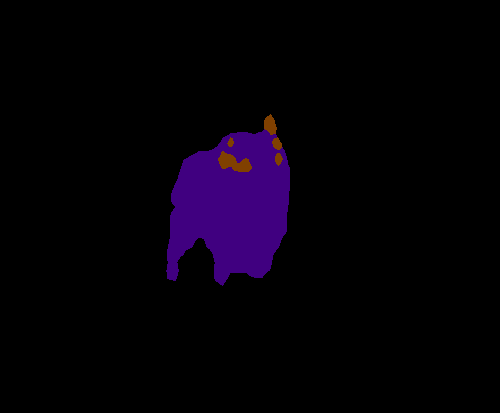} & 
   \includegraphics[width=\imgsize]{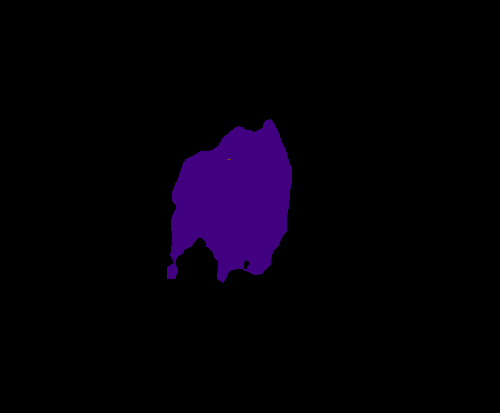} &
   \includegraphics[width=\imgsize]{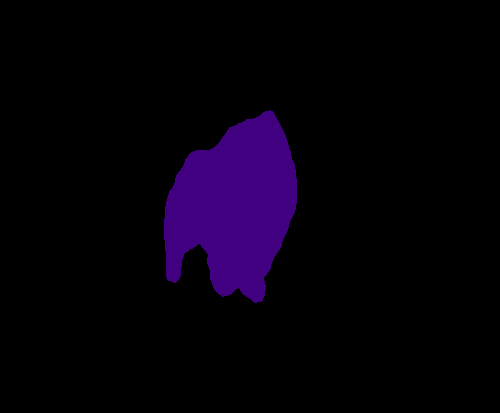}  \\ 
   \hline
   \multirow{2}{*}{\rotatebox{90}{\hspace{+4ex}10-1}} & 
   \includegraphics[trim = 0 0 0 -6mm,clip, width=\imgsize]{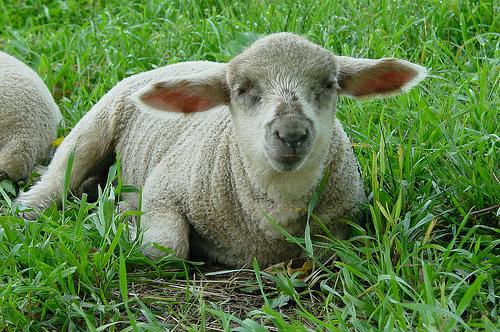} &
   \includegraphics[width=\imgsize]{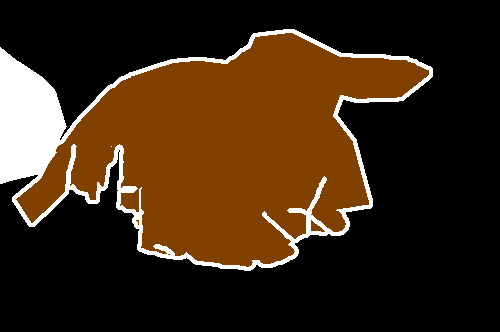} & 
   \includegraphics[width=\imgsize]{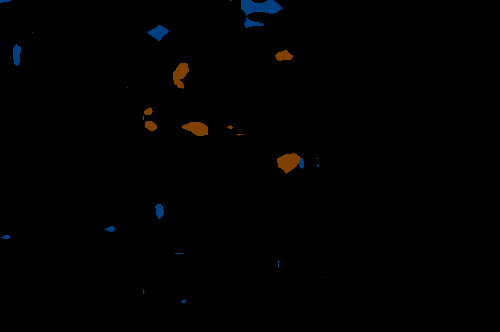} & 
   \includegraphics[width=\imgsize]{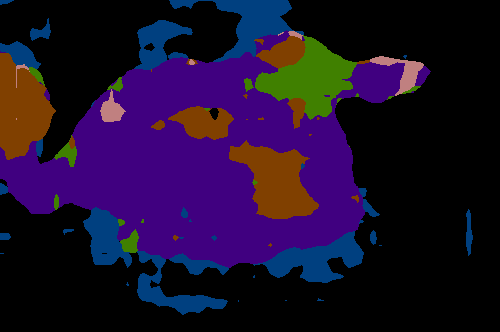} & 
   \includegraphics[width=\imgsize]{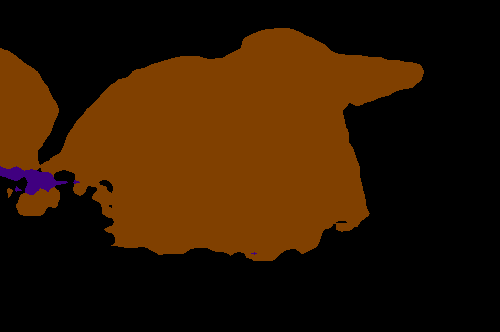} & 
   \includegraphics[width=\imgsize]{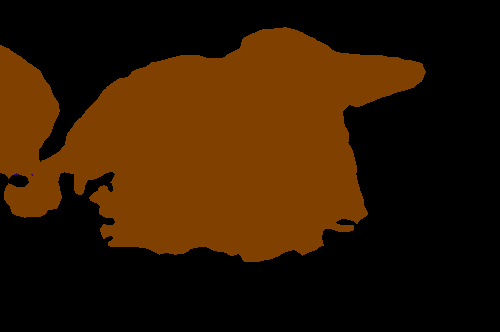} &
   \includegraphics[width=\imgsize]{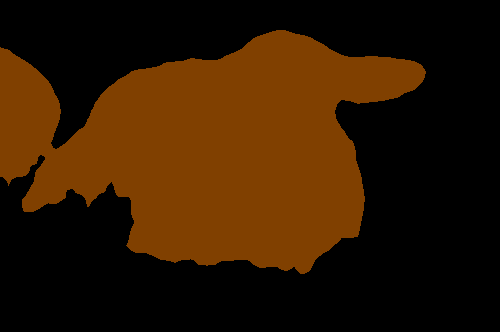}  \\ 
   & 
   \includegraphics[width=\imgsize]{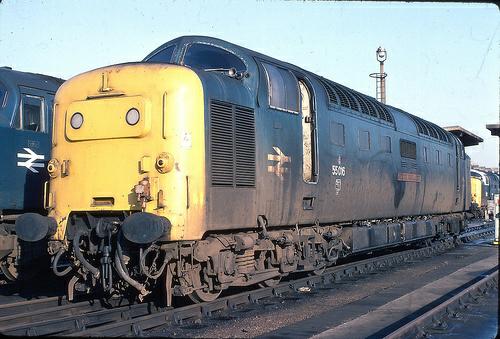} &
   \includegraphics[width=\imgsize]{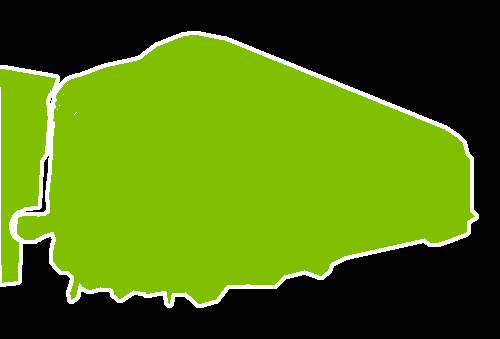} & 
   \includegraphics[width=\imgsize]{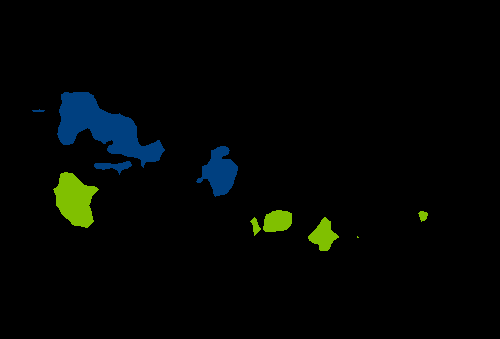} & 
   \includegraphics[width=\imgsize]{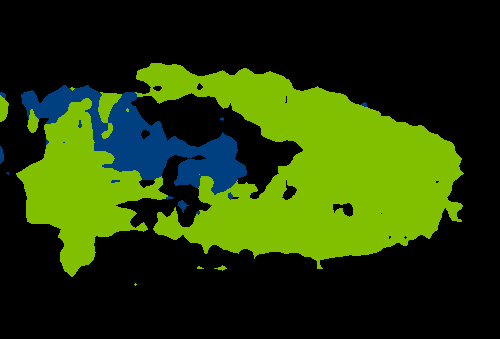} & 
   \includegraphics[width=\imgsize]{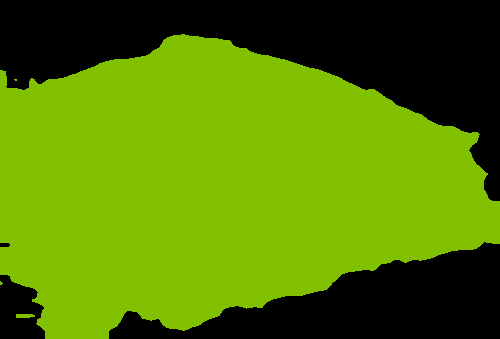} & 
   \includegraphics[width=\imgsize]{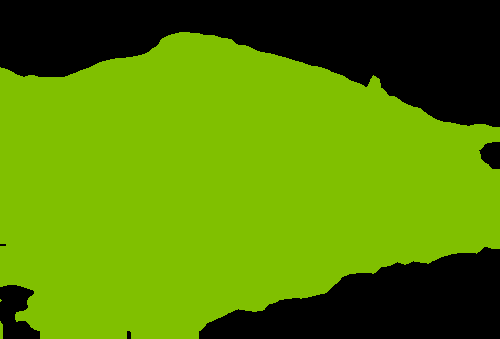} &
   \includegraphics[width=\imgsize]{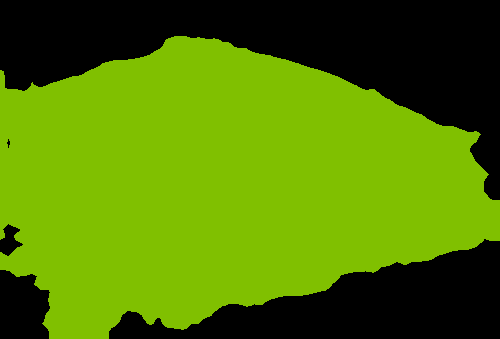}  \\

\end{tabular}
\end{subfigure}

\caption{Qualitative results on disjoint incremental setups (\textit{best viewed in colors}).}
\label{fig:qual_res}
\end{figure*}
\renewcommand{\imgsize}{25mm}
\begin{figure*}[htbp]
\centering

\setlength{\tabcolsep}{1.3pt} %
\renewcommand{\arraystretch}{0.6}
\centering
\begin{subfigure}[htbp]{1\textwidth}
\centering

\begin{tabular}{cccc}
   \includegraphics[width=\imgsize]{suppl_imgs/incr_res/IMG/0026.jpg} &
   \includegraphics[width=\imgsize]{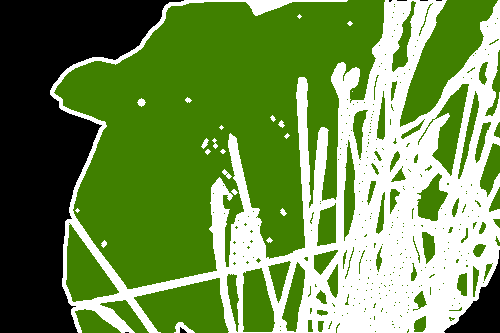} & 
   \includegraphics[width=\imgsize]{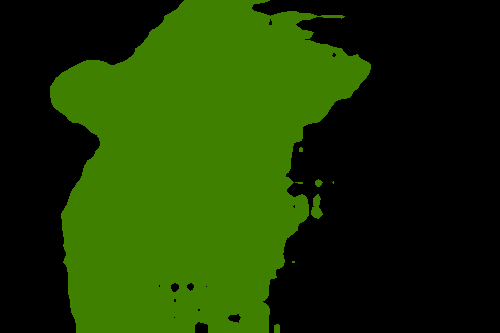} & 
   \includegraphics[width=\imgsize]{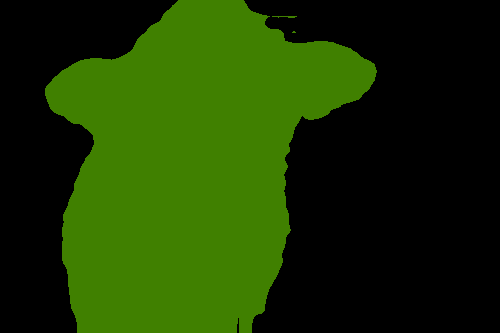} \\[0.5ex]
   RGB  & Ground Truth  & Joint Training  & Step 0 
\end{tabular}

\vspace{0.3cm}

\begin{tabular}{c<{\hspace{1mm}}ccccc}
   \multirow{1}{*}[2.5em]{\rotatebox{90}{FT}} &  
   \includegraphics[width=\imgsize]{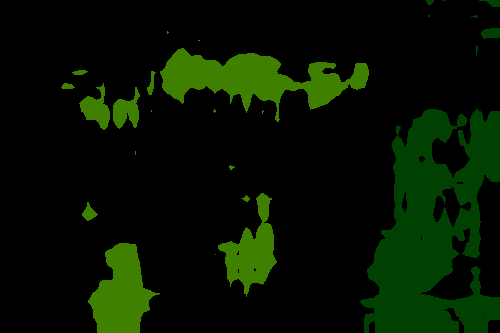} &
   \includegraphics[width=\imgsize]{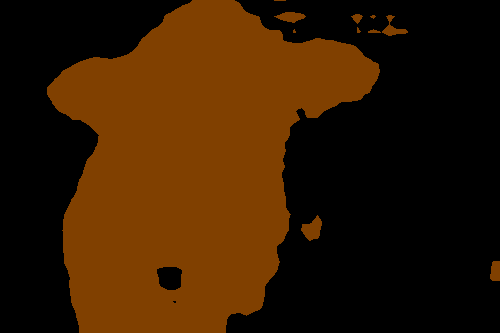} & 
   \includegraphics[width=\imgsize]{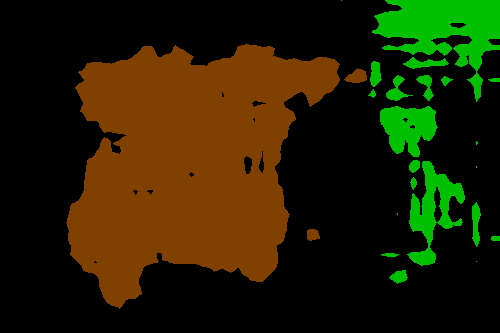} & 
   \includegraphics[width=\imgsize]{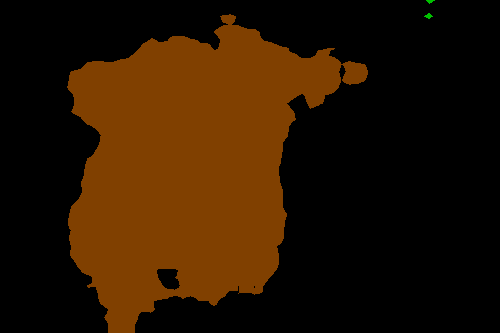} & 
   \includegraphics[width=\imgsize]{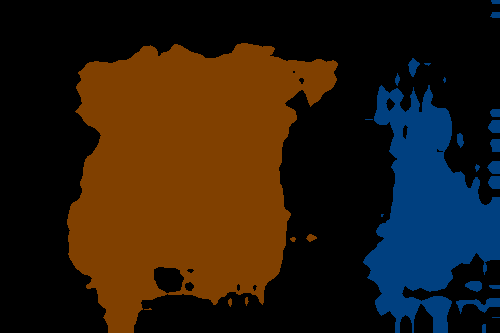} \\
   
   \multirow{1}{*}[3.4em]{\rotatebox{90}{Inpaint.}} &  
   \includegraphics[width=\imgsize]{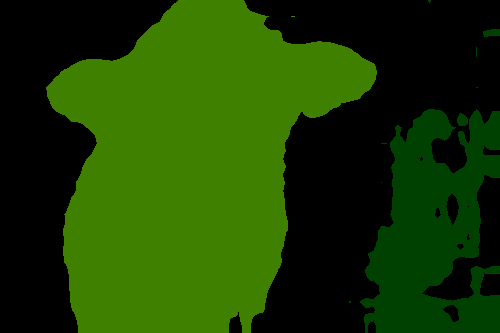} &
   \includegraphics[width=\imgsize]{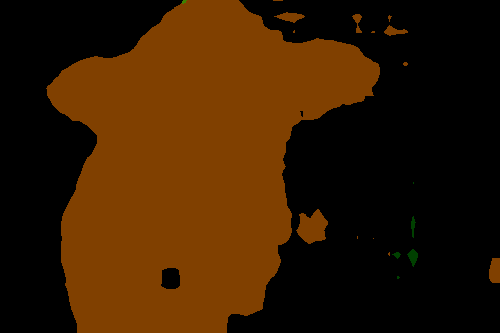} & 
   \includegraphics[width=\imgsize]{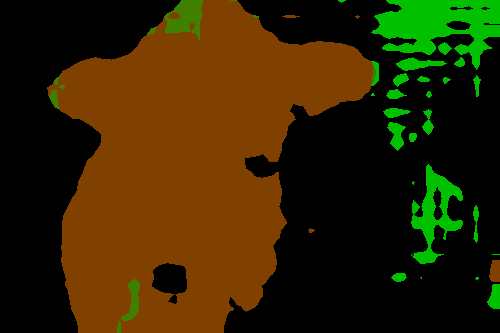} & 
   \includegraphics[width=\imgsize]{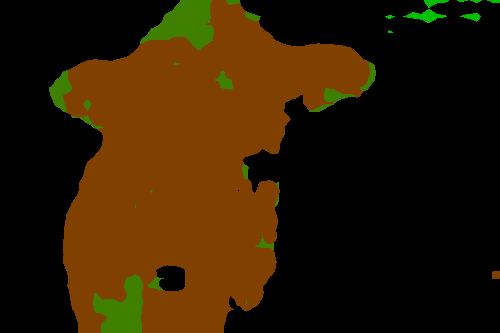} & 
   \includegraphics[width=\imgsize]{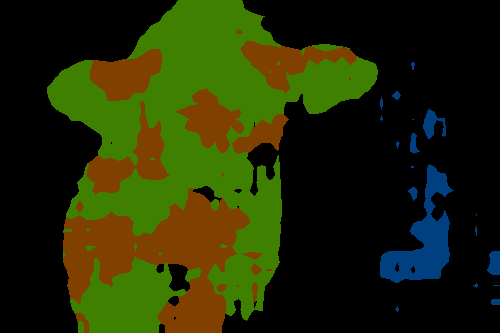} \\
   
   \multirow{1}{*}[4.4em]{\rotatebox{90}{Ours (GAN)}} &  
   \includegraphics[width=\imgsize]{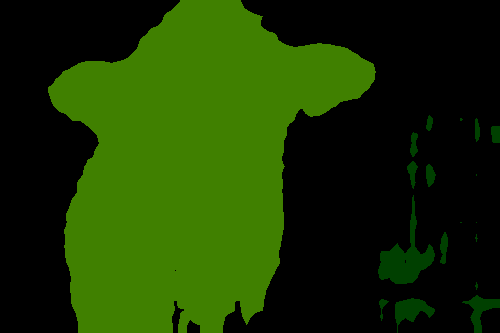} &
   \includegraphics[width=\imgsize]{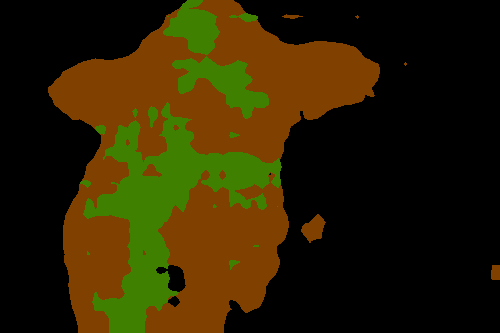} & 
   \includegraphics[width=\imgsize]{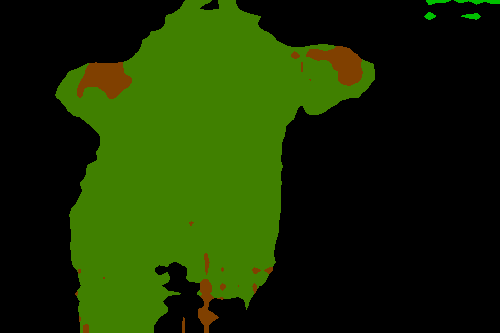} & 
   \includegraphics[width=\imgsize]{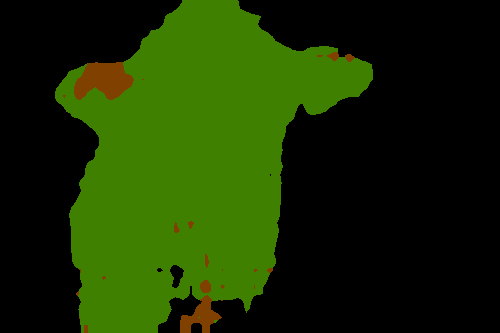} & 
   \includegraphics[width=\imgsize]{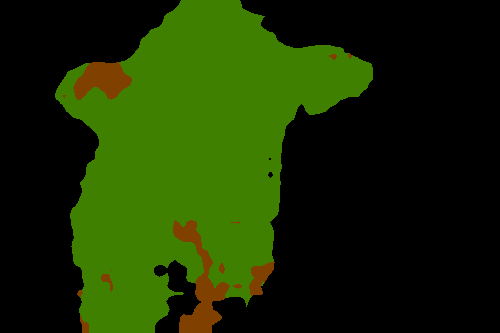} \\
  
   \multirow{1}{*}[4.1em]{\rotatebox{90}{Ours (Web)}} &  
   \includegraphics[width=\imgsize]{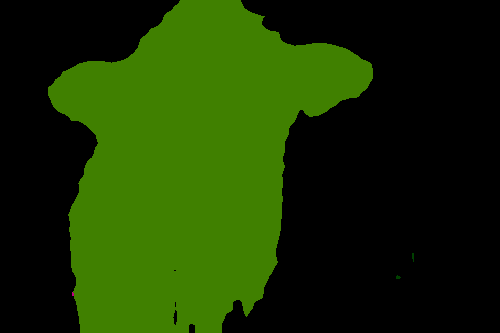} &
   \includegraphics[width=\imgsize]{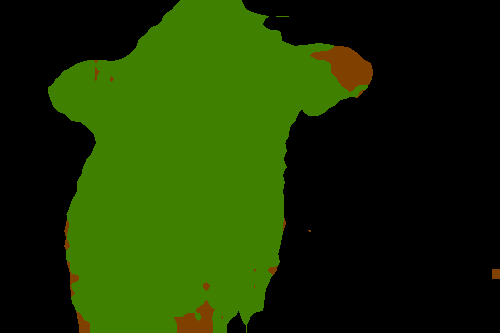} & 
   \includegraphics[width=\imgsize]{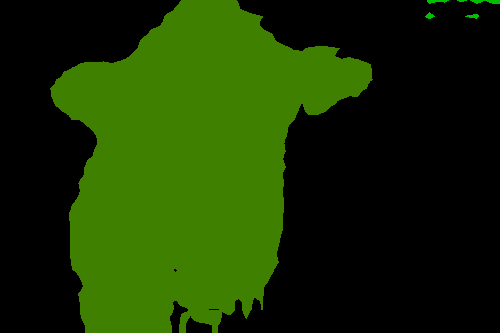} & 
   \includegraphics[width=\imgsize]{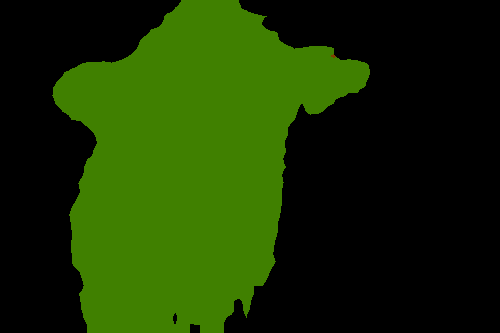} & 
   \includegraphics[width=\imgsize]{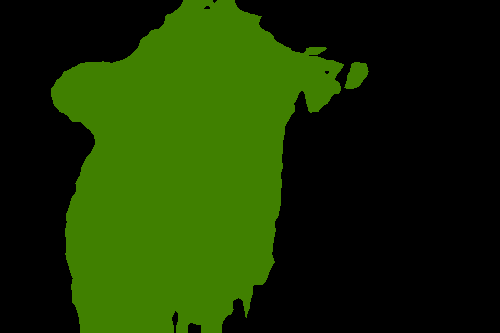} \\[0.5ex]
   
   & Step 1  & Step 2  & Step 3  & Step 4  & Step 5 \\
  &\footnotesize  \textit{(added: potted-plant)} & \footnotesize  \textit{(added: sheep)} &  \footnotesize  \textit{(added:sofa)} & \footnotesize  \textit{(added: train)} & \footnotesize  \textit{(added: tv/monitor)}  
\end{tabular}

\vspace{0.8cm}

\begin{tabular}{cccc}
   \includegraphics[width=\imgsize]{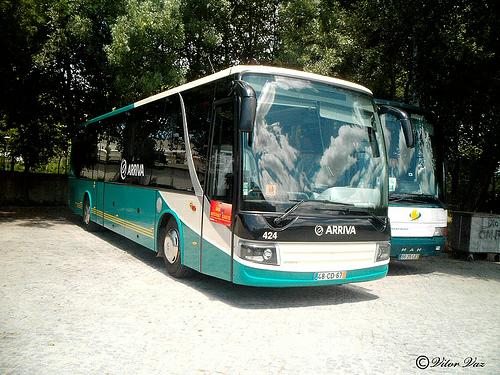} &
   \includegraphics[width=\imgsize]{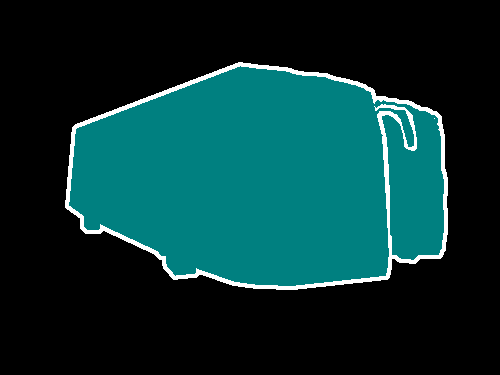} & 
   \includegraphics[width=\imgsize]{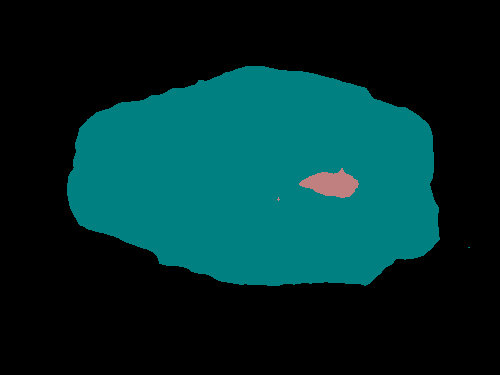} & 
   \includegraphics[width=\imgsize]{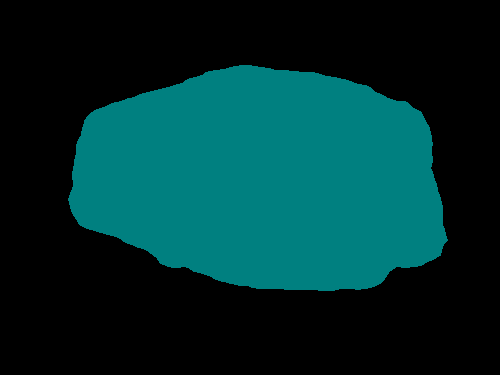} \\[0.5ex]
    RGB  & Ground Truth  & Joint Training  & Step 0 
\end{tabular}

\vspace{0.3cm}

\begin{tabular}{c<{\hspace{1mm}}ccccc}
   \multirow{1}{*}[2.7em]{\rotatebox{90}{FT}} &  
   \includegraphics[width=\imgsize]{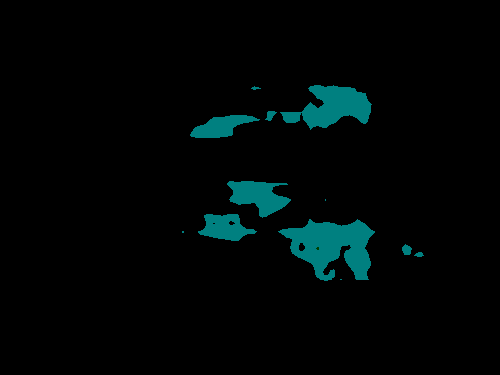} &
   \includegraphics[width=\imgsize]{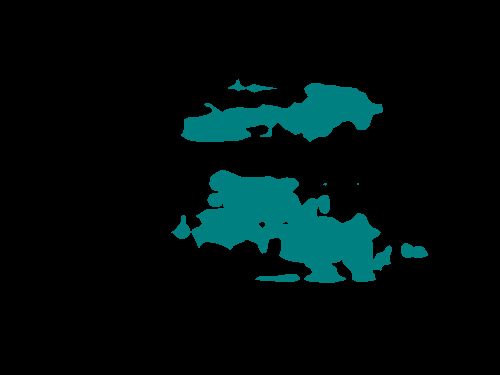} & 
   \includegraphics[width=\imgsize]{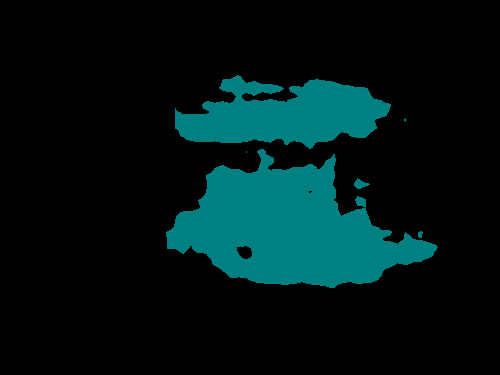} & 
   \includegraphics[width=\imgsize]{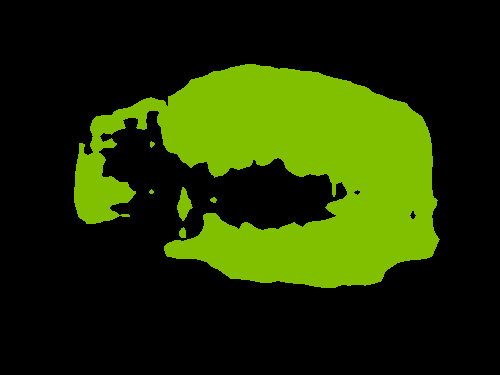} & 
   \includegraphics[width=\imgsize]{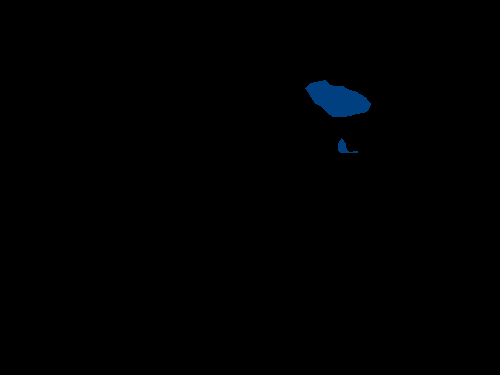} \\
   
   \multirow{1}{*}[3.8em]{\rotatebox{90}{Inpaint.}} &  
   \includegraphics[width=\imgsize]{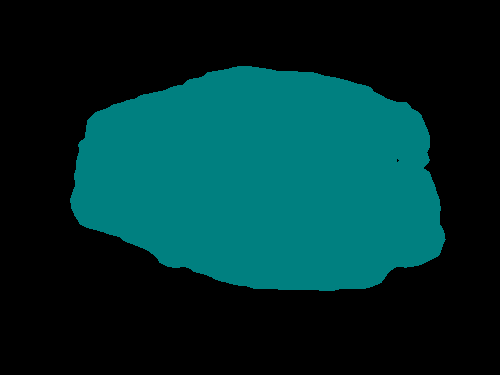} &
   \includegraphics[width=\imgsize]{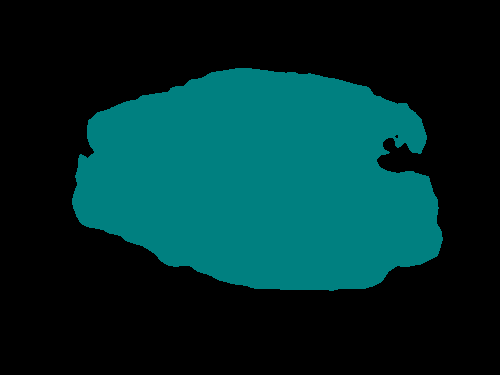} & 
   \includegraphics[width=\imgsize]{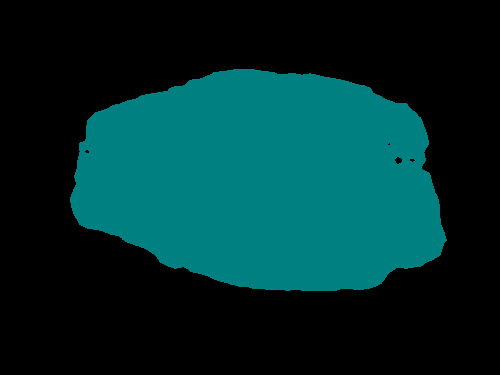} & 
   \includegraphics[width=\imgsize]{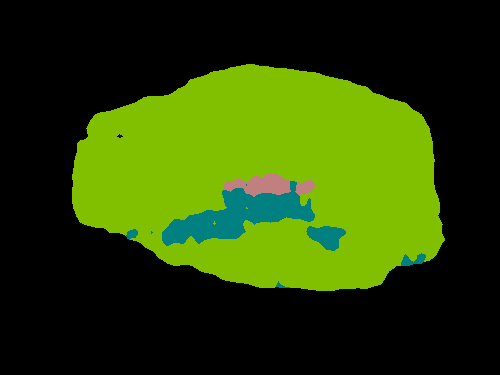} & 
   \includegraphics[width=\imgsize]{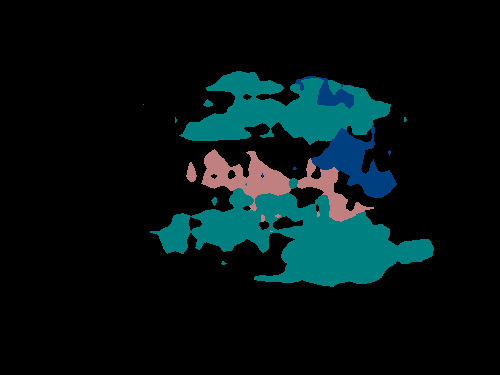} \\
   
   \multirow{1}{*}[4.6em]{\rotatebox{90}{Ours (GAN)}} &  
   \includegraphics[width=\imgsize]{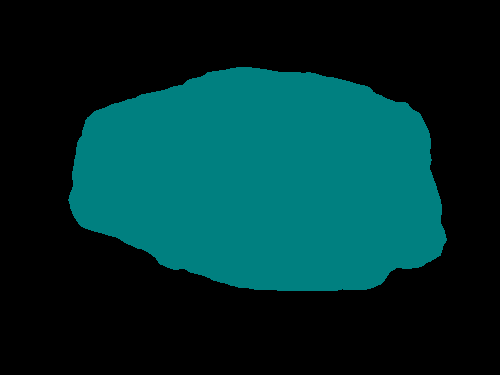} &
   \includegraphics[width=\imgsize]{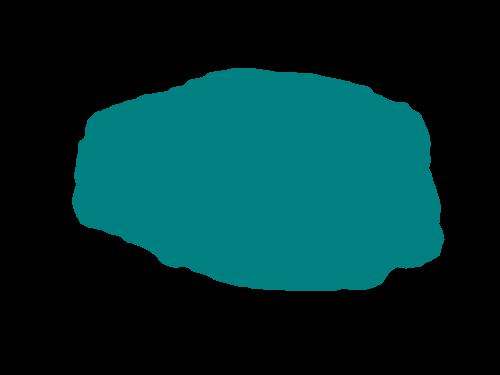} & 
   \includegraphics[width=\imgsize]{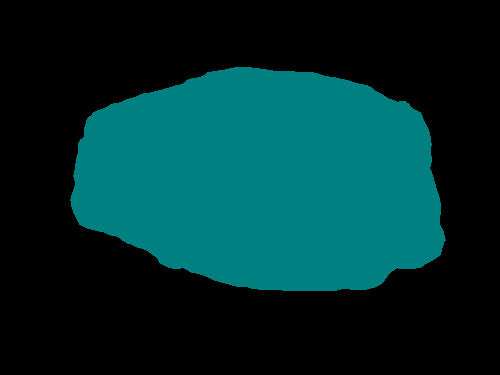} & 
   \includegraphics[width=\imgsize]{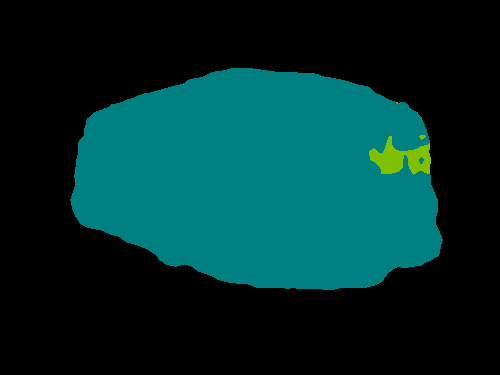} & 
   \includegraphics[width=\imgsize]{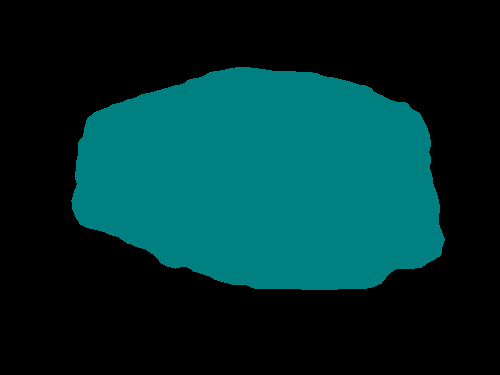} \\
  
   \multirow{1}{*}[4.3em]{\rotatebox{90}{Ours (Web)}} &  
   \includegraphics[width=\imgsize]{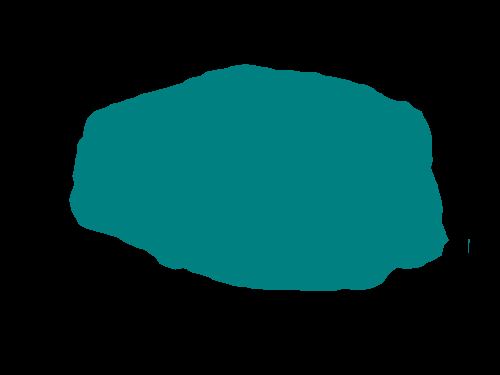} &
   \includegraphics[width=\imgsize]{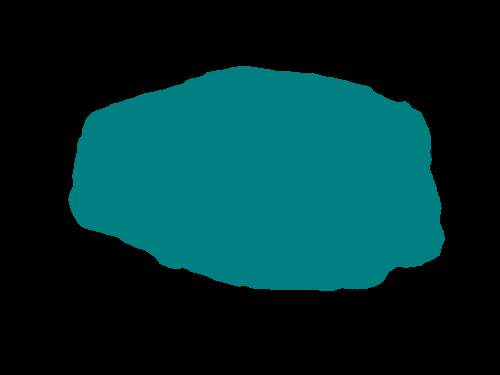} & 
   \includegraphics[width=\imgsize]{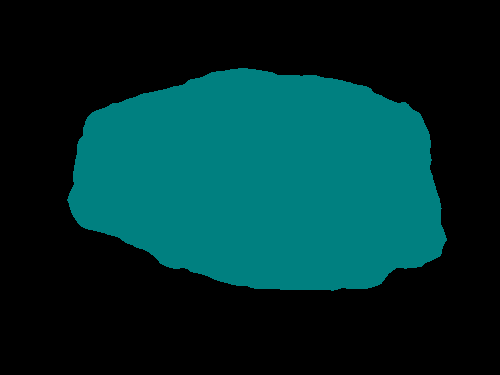} & 
   \includegraphics[width=\imgsize]{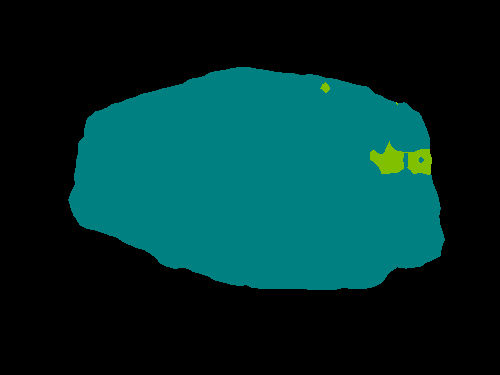} & 
   \includegraphics[width=\imgsize]{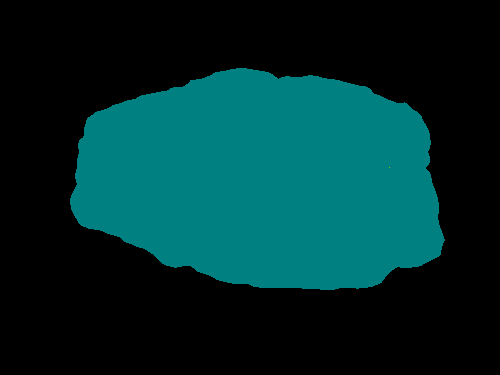} \\[0.5ex]
   
   & Step 1  & Step 2  & Step 3  & Step 4  & Step 5 \\
     &\footnotesize  \textit{(added: potted-plant)} & \footnotesize  \textit{(added: sheep)} &  \footnotesize  \textit{(added:sofa)} & \footnotesize  \textit{(added: train)} & \footnotesize  \textit{(added: tv/monitor)}  
\end{tabular}

\end{subfigure}

\caption{Per-step prediction maps on the 15-1 disjoint incremental setup for different training strategies. %
}

\label{fig:incr_res}
\end{figure*}

\end{document}